%% file: main.tex
\documentclass[]{externalized}

\input{common}

\renewcommand\correspondence[1]{\checkdata[\faEnvelopeO\;Email]{#1}}
\usepackage{array}
\usepackage{textcomp}
\usepackage{stfloats}
\usepackage{verbatim}
\usepackage{breakurl}
\usepackage{bbm}
\usepackage{dsfont}
\usepackage{latexsym}
\usepackage{soul}
\usepackage{float}
\usepackage{tabularx}
\usepackage{colortbl}
\usepackage{todonotes}
\usepackage[edges]{forest}
\usepackage[framemethod=tikz]{mdframed}
\usepackage{hyperref}
\usepackage{xurl}
\usepackage{booktabs,array,tabularx}
\usepackage{tikz}
\usetikzlibrary{svg.path}

\newcommand{\LinkGH}[1]{\href{#1}{\textcolor{black}{\faGithub}}}
\definecolor{hfyellow}{HTML}{FFD21E}
\definecolor{webblue}{HTML}{0000FF}
\newcommand{\LinkHF}[1]{\href{#1}{\textcolor{hfyellow}{\includegraphics[height=1em]{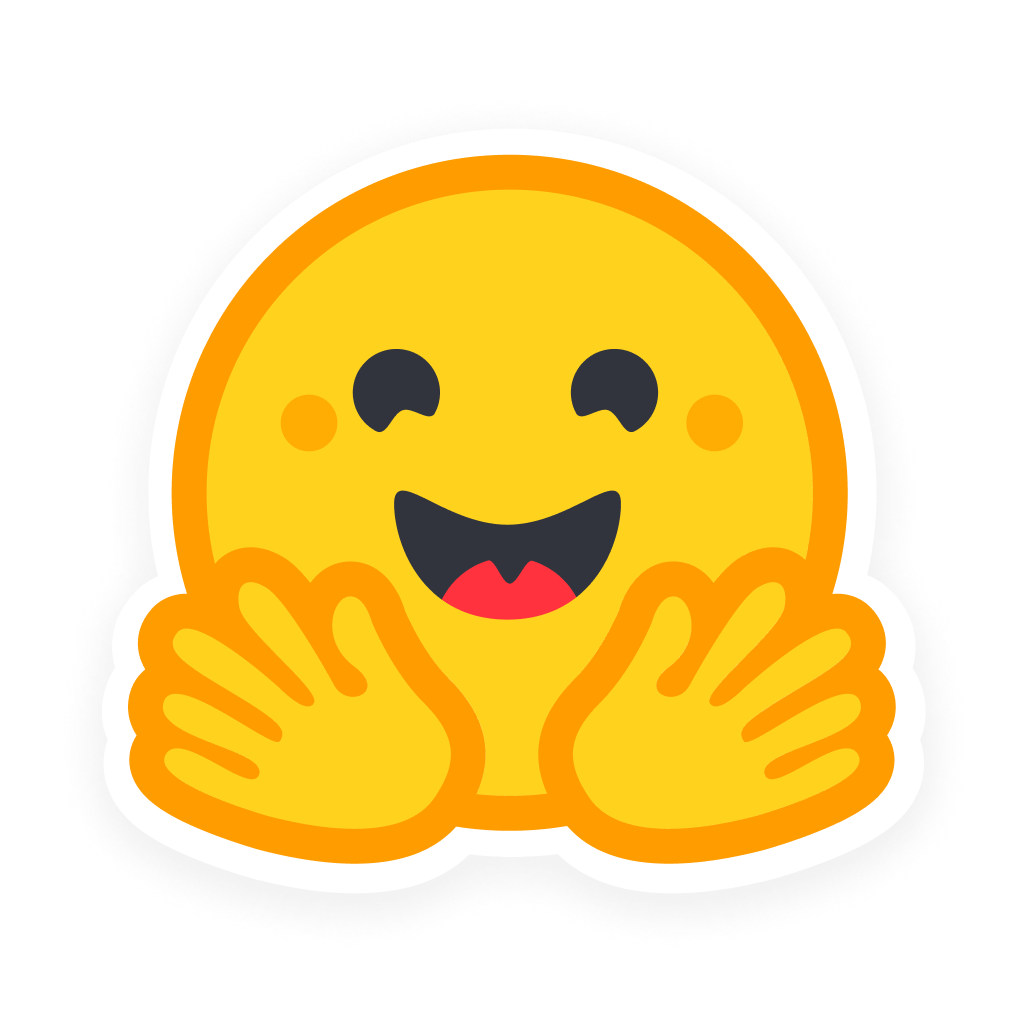}}}}
\newcommand{\LinkWEB}[1]{\href{#1}{\includegraphics[height=1em]{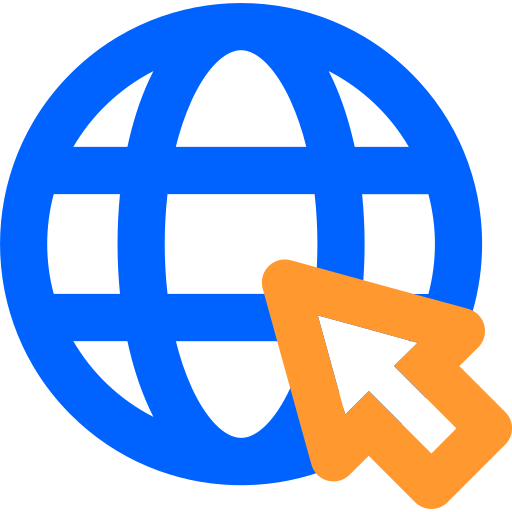}}}

\providecommand{\BenchArxiv}[1]{\href{#1}{arXiv}}
\providecommand{\BenchPaper}[1]{\href{#1}{paper}}
\providecommand{\BenchData}[1]{\href{#1}{data}}

\providecommand{\AppArxiv}[1]{\href{#1}{arXiv}}
\providecommand{\AppPaper}[1]{\href{#1}{paper}}
\definecolor{hidden-draw}{RGB}{20,68,106}
\definecolor{hidden-pink}{RGB}{255,245,247}
\definecolor{lightgray}{gray}{0.9}
\definecolor{lightgreen}{rgb}{0.9, 1, 0.9}
\definecolor{emb}{RGB}{209,228,252}
\definecolor{hidden-blue}{RGB}{194,232,247}
\definecolor{hidden-orange}{RGB}{224,224,224}
\definecolor{hidden-yellow}{RGB}{242,244,193}
\definecolor{output-purple}{RGB}{219,203,231}
\definecolor{output-green}{RGB}{204,231,207}
\definecolor{hiddendraw}{RGB}{10,128,122}
\definecolor{myred2}{HTML}{F875AA}
\definecolor{mypurple2}{HTML}{D2E0FB}
\definecolor{myred}{HTML}{F8F6F4}
\definecolor{mypurple}{HTML}{FFDFDF}
\definecolor{myyellow}{HTML}{FFF6F6}
\definecolor{mygreen}{HTML}{D2E0FB}

\input{math_commands}

\usepackage{tikz}
\usetikzlibrary{fadings}
\usetikzlibrary{mindmap,trees}
\usetikzlibrary{arrows,automata,shapes,positioning,shadows,trees}
\usetikzlibrary{shapes,snakes,shadows}
\usetikzlibrary{shapes.arrows}
\usetikzlibrary{calc,shapes, positioning}
\usetikzlibrary{decorations.text}
\usetikzlibrary{matrix,chains,positioning,decorations.pathreplacing,arrows}
\usetikzlibrary{bayesnet}
\usetikzlibrary{shadows.blur}
\usetikzlibrary{shapes.geometric}

\usepackage{pgfplots}
\usepackage{pgfplotstable}
\usepackage{pgfmath,pgffor}
\usepgfplotslibrary{colorbrewer}
\pgfplotsset{compat = 1.14, cycle list/Set1-8}
\usetikzlibrary{pgfplots.statistics, pgfplots.colorbrewer}
\pgfplotsset{compat=1.8}

\tikzstyle{edge}=[-latex',draw=black!90,shorten <=1pt,shorten >=1pt]
\tikzstyle{redge}=[latex'-,draw=black!90,shorten <=1pt,shorten >=1pt]
\tikzstyle{dedge}=[latex'-latex',draw=black!90,shorten <=1pt,shorten >=1pt]
\tikzstyle{block}=[draw, text width=5em,align=center,shape=rectangle, rounded corners, align=center]
\tikzstyle{nobox}=[align=center]

\title{What Improves, Who Drives It: \\A Roadmap Towards Reliable Recursive Self-Improvement}

\author[1,*]{Zhaoxu Liu}
\author[1,*]{Yuyuan Feng}
\author[1]{Sihan Huang}
\author[1]{Haocong Niu}
\author[2]{Chaobin Yang}
\author[3]{Yujing Zhang}
\author[3]{Chang Yang}
\author[1]{Wei Chen}
\author[4]{Jiwei Zheng}
\author[4]{Jiao Li}
\author[2]{Zhihong Zhang}
\author[3]{Xiao Huang}
\author[1]{Xin Wang}
\author[1,\dag]{Qinggang Zhang}
\author[1,\dag]{Yi Chang}

\affiliation[1]{Jilin University}
\affiliation[2]{Xiamen University}
\affiliation[3]{The Hong Kong Polytechnic University\\}
\affiliation[4]{Chinese Academy of Medical Sciences
and Peking Union Medical College}

\contribution[*]{Equal Contribution}
\contribution[\dag]{Corresponding Authors}

\correspondence{\email{qinggangzhang@jlu.edu.cn}}

\abstract{ Recent advances in Large Language Models (LLMs) have enabled AI systems to evolve beyond conventional question answering and content generation toward agents capable of performing complex, long-horizon tasks in dynamic environments. Along with this development, agent improvement has expanded from temporal output refinement to persistent updates of models, harnesses, and further to the continuous adaptation of environments and learning conditions. Meanwhile, feedback acquisition, utilization, and evolution have become increasingly automated, enabling agents to guide subsequent improvements more autonomously. However, sustained improvements in agent capabilities do not necessarily translate into enhanced abilities to generate further improvements. Recursive Self-Improvement extends the improvement to the mechanism responsible for producing subsequent improvements. Nevertheless, achieving reliable RSI remains challenging, particularly in identifying appropriate components for evolution and obtaining valid feedback signals to guide evolutionary updates. Inappropriate evolution targets and feedback sources may lead to ineffective modifications, capability degradation, and error accumulation across recursive iterations. To systematically analyze these two challenges for reliable RSI, this survey introduces a two-axis taxonomy comprising \textbf{What Improves} and \textbf{Who Drives Improve}, characterizing agent evolution from the complementary perspectives of evolution targets and driving mechanisms. Building on this taxonomy, we organize existing studies into seven interrelated milestones, covering output refinement, foundation model training, agent harness evolution, environment synthesis, feedback automation, agent–environment co-evolution, and recursive self-improvement. Furthermore, we examine the construction, self-modification, cross-generational reuse, and capability evaluation of improvement mechanisms, and systematically review representative benchmarks, practical applications, and promising research opportunities toward reliable RSI. All the related resources, including research papers, open-source data, and projects, are collected for the community at \textcolor{blue}{\url{https://github.com/DEEP-JLU/Awesome-RSI}}.

}

\begin{document}
\maketitle
\newpage
\tableofcontents
\newpage

\input{section/introduction}

\input{section/paper_list}
\input{section/overview}

\input{section/agent_evolution}

\input{section/environment_evolution}
\input{section/feedback_autonomy}
\input{section/co_evolution}

\input{section/rsi}

\input{section/benchmark}

\input{section/application}
\input{section/future_directions}
\input{section/conclusion}

\newpage
\bibliographystyle{abbrvnat}
\bibliography{reference}

\end{document}

%% file: common.tex
\usepackage[T1]{fontenc}
\usepackage[utf8]{inputenc}
\usepackage{microtype}
\usepackage{inconsolata}
\usepackage{graphicx}
\usepackage{hyperref}
\usepackage{url}
\usepackage{booktabs}
\usepackage{amsfonts}
\usepackage{amsmath}
\usepackage{amssymb}
\usepackage{amsthm}
\usepackage{mathtools}
\usepackage{enumitem}
\usepackage{multirow}
\usepackage{subcaption}
\usepackage{xcolor}
\usepackage{listings}
\usepackage{natbib}
\usepackage{multicol}
\usepackage{pifont}
\usepackage{xspace}
\usepackage{wrapfig}
\AtBeginDocument{%
  \providecommand\BibTeX{{%
    \normalfont B\kern-0.5em{\scshape i\kern-0.25em b}\kern-0.8em\TeX}}}

\makeatletter
\DeclareRobustCommand\onedot{\futurelet\@let@token\@onedot}
\def\@onedot{\ifx\@let@token.\else.\null\fi}

\makeatother

\definecolor{codegreen}{rgb}{0,0.6,0}
\definecolor{codegray}{rgb}{0.5,0.5,0.5}
\definecolor{codepurple}{rgb}{0.58,0,0.82}
\definecolor{backcolour}{rgb}{0.95,0.95,0.92}

\lstdefinestyle{mystyle}{
  backgroundcolor=\color{backcolour},
  commentstyle=\color{codegreen},
  keywordstyle=\color{magenta},
  numberstyle=\tiny\color{codegray},
  stringstyle=\color{codepurple},
  basicstyle=\ttfamily\footnotesize,
  breakatwhitespace=false,
  breaklines=true,
  captionpos=b,
  keepspaces=true,
  numbers=left,
  numbersep=5pt,
  showspaces=false,
  showstringspaces=false,
  showtabs=false,
  tabsize=2
}

%% file: math_commands.tex
\usepackage{amsmath,amsfonts,bm}

\def\eqref#1{equation~\ref{#1}}

\def\1{\bm{1}}

\DeclareMathAlphabet{\mathsfit}{\encodingdefault}{\sfdefault}{m}{sl}
\SetMathAlphabet{\mathsfit}{bold}{\encodingdefault}{\sfdefault}{bx}{n}



%% file: section/introduction.tex
\section{Introduction}
\label{sec:introduction}

Recent advances in large language models (LLMs) in language understanding, complex reasoning, and decision-making have enabled AI systems to go beyond conventional question answering and content generation toward agents capable of performing complex, long-horizon tasks in dynamic environments~\cite{openai2024gpt4technicalreport,DeepSeekR1,liu2023agentbench,zhou2023webarena}. Unlike conventional model inference, agent performance depends not only on the underlying foundation model but also on multiple system components, including task planning, workflow orchestration, memory management, tool use, and environmental interaction~\cite{yao2023reactsynergizingreasoningacting,schick2023toolformer,park2023generativeagentsinteractivesimulacra,zhang2024aflow,wang2023voyager}. Meanwhile, agents continuously generate task outcomes, interaction trajectories, and experiences of success and failure during execution, providing valuable information for subsequent system improvement~\cite{shinn2023reflexion,zhao2024expel,wang2023voyager}. How to transform such execution experiences into system capabilities that can be accumulated and reused across runs has become an important research question in agent improvement.

Toward this goal, recent studies have progressively expanded from temporal output refinement to persistent updates of models and harnesses, and further to the continuous adaptation of environments and learning conditions~\cite{SelfRefine,CRITIC,instructgpt,xu2025amem,poet}. Meanwhile, the acquisition, utilization, and evolution of feedback have also become increasingly automated, reducing reliance on human intervention and enabling systems to guide subsequent updates more autonomously~\cite{codet,constitutional,agent_judge,meta_rewarding}. Together, these advances have driven sustained and recursive improvements in the capabilities of AI agent systems. However, capability gains in one iteration do not guarantee continued improvement in subsequent iterations. What's more, improving agent capabilities does not necessarily enhance the ability to generate further improvements. Recursive self-improvement (RSI) enables agents to enhance not only their task-solving capabilities but also the mechanisms that drive further capability gains~\cite{good1966ultraintelligent,schmidhuber2003goedel,zhang2026hyperagents}.

Existing surveys have examined self-improvement from two complementary
perspectives. \emph{Mechanism-oriented surveys} map the technical
routes through which agent systems improve, organizing methods
by editable components and update
timing~\cite{fang2025comprehensivesurveyselfevolvingai,
gao2026surveyselfevolvingagentswhat,wang2026divingintoreliableselfevolvingagents},
update targets and learning
signals~\cite{ren2026selfimprovementsmodernagenticsystems},
or model--environment
co-evolution~\cite{xiang2026systematic}.
\emph{Recursion-oriented surveys} examine the boundaries and
development of RSI, emphasizing the distinction between repeated
optimization and recursive improvement, the degree of loop closure,
and increasing autonomy over the improvement process
itself~\cite{chen2026recursiveselfimprovementaibounded,
duan2026aibuilthumansgenuine,li2026towards}.
These studies provide important foundations for understanding the different levels and mechanisms of recursive agent improvement.

However, achieving reliable recursive self-improvement remains challenging and requires both selecting \emph{appropriate components
for evolution} and using valid \emph{signals to guide the evolution updates}.
On the one hand, capability bottlenecks may originate in different system components and their interactions, making it hard to identify what to improve. Noise and bias in execution experiences and automated feedback may further lead to inappropriate updates, preventing modifications from addressing the actual limitations of the system~\cite{reward_refinement,ICLR2024_8b4add8b,fair_evaluators}. On the other hand, even when appropriate components are selected for evolution, unreliable feedback signals or discrepancies between evaluation criteria and actual objectives may steer the improvement process away from its intended direction.
Moreover, modifications across iterations may cause capability degradation and interference among system components, undermining the long-term retention of improvements~\cite{shang2026backfires,wang2026compound}. Even when a system continuously produces better-performing agents, its improvement process may still rely on fixed update strategies, and such gains alone do not establish whether its ability to generate further improvements has increased~\cite{zelikman2024stop,srikanth2026aide2,zhang2026darwin,wang2026rethinking}. Therefore, achieving reliable RSI requires not only identifying appropriate evolution targets but also obtaining accurate and effective feedback signals to guide their evolution. Furthermore, whether these improvements can be sustained and translated into genuine gains in recursive improvement capability requires rigorous evaluation.

\begin{figure*}[t]
    \centering
    \includegraphics[width=\textwidth]{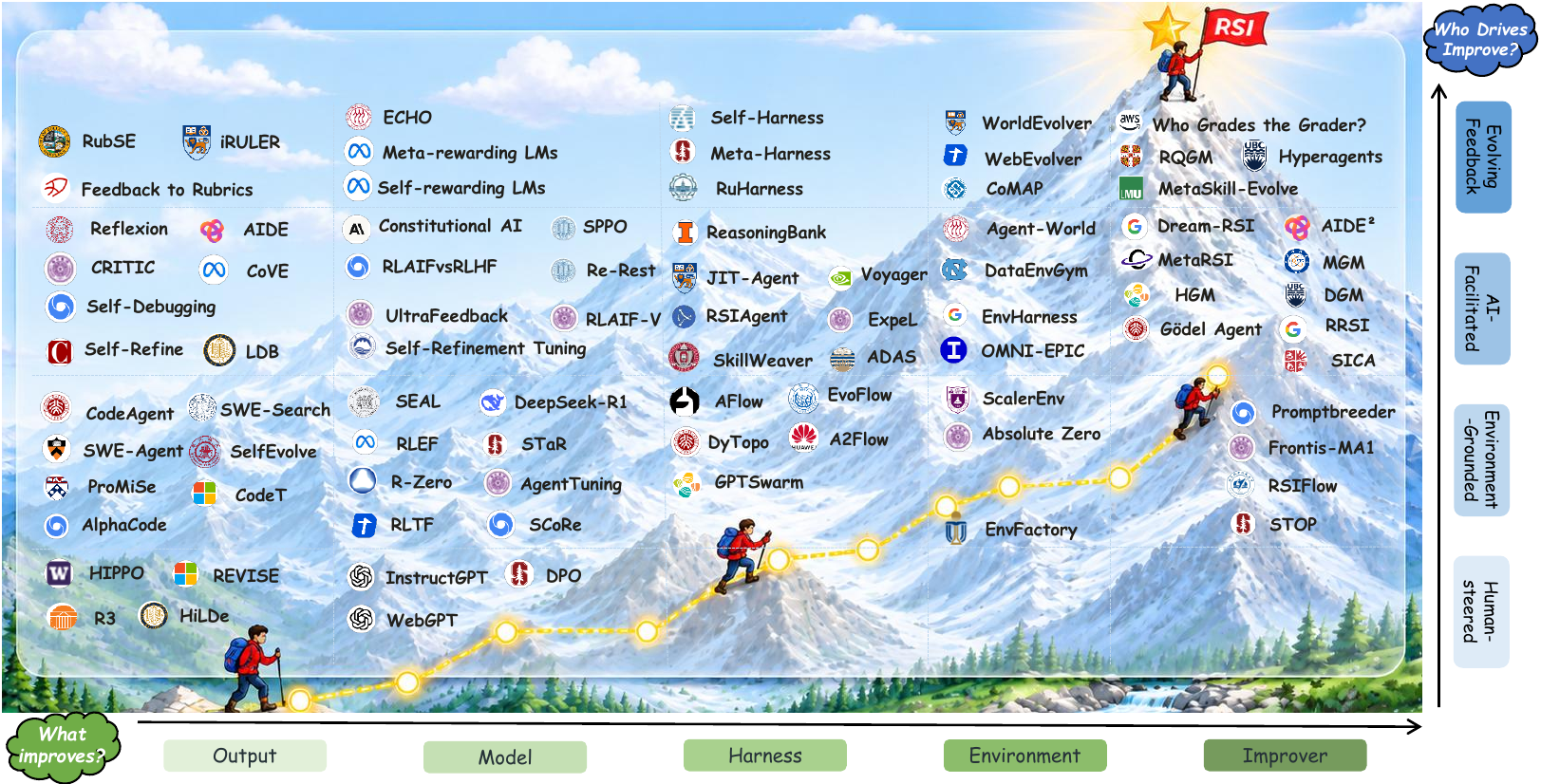}
    \caption{\textbf{Two-Axis Taxonomy Towards Recursive Self-Improvement.} What Improves and Who Drives Improve jointly characterize the scope and mechanisms of agent improvement, providing a unified view of how expanding improvement targets and increasingly autonomous feedback support progress toward RSI.}
    \label{fig:mountain}
\end{figure*}

Motivated by these observations, this survey systematically examines the technological roadmaps from task-local agent optimization toward \textbf{Reliable Recursive Self-Improvement}. We introduce a two-axis analytical taxonomy comprising What Improves and Who Drives Improve, which characterize the components subject to modification and the sources and mechanisms of feedback driving improvement, respectively, as illustrated in Fig.~\ref{fig:mountain}. Building on this taxonomy, we organize existing research into seven interrelated milestones according to the expansion of improvement capabilities and their underlying mechanistic requirements, as illustrated in Fig.~\ref{fig:road}. These milestones cover output refinement, foundation model training, agent harness evolution, environment synthesis, feedback automation, agent-environment co-evolution, and recursive self-improvement. Furthermore, we examine the construction, self-modification, cross-generational reuse, and capability evaluation of improvement mechanisms, and systematically review relevant benchmarks, applications, and open challenges. Through these analyses, this survey provides a systematic perspective and research roadmap for understanding existing advances in agent improvement and exploring reliable, sustainable, and verifiable RSI.

The main contributions of this survey are summarized as follows:

\begin{itemize}
    \item We systematically analyze existing improvement approaches along two complementary dimensions, What Improves and Who Drives Improve, characterizing the roles and interrelationships of different improvement mechanisms and clarifying the capability boundaries among task-local optimization, persistent system improvement, and recursive self-improvement.

    \item We organize existing research into seven interrelated milestones, systematically examining the improvement capabilities and limitations of different approaches and elucidating the key connections between the accumulation of agent capabilities and the recursive enhancement of improvement mechanisms.

    \item We further investigate the cross-generational inheritance and capability evaluation of improvement mechanisms, summarize representative benchmarks, practical applications, and open challenges, and discuss research directions toward reliable, sustainable, and verifiable RSI.
\end{itemize}

%% file: section/paper_list.tex
\definecolor{hidden-draw}{RGB}{255,255,255}
\newcommand{\taxwrap}[1]{\parbox[t]{50.7em}{\raggedright #1\par}}
\newcommand{\taxwrapwide}[1]{\parbox[t]{61.4em}{\raggedright #1\par}}
\tikzset{
    my-box/.style={
        rectangle,
        draw=hidden-draw,
        rounded corners,
        align=left,
        text opacity=1,
        minimum height=1.5em,
        minimum width=5em,
        inner sep=2pt,
        fill opacity=.8,
        line width=0.8pt,
    },
    leaf-head/.style={
        my-box,
        minimum height=1.5em,
        draw=gray!80,
        fill=gray!15,
        text=black,
        font=\normalsize,
        inner xsep=2pt,
        inner ysep=4pt,
        line width=0.8pt,
    },
    leaf-datasets/.style={
        my-box,
        minimum height=1.5em,
        draw=orange!80,
        fill=orange!15,
        text=black,
        font=\normalsize,
        inner xsep=2pt,
        inner ysep=4pt,
        line width=0.8pt,
    },
    leaf-methods/.style={
        my-box,
        minimum height=1.5em,
        draw=red!70,
        fill=red!15,
        text=black,
        font=\normalsize,
        inner xsep=2pt,
        inner ysep=4pt,
        line width=0.8pt,
    },
    leaf-metrics/.style={
        my-box,
        minimum height=1.5em,
        draw=cyan!70,
        fill=cyan!15,
        text=black,
        font=\normalsize,
        inner xsep=2pt,
        inner ysep=4pt,
        line width=0.8pt,
    },
    modelnode-datasets/.style={
        my-box,
        minimum height=1.5em,
        draw=orange!80,
        fill=white,
        text=black,
        font=\normalsize,
        inner xsep=2pt,
        inner ysep=4pt,
        line width=0.8pt,
    },
    modelnode-methods/.style={
        my-box,
        minimum height=1.5em,
        draw=red!70,
        fill=white,
        text=black,
        font=\normalsize,
        inner xsep=2pt,
        inner ysep=4pt,
        line width=0.8pt,
    },
    modelnode-metrics/.style={
        my-box,
        minimum height=1.5em,
        draw=cyan!70,
        fill=white,
        text=black,
        font=\normalsize,
        inner xsep=2pt,
        inner ysep=4pt,
        line width=0.8pt,
    }
}
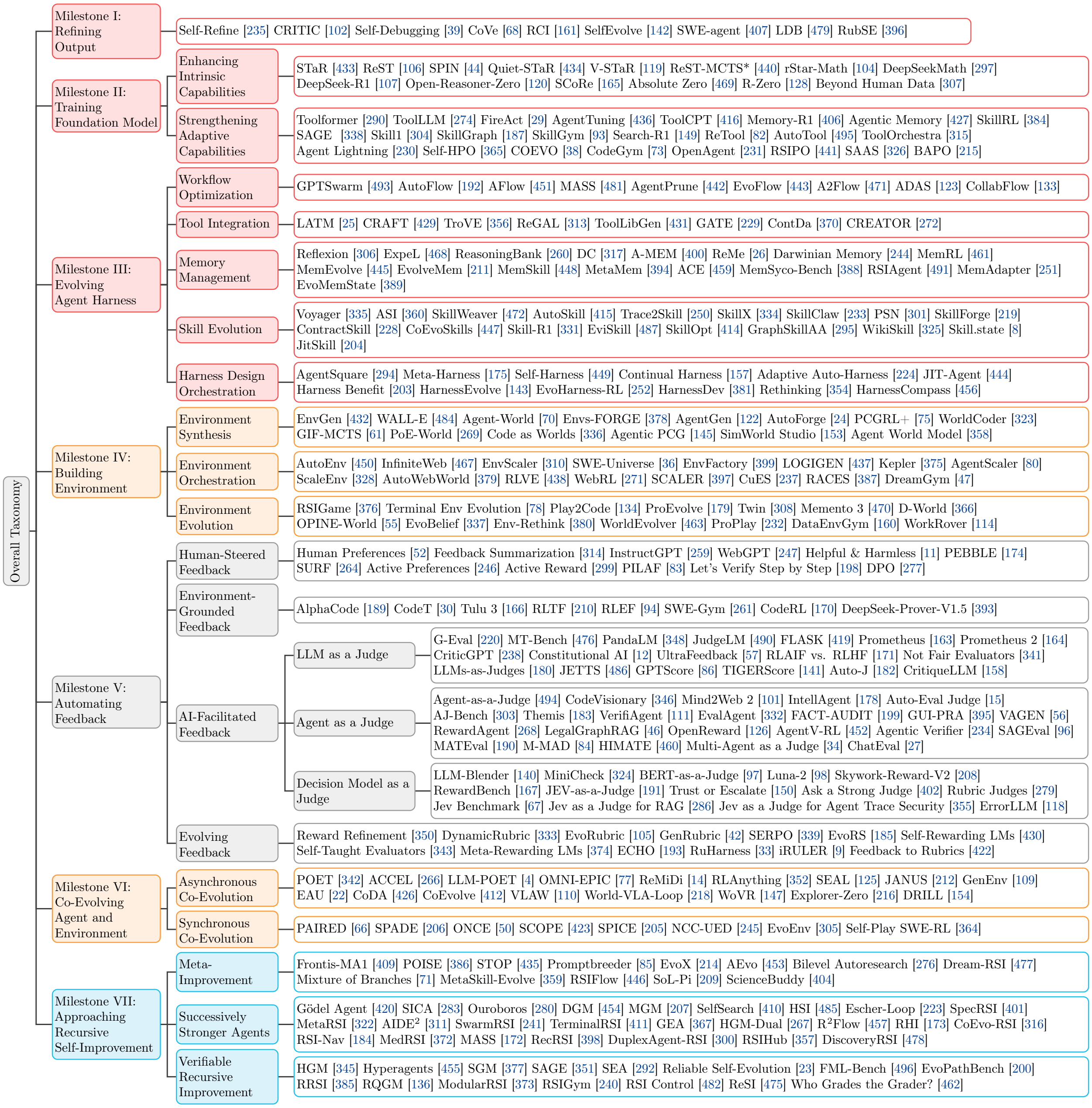
\begin{figure*}[!t]
    \centering
    \vspace{-5mm}
    \resizebox{1\textwidth}{!}{
        \begin{forest}
            forked edges,
            for tree={
                grow=east,
                reversed=true,
                anchor=base west,
                parent anchor=east,
                child anchor=west,
                base=left,
                font=\normalsize,
                rectangle,
                draw=hidden-draw,
                rounded corners,
                align=left,
                minimum width=1em,
                edge+={darkgray, line width=1pt},
                s sep=3pt,
                inner xsep=0pt,
                inner ysep=3pt,
                line width=0.8pt,
                ver/.style={
                    rotate=90,
                    child anchor=north,
                    parent anchor=south,
                    anchor=center
                },
            },
            [
                 Overall Taxonomy, leaf-head, ver
                [
                    {\parbox[t]{8em}{\raggedright Milestone I:\\ Refining\\Output}},
                    leaf-methods
                    [
                        {\taxwrapwide{
                        Self-Refine~\cite{SelfRefine}
                        CRITIC~\cite{CRITIC}
                        Self-Debugging~\cite{SelfDebugging}
                        CoVe~\cite{dhuliawala-etal-2024-chain}
                        RCI~\cite{NEURIPS2023_7cc1005e}
                        SelfEvolve~\cite{jiang2023selfevolvecodeevolutionframework}
                        SWE-agent~\cite{yang2024sweagent}
                        LDB~\cite{zhong-etal-2024-debug}
                        RubSE~\cite{xiong2026rubricsvisualrepaircontextselfevolving}
                        }},
                        modelnode-methods,
                        text width=61.4em
                    ]
                ]
                [
                    {\parbox[t]{8em}{\raggedright Milestone II:\\Training\\Foundation Model}},
                    leaf-methods
                    [
                        {\parbox[t]{7.5em}{\raggedright Enhancing Intrinsic Capabilities}},
                        leaf-methods
                        [
                            {\taxwrapwide{
                            STaR~\cite{STaR}
                            ReST~\cite{gulcehre2023rest}
                            SPIN~\cite{chen2024spin}
                            Quiet-STaR~\cite{zelikman2024quietstar}
                            V-STaR~\cite{hosseini2024vstar}
                            ReST-MCTS*~\cite{zhang2024restmcts}
                            rStar-Math~\cite{guan2025rstarmath}
                            DeepSeekMath~\cite{shao2024deepseekmath}
                            DeepSeek-R1~\cite{DeepSeekR1}
                            Open-Reasoner-Zero~\cite{hu2025openreasonerzero}
                            SCoRe~\cite{SCoRe}
                            Absolute Zero~\cite{zhao2025absolutezero}
                            R-Zero~\cite{huang2026rzero}
                            Beyond Human Data~\cite{ReSTEM}
                            }},
                            modelnode-methods,
                            text width=61.4em
                        ]
                    ]
                    [
                        {\parbox[t]{7.5em}{\raggedright Strengthening Adaptive Capabilities}},
                        leaf-methods
                        [
                            {\taxwrapwide{
                            Toolformer~\cite{schick2023toolformer}
                            ToolLLM~\cite{qin2024toolllm}
                            FireAct~\cite{chen2023fireact}
                            AgentTuning~\cite{AgentTuning}
                            ToolCPT~\cite{yang2026toolcpt}
                            Memory-R1~\cite{yan2026memoryr1}
                            Agentic Memory~\cite{yu2026agemem}
                            SkillRL~\cite{xia2026skillrl}
                            SAGE ~\cite{wang2026sage}
                            Skill1~\cite{shi2026skill1}
                            SkillGraph~\cite{li2026skillgraph}
                            SkillGym~\cite{ge2026skillgym}
                            Search-R1~\cite{SearchR1}
                            ReTool~\cite{feng2025retool}
                            AutoTool~\cite{zou2026autotool}
                            ToolOrchestra~\cite{su2026toolorchestra}
                            \\
                            Agent Lightning~\cite{luo2025agentlightning}
                            Self-HPO~\cite{wei2026selfevolvingcodingagentsharnesspolicy}
                            COEVO~\cite{chen2026coevocoevolvingcontextparameters}
                            CodeGym~\cite{du2026generalizableendtoendtooluserl}
                            OpenAgent~\cite{lv2026agentsgeneralizeopenworld}
                            RSIPO~\cite{zhang2026rsiporecursiveselfimprovingpolicy}
                            SAAS~\cite{tang2026saasselfawarereinforcementlearning}
                            BAPO~\cite{liu2026bapoboundaryawarepolicyoptimization}
                            }},
                            modelnode-methods,
                            text width=61.4em
                        ]
                    ]
                ]
                [
                    {\parbox[t]{8em}{\raggedright Milestone III:\\Evolving\\Agent Harness}},
                    leaf-methods
                    [
                        {\parbox[t]{7.5em}{\raggedright 
                        Workflow Optimization}},
                        leaf-methods
                        [
                            {\taxwrapwide{
                            GPTSwarm~\cite{zhuge2024gptswarm}
                            AutoFlow~\cite{li2024autoflow}
                            AFlow~\cite{zhang2024aflow}
                            MASS~\cite{zhou2026mass}
                            AgentPrune~\cite{zhang2024agentprune}
                            EvoFlow~\cite{zhang2025evoflow}
                            A2Flow~\cite{zhao2026a2flow}
                            ADAS~\cite{hu2024adas}
                            CollabFlow~\cite{huang2026collabflowrecursiveselfimprovementagent}
                            }},
                            modelnode-methods,
                            text width=61.4em
                        ]
                    ]
                    [
                        {\parbox[t]{7.5em}{\raggedright Tool Integration}},
                        leaf-methods
                        [
                            {\taxwrapwide{
                            LATM~\cite{cai2024large}
                            CRAFT~\cite{yuan2024craft}
                            TroVE~\cite{wang2024trove}
                            ReGAL~\cite{stengeleskin2024regal}
                            ToolLibGen~\cite{yue2025toollibgen}
                            GATE~\cite{luo2026gate}
                            ContDa~\cite{wu2026contda}
                            CREATOR~\cite{qian-etal-2023-creator}
                            }},
                            modelnode-methods,
                            text width=61.4em
                        ]
                    ]
                    [
                        {\parbox[t]{7.5em}{\raggedright Memory Management}},
                        leaf-methods
                        [
                            {\taxwrapwide{
                            Reflexion~\cite{shinn2023reflexion}
                            ExpeL~\cite{zhao2024expel}
                            ReasoningBank~\cite{ouyang2026reasoningbank}
                            DC~\cite{suzgun2026dynamiccheatsheet}
                            A-MEM~\cite{xu2025amem}
                            ReMe~\cite{cao2026reme}
                            Darwinian Memory~\cite{mi2026darwinianmemory}
                            MemRL~\cite{zhang2026memrl}
                            MemEvolve~\cite{zhang2026memevolve}
                            EvolveMem~\cite{liu2026evolvemem}
                            MemSkill~\cite{zhang2026memskill}
                            MetaMem~\cite{xin2026metamemevolvingmetamemoryknowledge}
                            ACE~\cite{zhang2026agenticcontextengineeringevolving}
                            MemSyco-Bench~\cite{xiang2026memsycobenchbenchmarkingsycophancyagent}
                            RSIAgent~\cite{zhu2026rsiagentautonomousexplorationrecursive}
                            MemAdapter~\cite{ning2026memadaptercounterfactualadaptationmemoryinduced}
                            EvoMemState~\cite{xiang2026evomemstateevolvingmemorystate}
                            }},
                            modelnode-methods,
                            text width=61.4em
                        ]
                    ]
                    [
                        {\parbox[t]{7.5em}{\raggedright Skill Evolution}},
                        leaf-methods
                        [
                            {\taxwrapwide{
                            Voyager~\cite{wang2023voyager}
                            ASI~\cite{wang2025asi}
                            SkillWeaver~\cite{zheng2025skillweaver}
                            AutoSkill~\cite{yang2026autoskill}
                            Trace2Skill~\cite{ni2026trace2skill}
                            SkillX~\cite{wang2026skillx}
                            SkillClaw~\cite{ma2026skillclaw}
                            PSN~\cite{shi2026psn}
                            SkillForge~\cite{liu2026skillforge}
                            ContractSkill~\cite{lu2026contractskill}
                            CoEvoSkills~\cite{zhang2026coevoskills}
                            Skill-R1~\cite{vishe2026skillr1}
                            EviSkill~\cite{zhou2026eviskillgroundingskillevolution}
                            SkillOpt~\cite{yang2026skilloptexecutivestrategyselfevolving}
                            GraphSkillAA~\cite{shang2026graphskillaaattributionguidedskillgraphupdating}
                            WikiSkill~\cite{tang2026wikiskillcompilingagentexperience}
                            Skill.state~\cite{badhe2026skillstatescalablelonghorizonagent}
                            JitSkill~\cite{lin2026justintimeskillretrievallonghorizon}
                            }},
                            modelnode-methods,
                            text width=61.4em
                        ]
                    ]
                    [
                        {\parbox[t]{7.5em}{\raggedright Harness Design Orchestration}},
                        leaf-methods
                        [
                            {\taxwrapwide{
                            AgentSquare~\cite{shang2025agentsquare}
                            Meta-Harness~\cite{lee2026metaharness}
                            Self-Harness~\cite{zhang2026selfharness}
                            Continual Harness~\cite{karten2026continualharness}
                            Adaptive Auto-Harness~\cite{liu2026adaptiveautoharness}
                            JIT-Agent~\cite{zhang2026jitagent}
                            \\
                            Harness Benefit~\cite{lin2026harnessbenefit}
                            HarnessEvolve~\cite{jiang2026harnessevolvelearningreferencetrajectories}
                            EvoHarness-RL~\cite{ning2026evoharnessrllearningruntimeharness}
                            HarnessDev~\cite{wu2026harnessdev}
                            Rethinking~\cite{wang2026rethinking}
                            HarnessCompass~\cite{zhang2026harnesscompassguidingautomaticharness}
                            }},
                            modelnode-methods,
                            text width=61.4em
                        ]
                    ]
                ]
                [
                    {\parbox[t]{8em}{\raggedright Milestone IV:\\Building\\Environment}},
                    leaf-datasets
                    [
                        {\parbox[t]{7.5em}{\raggedright Environment Synthesis}},
                        leaf-datasets
                        [
                            {\taxwrapwide{
                            EnvGen~\cite{envgen}
                            WALL-E~\cite{walle2}
                            Agent-World~\cite{agentworld}
                            Envs-FORGE~\cite{envsforge}
                            AgentGen~\cite{agentgen}
                            AutoForge~\cite{cai-etal-2026-autoforge}
                            PCGRL+~\cite{earle2024pcgrlscalingcontrolgeneralization}
                            WorldCoder~\cite{worldcoder}
                            GIF-MCTS~\cite{gifmcts}
                            PoE-World~\cite{poeworld}
                            Code as Worlds~\cite{codeasworlds}
                            Agentic PCG~\cite{jiang2026agentic}
                            SimWorld Studio~\cite{kang2026simworldstudioautomaticenvironment}
                            Agent World Model~\cite{pmlr-v306-wang26jh}
                            }},
                            modelnode-datasets,
                            text width=61.4em
                        ]
                    ]
                    [
                        {\parbox[t]{7.5em}{\raggedright Environment Orchestration}},
                        leaf-datasets
                        [
                            {\taxwrapwide{
                            AutoEnv~\cite{autoenv}
                            InfiniteWeb~\cite{zhang-etal-2026-infiniteweb}
                            EnvScaler~\cite{song-etal-2026-envscaler}
                            SWE-Universe~\cite{chen2026sweuniversescalerealworldverifiable}
                            EnvFactory~\cite{xu2026envfactoryscalingtooluseagents}
                            LOGIGEN~\cite{zeng2026logigenlogicdrivengenerationverifiable}
                            Kepler~\cite{kepler}
                            AgentScaler~\cite{fang-etal-2026-towards}
                            ScaleEnv~\cite{pmlr-v306-tu26a}
                            AutoWebWorld~\cite{pmlr-v306-wu26ar}
                            RLVE~\cite{rlve}
                            WebRL~\cite{webrl}
                            SCALER~\cite{scaler}
                            CuES~\cite{cues}
                            RACES~\cite{xiang2026verifiableenvironmentslegobricks}
                            DreamGym~\cite{dreamgym}
                            }},
                            modelnode-datasets,
                            text width=61.4em
                        ]
                    ]
                    [
                        {\parbox[t]{7.5em}{\raggedright Environment Evolution}},
                        leaf-datasets
                        [
                            {\taxwrapwide{
                            RSIGame~\cite{wu2026rsigameautonomousagenticgame}
                            Terminal Env Evolution~\cite{terminalenv_evolution}
                            Play2Code~\cite{huang-etal-2026-gui-agents}
                            ProEvolve~\cite{proevolve}
                            Twin~\cite{twin}
                            Memento 3~\cite{memento3}
                            D-World~\cite{dworld}
                            OPINE-World~\cite{opineworld}
                            EvoBelief~\cite{evobelief}
                            Env-Rethink~\cite{envrethink}
                            WorldEvolver~\cite{worldevolver}
                            ProPlay~\cite{proplay}
                            DataEnvGym~\cite{khan2025dataenvgymdatagenerationagents}
                            WorkRover~\cite{he2026workroverscalinglonghorizoncowork}
                            }},
                            modelnode-datasets,
                            text width=61.4em
                        ]
                    ]
                ]
                [
                    {\parbox[t]{8em}{\raggedright Milestone V:\\Automating\\Feedback}},
                    leaf-head
                    [
                        {\parbox[t]{7.5em}{\raggedright Human-Steered Feedback}},
                        leaf-head
                        [
                            {\taxwrapwide{
                            Human Preferences~\cite{human_preferences}
                            Feedback Summarization~\cite{summarize}
                            InstructGPT~\cite{instructgpt}
                            WebGPT~\cite{webgpt}
                            Helpful \& Harmless~\cite{helpful_harmless}
                            PEBBLE~\cite{pebble}
                            SURF~\cite{surf}
                            Active Preferences~\cite{active_preference}
                            Active Reward~\cite{active_reward}
                            PILAF~\cite{pilaf}
                            Let's Verify Step by Step~\cite{lightman2023letsverifystepstep}
                            DPO~\cite{NEURIPS2023_a85b405e}
                            }},
                            modelnode-datasets,
                            draw=gray!80,
                            text width=61.4em
                        ]
                    ]
                    [
                        {\parbox[t]{7.5em}{\raggedright Environment-Grounded Feedback}},
                        leaf-head
                        [
                            {\taxwrapwide{
                            AlphaCode~\cite{alphacode}
                            CodeT~\cite{codet}
                            Tulu 3~\cite{tulu3}
                            RLTF~\cite{rltf}
                            RLEF~\cite{rlef}
                            SWE-Gym~\cite{swe_gym}
                            CodeRL~\cite{NEURIPS2022_8636419d}
                            DeepSeek-Prover-V1.5~\cite{ICLR2025_b3b55c36}
                            }},
                            modelnode-datasets,
                            draw=gray!80,
                            text width=61.4em
                        ]
                    ]
                    [
                        {\parbox[t]{7.5em}{\raggedright AI-Facilitated Feedback}},
                        leaf-head
                        [
                            {\parbox[t]{9em}{\raggedright LLM as a Judge}},
                            leaf-head
                            [
                                {\taxwrap{
                                G-Eval~\cite{geval}
                                MT-Bench~\cite{mtbench}
                                PandaLM~\cite{pandalm}
                                JudgeLM~\cite{judgelm}
                                FLASK~\cite{flask}
                                Prometheus~\cite{prometheus}
                                Prometheus 2~\cite{prometheus2}
                                CriticGPT~\cite{criticgpt}
                                Constitutional AI~\cite{constitutional}
                                UltraFeedback~\cite{ultrafeedback}
                                RLAIF vs. RLHF~\cite{rlaif}
                                Not Fair Evaluators~\cite{fair_evaluators}
                                LLMs-as-Judges~\cite{judge_survey}
                                JETTS~\cite{jetts}
                                GPTScore~\cite{fu-etal-2024-gptscore}
                                TIGERScore~\cite{jiang2024tigerscorebuildingexplainablemetric}
                                Auto-J~\cite{li2023generativejudgeevaluatingalignment}
                                CritiqueLLM~\cite{ke2024critiquellminformativecritiquegeneration}
                                }},
                                modelnode-datasets,
                                draw=gray!80,
                                text width=50.7em
                            ]
                        ]
                        [
                            {\parbox[t]{9em}{\raggedright Agent as a Judge}},
                            leaf-head
                            [
                                {\taxwrap{
                                Agent-as-a-Judge~\cite{agent_judge}
                                CodeVisionary~\cite{codevisionary}
                                Mind2Web 2~\cite{mind2web2}
                                IntellAgent~\cite{intellagent}
                                Auto-Eval Judge~\cite{auto_eval}
                                AJ-Bench~\cite{ajbench}
                                Themis~\cite{themis}
                                VerifiAgent~\cite{verifiagent}
                                EvalAgent~\cite{evalagent}
                                FACT-AUDIT~\cite{fact_audit}
                                GUI-PRA~\cite{gui_pra}
                                VAGEN~\cite{vagen}
                                RewardAgent~\cite{rewardagent}
                                LegalGraphRAG~\cite{chen-etal-2026-legalgraphrag}
                                OpenReward~\cite{openreward}
                                AgentV-RL~\cite{agentv_rl}
                                Agentic Verifier~\cite{scaling_verifier}
                                SAGEval~\cite{ghosh2024sagevalfrontierssatisfactoryagent}
                                MATEval~\cite{li2024matevalmultiagentdiscussionframework}
                                M-MAD~\cite{feng-etal-2025-mad}
                                HIMATE~\cite{zhang-etal-2025-himate}
                                Multi-Agent as a Judge~\cite{chen-etal-2026-multi-agent}
                                ChatEval~\cite{chan2023chatevalbetterllmbasedevaluators}
                                }},
                                modelnode-datasets,
                                draw=gray!80,
                                text width=50.7em
                            ]
                        ]
                        [
                            {\parbox[t]{9em}{\raggedright Decision Model as a Judge}},
                            leaf-head
                            [
                                {\taxwrap{
                                LLM-Blender~\cite{pairranker}
                                MiniCheck~\cite{minicheck}
                                BERT-as-a-Judge~\cite{bert_judge}
                                Luna-2~\cite{luna2}
                                Skywork-Reward-V2~\cite{skywork}
                                RewardBench~\cite{rewardbench}
                                JEV-as-a-Judge~\cite{jev}
                                Trust or Escalate~\cite{trust_escalate}
                                Ask a Strong Judge~\cite{ask_strong}
                                Rubric Judges~\cite{jev_rubric}
                                \\
                                Jev Benchmark~\cite{jev_benchmark}
                                Jev as a Judge for RAG~\cite{rosario2026jevasajudgeragclaimverification}
                                Jev as a Judge for Agent Trace Security~\cite{wang2026jevjudgeagenttrace}
                                ErrorLLM~\cite{hong2026errorllmmodelingsqlerrors}
                                }},
                                modelnode-datasets,
                                draw=gray!80,
                                text width=50.7em
                            ]
                        ]
                    ]
                    [
                        {\parbox[t]{7.5em}{\raggedright Evolving Feedback}},
                        leaf-head
                        [
                            {\taxwrapwide{
                            Reward Refinement~\cite{reward_refinement}
                            DynamicRubric~\cite{dynamic_rubric}
                            EvoRubric~\cite{evo_rubric}
                            GenRubric~\cite{genrubric}
                            SERPO~\cite{serpo}
                            EvoRS~\cite{evors}
                            Self-Rewarding LMs~\cite{self_rewarding}
                            Self-Taught Evaluators~\cite{self_taught}
                            Meta-Rewarding LMs~\cite{meta_rewarding}
                            ECHO~\cite{echo}
                            RuHarness~\cite{chen2026criticfallsbehindcoevolving}
                            iRULER~\cite{bai2026irulerintelligiblerubricbaseduserdefined}
                            Feedback to Rubrics~\cite{yoshida2026feedbacktorubricslearnexpertcriteria}
                            }},
                            modelnode-datasets,
                            draw=gray!80,
                            text width=61.4em
                        ]
                    ]
                ]
                [
                    {\parbox[t]{8em}{\raggedright Milestone VI:\\Co-Evolving\\Agent and Environment}},
                    leaf-datasets
                    [
                        {\parbox[t]{7.5em}{\raggedright Asynchronous Co-Evolution}},
                        leaf-datasets
                        [
                            {\taxwrapwide{
                            POET~\cite{poet}
                            ACCEL~\cite{accel}
                            LLM-POET~\cite{llmpoet}
                            OMNI-EPIC~\cite{omniepic}
                            ReMiDi~\cite{pmlr-v235-beukman24a}
                            RLAnything~\cite{wang2026rlanythingforgeenvironmentpolicy}
                            SEAL~\cite{seal}
                            JANUS~\cite{janus}
                            GenEnv~\cite{guo2025genenv}
                            EAU~\cite{cai2026eauenvironmentagentusercoevolutiontraining}
                            CoDA~\cite{yu2026codacoevolutiondataengines}
                            CoEvolve~\cite{yang-etal-2026-coevolve}
                            VLAW~\cite{vlaw}
                            World-VLA-Loop~\cite{worldvlaloop}
                            WoVR~\cite{jiang2026wovrworldmodelsreliable}
                            Explorer-Zero~\cite{liu2026explorerzeroimprovingselfplaylonghorizon}
                            DRILL~\cite{kang2026drilltrainingworldmodels}
                            }},
                            modelnode-datasets,
                            text width=61.4em
                        ]
                    ]
                    [
                        {\parbox[t]{7.5em}{\raggedright Synchronous Co-Evolution}},
                        leaf-datasets
                        [
                            {\taxwrapwide{
                            PAIRED~\cite{paired}
                            SPADE~\cite{spade}
                            ONCE~\cite{chi-etal-2022-offline}
                            SCOPE~\cite{yu2026frozensimulatorenoughsimulator}
                            SPICE~\cite{liu2025spiceselfplaycorpusenvironments}
                            NCC-UED~\cite{monette2025optimisation}
                            EvoEnv~\cite{shi2026learningbuildenvironmentselfevolving}
                            Self-Play SWE-RL~\cite{pmlr-v306-wei26x}
                            }},
                            modelnode-datasets,
                            text width=61.4em
                        ]
                    ]
                ]
                [
                    {\parbox[t]{8em}{\raggedright Milestone VII:\\Approaching\\Recursive Self-Improvement}},
                    leaf-metrics
                    [
                        {\parbox[t]{7.5em}{\raggedright Meta-Improvement}},
                        leaf-metrics
                        [
                            {\taxwrapwide{
                            Frontis-MA1~\cite{yang2026frontisma1}
                            POISE~\cite{xia2026poise}
                            STOP~\cite{zelikman2024stop}
                            Promptbreeder~\cite{fernando2024promptbreeder}
                            EvoX~\cite{liu2026evox}
                            AEvo~\cite{zhang2026aevo}
                            Bilevel Autoresearch~\cite{qu2026bilevel}
                            Dream-RSI~\cite{zheng2026dreamrsi}
                            Mixture of Branches~\cite{dong2026mixture}
                            MetaSkill-Evolve~\cite{wang2026metaskillevolve}
                            RSIFlow~\cite{zhang2026rsiflowadaptiveselfimprovementworkflows}
                            SoL-Pi~\cite{liu2026solpirecursivelyscalingautoresearch}
                            ScienceBuddy~\cite{xue2026sciencebuddyrecursiveinrecursiveselfimprovementinteractive}
                            }},
                            modelnode-metrics,
                            text width=61.4em
                        ]
                    ]
                    [
                        {\parbox[t]{7.5em}{\raggedright Successively Stronger Agents}},
                        leaf-metrics
                        [
                            {\taxwrapwide{
                            G{\"o}del Agent~\cite{yin2025goedelagent}
                            SICA~\cite{robeyns2025sica}
                            Ouroboros~\cite{razzhigaev2026ouroboros}
                            DGM~\cite{zhang2026darwin}
                            MGM~\cite{liu2026mendel}
                            SelfSearch~\cite{yang2026selfsearch}
                            HSI~\cite{zhou2026hsi}
                            Escher-Loop~\cite{liu2026escher}
                            SpecRSI~\cite{xu2026specrsi}
                            MetaRSI~\cite{tan2026metarsi}
                            AIDE$^2$~\cite{srikanth2026aide2}
                            SwarmRSI~\cite{meng2026swarmrsirecursiveselfimprovementevolving}
                            TerminalRSI~\cite{yang2026terminalrsirecursiveselfimprovingterminal}
                            GEA~\cite{weng2026groupevolvingagentsopenendedselfimprovement}
                            HGM-Dual~\cite{paul2026hgmduallineageawareguidedagent}
                            R$^2$Flow~\cite{zhang2026r2flowrecursiveselfimprovement}
                            RHI~\cite{lee2026recursiveharnessselfimprovement}
                            CoEvo-RSI~\cite{su2026coevorsiharnessesevolvemodels}
                            RSI-Nav~\cite{li2026datacentricrecursiveselfimprovementnavigation}
                            MedRSI~\cite{wu2026medrsirecursiveselfimprovementmedical}
                            MASS~\cite{lee2026recursiveselfimprovementmultiagentselfsupervision}
                            RecRSI~\cite{xu2026successoraccuracystateretention}
                            DuplexAgent-RSI~\cite{shen2026duplexagentrsirecursiveharnessimprovement}
                            RSIHub~\cite{wang2026rsihubcomposableframeworkagentevolution}
                            DiscoveryRSI~\cite{zhong2026discoveryrsirecursivelylearningdiscover}
                            }},
                            modelnode-metrics,
                            text width=61.4em
                        ]
                    ]
                    [
                        {\parbox[t]{7.5em}{\raggedright Verifiable Recursive Improvement}},
                        leaf-metrics
                        [
                            {\taxwrapwide{
                            HGM~\cite{wang2025huxley}
                            Hyperagents~\cite{zhang2026hyperagents}
                            SGM~\cite{wu2025sgm}
                            SAGE~\cite{sage2026}
                            SEA~\cite{sea2026}
                            Reliable Self-Evolution~\cite{cai2026reliable}
                            FML-Bench~\cite{zou2026fmlbench}
                            EvoPathBench~\cite{evopathbench2026}
                            RRSI~\cite{xia2026rrsi}
                            RQGM~\cite{iacob2026redqueen}
                            ModularRSI~\cite{wu2026modularrsi}
                            RSIGym~\cite{meng2026rsigymflexibleenvironmentrecursive}
                            RSI Control~\cite{zhou2026rsicontrolaimodels}
                            ReSI~\cite{zheng2026resirecursivesafetyimprovement}
                            Who Grades the Grader?~\cite{zhang2026whogrades}
                            }},
                            modelnode-metrics,
                            text width=61.4em
                        ]
                    ]
                ]
            ]
        \end{forest}
    }
    \vspace{-5mm}
    \caption{ \textbf{A Taxonomy of Representative Studies Across Seven Milestones.}}
    \label{fig:taxonomy}
    \vspace{-5mm}
\end{figure*}

%% file: section/overview.tex
\section{Preliminaries}

\begin{figure*}[t]
    \centering
    \includegraphics[width=\textwidth]{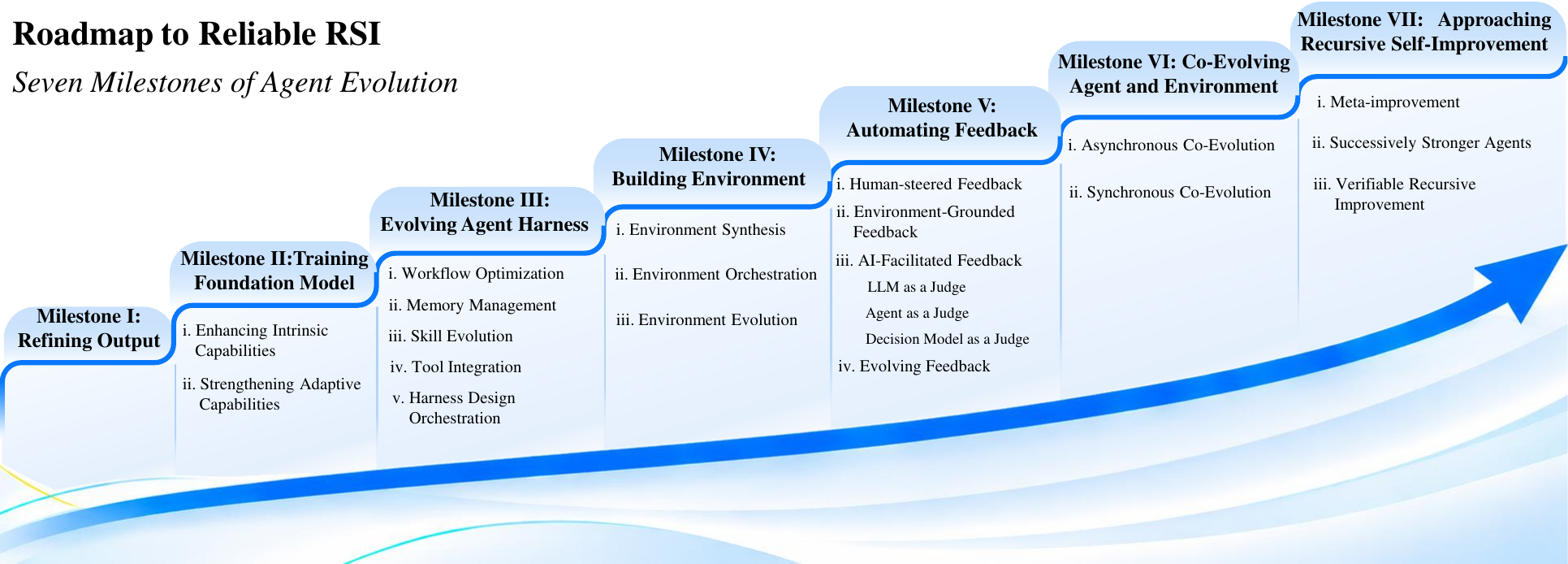}
    \caption{\textbf{Roadmap Toward Recursive Self-Improvement.} Seven interrelated milestones trace the expansion of improvement capabilities from task-local refinement to persistent system evolution and recursive improvement, highlighting the capabilities established by different research directions and their interrelationships.}
    \label{fig:road}
\end{figure*}

We clarify the core concepts and scope of agent
improvement, and then introduce two complementary views,
What Improves and Who Drives Improve, to establish the conceptual
foundation for the roadmap toward recursive self-improvement
(RSI).

\subsection{Definitions and Scope}

An Agent is an interactive computational system that perceives
its environment, makes decisions, and performs tasks through
iterative execution. It primarily consists of a Foundation Model
and an Agent Harness, operating through sustained interaction
with its environment. At system version $k$, an Agent can be
abstracted as
\begin{equation}
    A^{(k)} = (M^{(k)}, H^{(k)}; E^{(k)}),
    \label{eq:agent}
\end{equation}
where the following core components are considered:
\begin{itemize}
    \item \textbf{Foundation Model $M^{(k)}$:}
    The cognitive core of the agent, providing reasoning,
    planning, generation, and decision-making capabilities.

    \item \textbf{Agent Harness $H^{(k)}$:}
    The execution structure that organizes model invocations
    and external resources, including workflows, memory,
    skills, tools, and runtime control mechanisms.

    \item \textbf{Environment $E^{(k)}$:}
    The task and interaction setting in which the agent operates, encompassing task instances and distributions, learning curricula, environmental states, interaction rules, operational conditions, and environment models.
\end{itemize}

The Model and Harness constitute the internal agent system,
whereas the Environment represents its external interaction
setting. Agent-environment interactions generate task outcomes,
execution experience, and feedback that inform subsequent
improvement.

Building on this abstraction, we define Agent Improvement as
the process of modifying system-related targets based on
execution experience and feedback. We refer to the mechanism
responsible for proposing, evaluating, selecting, and applying
candidate modifications as the Improver. The improvement
process can be abstracted as
\begin{equation}
    X^{(k+1)} =
    \mathcal{I}^{(k)}(X^{(k)}; \Phi^{(k)}),
    \label{eq:agent_improvement}
\end{equation}
where:
\begin{itemize}
    \item \textbf{Evolution Target $X^{(k)}$:}
    The object being modified at system version $k$, including task-local Outputs and persistent system components: the Foundation Model, Agent Harness, Environment, and Improver. Model Evolution concerns changes to Foundation Model parameters and their internal capabilities, regardless of the semantic type of the capability being learned. Harness Evolution concerns persistent modifications to external resources and execution structures, while Environment Evolution concerns changes to tasks, task distributions, environmental states, interaction rules, and environment models.

    \item \textbf{Feedback $\Phi^{(k)}$:}
    The information used to evaluate outcomes, identify
    deficiencies, and guide subsequent modifications, obtained
    through human guidance, direct environmental signals, 
    AI-facilitated judgments, or evolving feedback.

    \item \textbf{Improver $\mathcal{I}^{(k)}$:}
    The mechanism that uses experience and feedback to propose,
    evaluate, select, and apply candidate modifications.

    \item \textbf{Updated Target $X^{(k+1)}$:}
    The resulting target after modification, which may affect the current execution for task-local output refinement or be retained across executions for persistent system improvement.
\end{itemize}

\begin{figure*}[!t]
    \centering
    \includegraphics[
        width=\textwidth,
        height=0.55\textheight,
        keepaspectratio
    ]{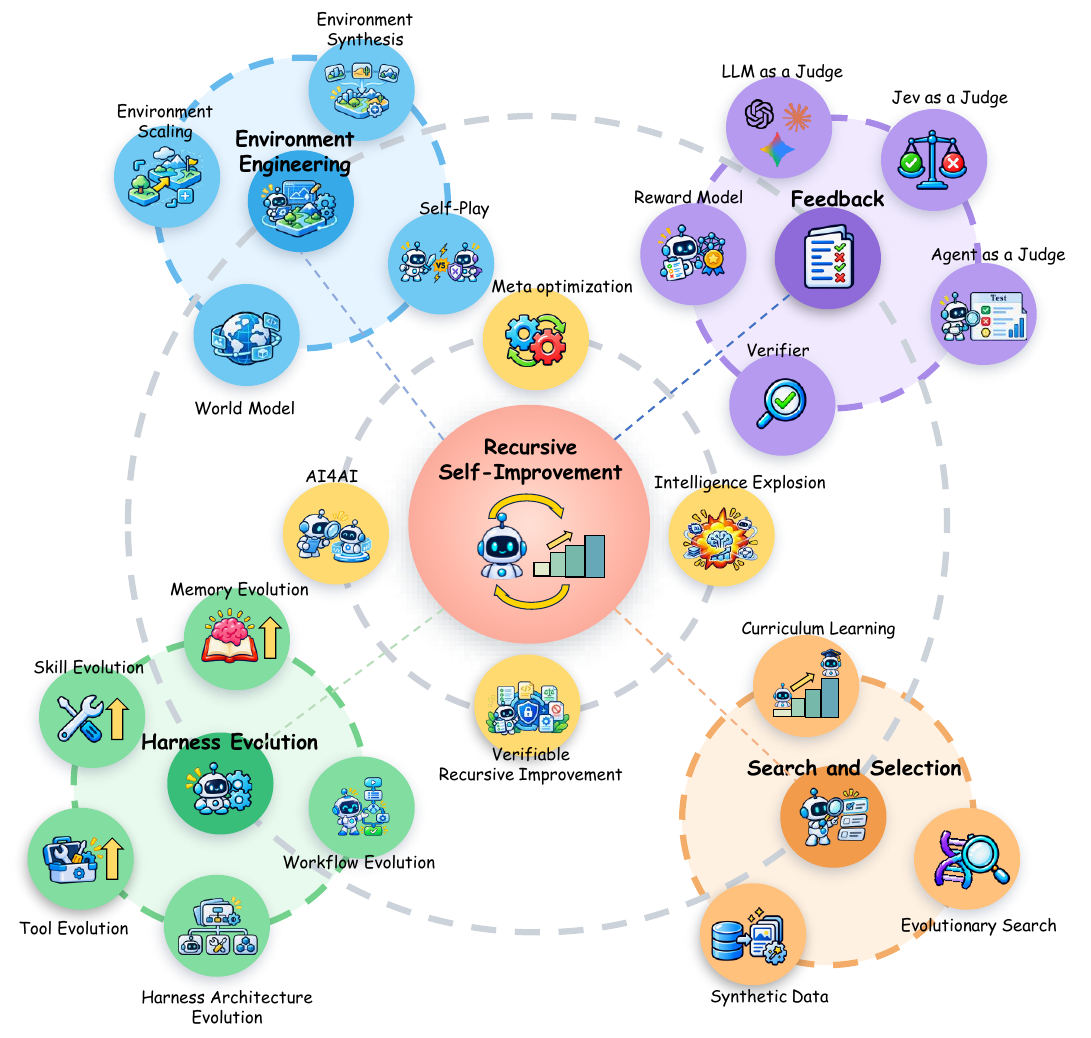}
    \caption{\textbf{Representative Techniques Related to Recursive Self-Improvement.} An overview of selected techniques spanning environment engineering, harness evolution, feedback mechanisms, and search and selection, illustrating their complementary roles in supporting agent improvement and progress toward RSI.}
    \label{fig:techniques}
\end{figure*}

\subsection{Two Complementary Views of Agent Improvement towards RSI}

As illustrated in Fig.~\ref{fig:mountain}, we examine agent improvement through two complementary views: What Improves and Who Drives Improve. What Improves characterizes the objects and scope of improvement, starting from temporal refinement and extending to the model, agent harness, environment, and improver. Temporal refinement primarily optimizes outcomes within the current execution, whereas persistent updates to models and harnesses enable improvement capabilities to accumulate and be reused across runs. The scope of improvement can further extend to environments and learning conditions, and ultimately to the Improver responsible for generating subsequent modifications. These targets represent different scopes of improvement and may evolve jointly within the same system.

Who Drives Improve characterizes the acquisition, diagnosis, interpretation, evaluation, and evolution of improvement feedback. Human-Steered Feedback relies on human evaluation and guidance; Environment-Grounded Feedback derives direct signals from task outcomes and environmental interactions; AI-Facilitated Feedback employs language models, agents, or other decision models to generate analyses and evaluations; and Evolving Feedback persistently updates mechanisms for feedback generation, diagnosis, interpretation, evaluation, and criteria formulation based on accumulated experience. These four feedback categories provide complementary perspectives for analyzing feedback mechanisms in agent improvement. Who Drives Improve and What Improves represent orthogonal dimensions: the former specifies the source and adaptation of improvement signals, whereas the latter identifies the objects being modified. A single feedback mechanism can therefore guide updates to different evolution targets.

Together, these two views establish a unified taxonomy of agent improvement, characterizing existing approaches and their interrelationships. Building on this taxonomy, we organize the survey around a roadmap comprising seven interrelated milestones, as illustrated in Fig.~\ref{fig:road}. The roadmap traces the expansion of improvement targets from output refinement to models, harnesses, and environments, followed by feedback automation and evolution, agent–environment co-evolution, and ultimately RSI. 
Complementing these views, Fig.~\ref{fig:taxonomy} organizes representative studies across the seven milestones, while Fig.~\ref{fig:techniques} summarizes the enabling techniques underlying agent improvement.

%% file: section/agent_evolution.tex
\section{Milestone I: Refining Output}
\label{sec::agent_evolution}




Agent improvement initially focuses on refining task-specific outcomes through iterative critique and revision, without necessarily modifying the underlying system used in subsequent executions. Self-Refine iterates generation, feedback, and revision within one model; RCI extends recursive critique to plans and actions; and Chain-of-Verification uses independently answered questions to check factual claims~\cite{SelfRefine,NEURIPS2023_7cc1005e,dhuliawala-etal-2024-chain}. The common improvement operation is to revise a candidate in the same task, but the location and quality of diagnostic evidence determine what can be repaired. To address the limitations of model-generated critiques, subsequent approaches incorporate external evidence: CRITIC uses search and execution tools, while SelfEvolve and Self-Debugging exploit interpreter errors, execution outcomes, or code explanations for program repair~\cite{CRITIC,jiang2023selfevolvecodeevolutionframework,SelfDebugging}. Fine-grained runtime diagnosis in LDB and repository-level inspection, editing, and testing in SWE-agent extend such repair to complex code artifacts~\cite{zhong-etal-2024-debug,yang2024sweagent}.

Nevertheless, intrinsic self-correction can fail or degrade performance, whereas reliable error identification can improve revision outcomes~\cite{ICLR2024_8b4add8b,tyen-etal-2024-llms,kamoi-etal-2024-llms}. The important boundary is the object being updated, not whether a critique comes from the model or an external tool. Even when a revised task artifact is retained, it does not necessarily represent persistent agent improvement unless the revision also changes reusable system capabilities, such as model parameters or harness components. These limitations motivate two immediate pathways of persistent improvement: internalizing experience into Foundation Model parameters and retaining reusable changes in external Agent Harness components.

\begin{figure*}[t]
    \centering
    \includegraphics[width=1.\linewidth]{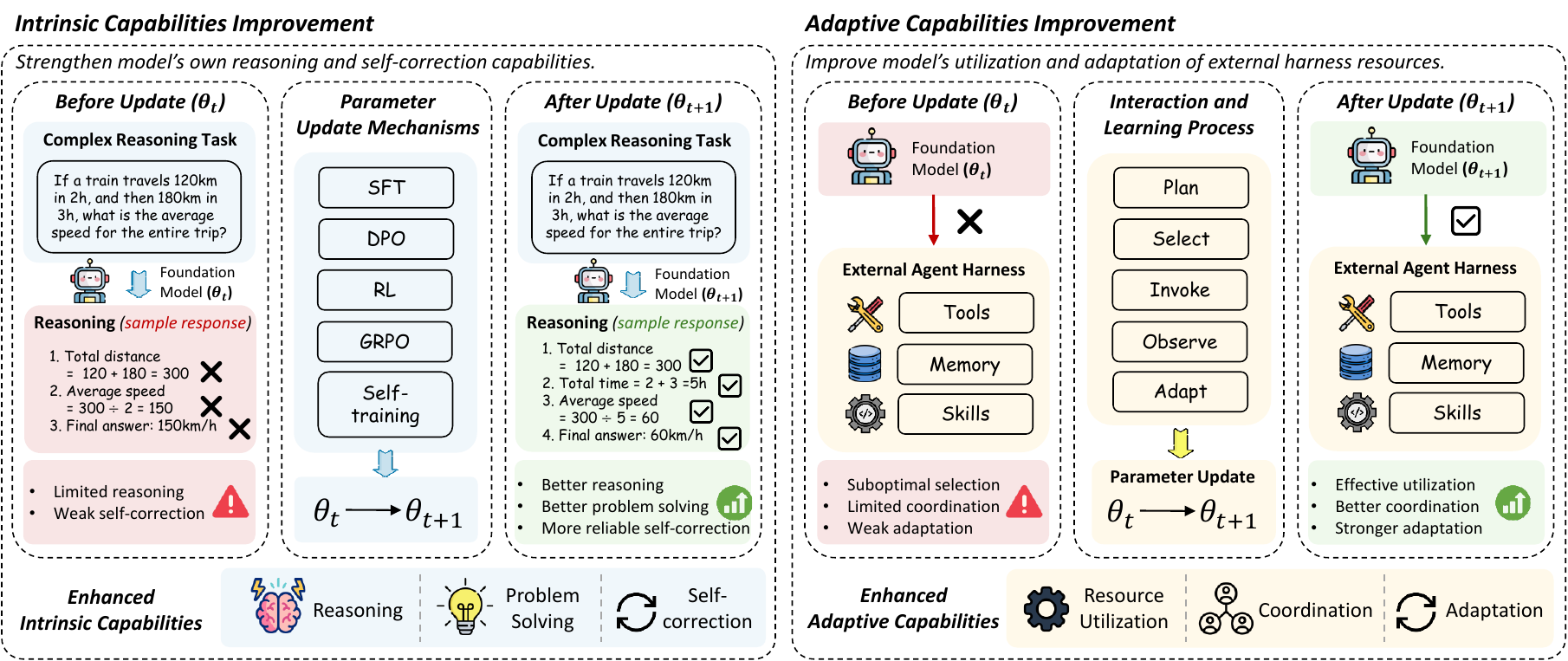}
    \caption{\textbf{Overview of Training Foundation Model.} Intrinsic Capabilities Improvement strengthens reasoning, problem solving and self-correction, while Adaptive Capabilities Improvement enhances the utilization and coordination of external resources. Both establish reusable capabilities through model parameter updates.}
    \label{fig:foundation_model}
\end{figure*}

\section{Milestone II: Training Foundation Model}

Moving beyond output refinement, one route to persistent improvement is to update the foundation model’s parameters using experience from earlier runs. Foundation model evolution internalizes reasoning, execution, and feedback experience through parameter updates, establishing reusable capabilities for subsequent executions. 
As shown in Fig.~\ref{fig:foundation_model}, foundation model evolution follows two complementary pathways: strengthening the model's own problem-solving and correction strategies, and improving its understanding, selection, utilization, and coordination of external harness resources.

\subsection{Enhancing Intrinsic Capabilities}

Enhancing intrinsic capability internalizes training experience and feedback into reusable model capabilities through parameter updates. FLAN and Self-Instruct improve instruction following and task generalization through multi-task instruction tuning and self-generated instruction data, respectively, while InstructGPT combines supervised fine-tuning (SFT), reward modeling, and proximal policy optimization (PPO) to align model behavior with human preferences~\cite{chung2022scalinginstructionfinetunedlanguagemodels,wang2023selfinstructaligninglanguagemodels,instructgpt}. Beyond supervised learning, DPO directly optimizes model policies from preference pairs, whereas KTO, ORPO, and SimPO extend preference optimization through binary desirability signals, joint supervised and preference learning, and reference-free objectives, respectively~\cite{NEURIPS2023_a85b405e,ethayarajh2026ktomodelalignmentprospect,hong2024orpomonolithicpreferenceoptimization,meng2024simposimplepreferenceoptimization}. These methods establish the foundation for feedback-driven parameter improvement.

Beyond external supervision, iterative self-training enables models to improve through self-generated experience. STaR iteratively generates and filters reasoning trajectories for fine-tuning, ReST alternates data generation and policy optimization, and SPIN improves models through self-play against previous generations~\cite{STaR,gulcehre2023rest,chen2024spin}. Quiet-STaR extends this approach to learning internal reasoning during general text prediction~\cite{zelikman2024quietstar}. To enhance training quality, V-STaR incorporates incorrect solutions into verifier training, while ReST-MCTS* and rStar-Math leverage search-generated trajectories and process feedback for iterative policy improvement~\cite{hosseini2024vstar,zhang2024restmcts,guan2025rstarmath}. These approaches progressively internalize self-generated reasoning experience into persistent model capabilities.

Reinforcement learning (RL) further improves reasoning and self-correction through reward-driven parameter optimization. Building on policy optimization methods such as PPO, DeepSeekMath introduces group relative policy optimization (GRPO), which optimizes reasoning policies using group-relative rewards without a separate value model~\cite{shao2024deepseekmath}. DeepSeek-R1 and Open-Reasoner-Zero further demonstrate the effectiveness of reward-driven reasoning optimization~\cite{DeepSeekR1,hu2025openreasonerzero}, while SCoRe employs multi-turn RL to internalize self-correction as a reusable model capability~\cite{SCoRe}. These advances extend parameter-based improvement from supervised behavior learning toward outcome-driven reasoning and correction.

Recent approaches further integrate parameter optimization with autonomous training experience generation. Absolute Zero generates and verifies reasoning tasks through executable environments, while R-Zero alternates Challenger and Solver optimization to jointly improve task generation and problem-solving capabilities~\cite{zhao2025absolutezero,huang2026rzero}. Together, these methods extend model improvement toward autonomous experience generation and learning.

\subsection{Strengthening Adaptive Capabilities}

Beyond intrinsic reasoning, model parameters also determine how agents understand, select, and utilize external resources, including tools, memory, and skills. Toolformer and ToolLLM learn tool-invocation behavior from generated or collected calls, while FireAct and AgentTuning internalize longer agent--tool interaction trajectories; ToolCPT further incorporates tool functionality and protocol knowledge during continued pretraining~\cite{schick2023toolformer,qin2024toolllm,chen2023fireact,AgentTuning,yang2026toolcpt}. These studies establish learning to use available tools, but do not alone establish adaptation to tools that evolve after training. Beyond learning individual tool interactions, parameter optimization can improve how agents manage and coordinate external resources across successive decisions. Memory-R1 and AgeMem train decisions for storing, updating, retrieving, and using memory~\cite{yan2026memoryr1,yu2026agemem}. SkillRL and SAGE connect skill use to policy optimization, while Skill1 and SkillGraph couple learned skill selection or composition with updates to external skill resources~\cite{xia2026skillrl,wang2026sage,shi2026skill1,li2026skillgraph}. SkillGym instead turns external procedures into training experience, allowing some skill knowledge to be internalized into model behavior~\cite{ge2026skillgym}.

As external harness resources become more diverse and dynamic, model adaptation increasingly involves flexible resource selection and coordination across changing execution conditions. Multi-turn tool learning further improves the integration of search and code execution into reasoning~\cite{SearchR1,feng2025retool}; AutoTool targets selection across changing inventories, and ToolOrchestra trains a dedicated controller to coordinate heterogeneous models and tools~\cite{zou2026autotool,su2026toolorchestra}. Agent Lightning provides a complementary means of converting trajectories from existing harnesses into parameter-training signals~\cite{luo2025agentlightning}. Together, these approaches extend model learning from individual resource utilization toward adaptive selection, multi-step coordination, and interaction with evolving external capabilities, providing complementary foundations for the harness evolution discussed in the next milestone.

\section{Milestone III: Evolving Agent Harness}

\begin{figure*}[t]
    \centering
    \includegraphics[width=1.\linewidth]{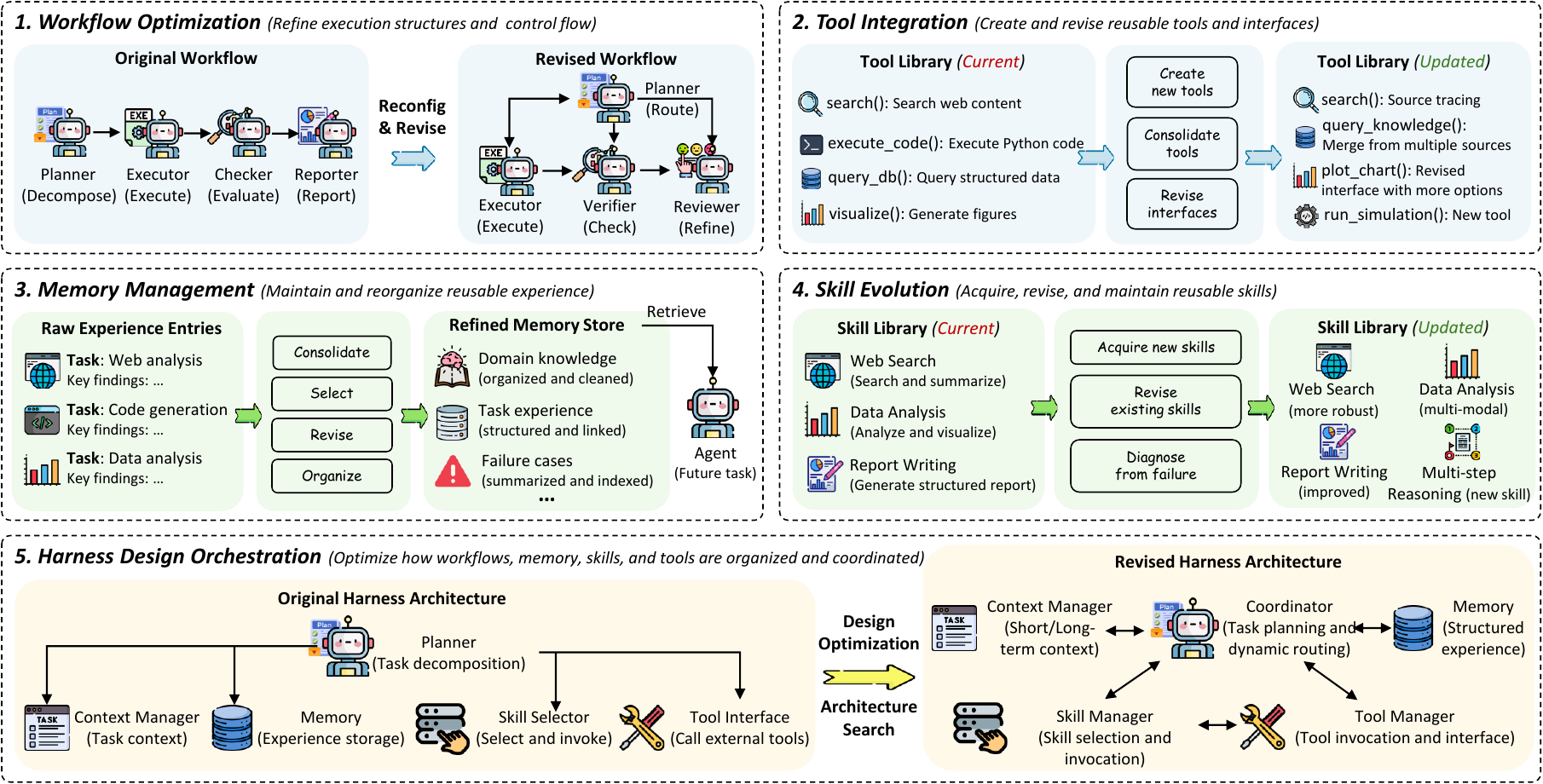}
    \caption{\textbf{Overview of Evolving Agent Harness.} Workflow Optimization reorganizes execution structures and control flow to improve task processing. Tool Integration expands reusable tool libraries through tool creation, consolidation, and interface revision. Memory Management transforms raw execution experiences into structured and retrievable knowledge, task experiences, and failure cases. Skill Evolution acquires new skills, revises existing ones, and diagnoses failures to improve reusable capabilities. Finally, Harness Design Orchestration optimizes the overall harness architecture and coordinates the interactions among context management, planning, memory, skills, and tools, enabling more effective and reusable agent systems.}
    \label{fig:harness}
\end{figure*}

Although foundation model evolution enables persistent capability improvement, agent performance also depends on the external structures and resources supporting execution. Agent Harness Evolution extends improvement to workflows, tools, memory, skills, and integrated harness architectures, transforming experience into reusable system-level capabilities, as illustrated in Fig.~\ref{fig:harness}. This section reviews these evolution mechanisms and their roles and limitations in supporting sustained agent improvement.


\subsection{Workflow Optimization}

Execution performance depends partly on how existing capabilities are organized, making workflows themselves objects of optimization. GPTSwarm optimizes computational graphs, while ADAS, AutoFlow, and AFlow explore reusable agent workflows and programmatic execution structures through automated generation and search~\cite{hu2024adas,zhuge2024gptswarm,li2024autoflow,zhang2024aflow}. These methods expose the structure used to generate future outputs, rather than merely rewriting the outputs themselves. Joint prompt--topology optimization and communication-graph pruning further revise component configurations and execution connections~\cite{zhou2026mass,zhang2024agentprune}. Rather than optimizing a single fixed structure, EvoFlow evolves heterogeneous workflow populations, while A$^2$Flow abstracts reusable operators that support later workflow construction~\cite{zhang2025evoflow,zhao2026a2flow}. Search can modify a single program, maintain multiple alternatives, or discover reusable construction units; each leaves a different artifact for later execution. The resulting graphs, programs, and operators can persist across runs, although offline workflow search and sustained adaptation under deployment shifts require distinct evaluation.

\subsection{Tool Integration}

Tool integration changes the callable functions and interfaces available in subsequent executions, rather than merely training models to invoke fixed tools. LATM and CRAFT create, validate, and retain reusable functions, while TroVE and ReGAL extend tool reuse through toolbox maintenance and programmatic abstraction~\cite{cai2024large,yuan2024craft,wang2024trove,stengeleskin2024regal}. However, creating a new function is insufficient when a growing library becomes redundant, difficult to retrieve, or incompatible with changed usage requirements. As inventories grow, ToolLibGen consolidates shared functionality and GATE evolves tool implementations and hierarchical dependencies through merging, checking, and pruning~\cite{yue2025toollibgen,luo2026gate}. Tool behavior and the knowledge required to invoke it are separate evolution targets. ContDa additionally updates tool documentation, usage preferences, and fallback relations as capabilities change~\cite{wu2026contda}. These updates preserve external executable functionality and its accessibility, but a larger toolset alone need not yield better utilization.

\subsection{Memory Management}

Memory management goes beyond retaining past interactions to continually refining the experience available to future decisions. Reflexion preserves feedback across attempts, whereas ExpeL and ReasoningBank extract reusable guidance from successful and failed task trajectories~\cite{shinn2023reflexion,zhao2024expel,ouyang2026reasoningbank}. As experience accumulates, Dynamic Cheatsheet consolidates strategies and A-MEM revises associations and contextual attributes of existing records~\cite{suzgun2026dynamiccheatsheet,xu2025amem}. Accumulation alone can introduce duplicated, obsolete, or low-value experience; later task performance must therefore also shape which records remain influential. ReMe and Darwinian Memory maintain experience through utility-sensitive refinement and selection, while MemRL explicitly updates memory-value estimates from later task feedback~\cite{cao2026reme,mi2026darwinianmemory,zhang2026memrl}. These updates change not just what is remembered but what the agent will preferentially retrieve or reuse. Beyond individual records, MemEvolve optimizes memory architectures, EvolveMem revises retrieval configurations following failures, and MemSkill adapts memory-management skills with a learned selector~\cite{zhang2026memevolve,liu2026evolvemem,zhang2026memskill}. Thus, persistent memory improvement concerns the content, organization, value, and utilization procedures of accumulated knowledge, not storage growth alone.

\subsection{Skill Evolution}

Skill evolution transforms execution experience into reusable task-solving procedures that agents can retrieve, adapt, or execute in subsequent tasks. Voyager and Agent Skill Induction acquire executable skills, while SkillWeaver and AutoSkill extend acquisition through exploration, practice, and accumulated interactions~\cite{wang2023voyager,wang2025asi,zheng2025skillweaver,yang2026autoskill}. However, accumulating reusable skills is insufficient when existing procedures contain incorrect assumptions, become unsuitable for new tasks, or conflict with other skills, motivating their continual revision and maintenance. Subsequent evolution modifies existing procedures and libraries: Trace2Skill consolidates trajectory-derived patches, SkillX refines hierarchical skills, and SkillClaw updates shared skills from cross-user experience~\cite{ni2026trace2skill,wang2026skillx,ma2026skillclaw}. Reliability requires more than adding skills. PSN revises skill networks with fault localization and rollback, while SkillForge and ContractSkill support diagnosis-driven or contract-guided local repair~\cite{shi2026psn,liu2026skillforge,lu2026contractskill}. CoEvoSkills iteratively refines complex skill packages through joint generation and verification; Skill-R1 instead trains an auxiliary generator using successive skill outcomes, without necessarily establishing persistent cross-task library evolution~\cite{zhang2026coevoskills,vishe2026skillr1}. These approaches also illustrate two different update carriers: a revised external skill and, in some cases, a trained model that proposes revisions. Skill evolution therefore encompasses acquisition, revision, consolidation, and compositional maintenance of reusable operational capabilities.

\subsection{Harness Design Orchestration}

Although individual harness components can evolve independently, their effectiveness also depends on how they are organized and coordinated within the complete agent system, making integrated harness design an additional improvement target. AgentSquare evolves and recombines functional modules, while Meta-Harness searches executable harness code using previous implementations, scores, and execution traces~\cite{shang2025agentsquare,lee2026metaharness}. This broadens the scope of modification beyond workflow topology to the resource presentation, context handling, and cross-component settings that determine whether a given model can effectively execute a task. Execution-driven approaches revise deployed configurations: Self-Harness validates weakness-driven edits through regression tests, while Continual Harness updates prompts, memory, skills, and subagents during uninterrupted interaction~\cite{zhang2026selfharness,karten2026continualharness}. Across changing tasks, Adaptive Auto-Harness maintains and routes among alternative designs, whereas JIT-Agent uses a separately trained generator and a persistent archive for task-conditioned harness construction~\cite{liu2026adaptiveautoharness,zhang2026jitagent}. The timescale of adaptation also matters: updates retained after a bounded design search, changes made within an uninterrupted session, and revisions inherited by separate future runs provide different kinds of evidence. Crucially, producing an update is not equivalent to using it beneficially: effectiveness depends on whether the underlying model can activate and follow the revised artifacts~\cite{lin2026harnessbenefit}. Integrated harness evolution thus requires assessing retained design changes and their downstream use, complementing parameter-based improvement.

%% file: section/environment_evolution.tex
\section{Milestone IV: Building Environment}
\label{sec::environment_evolution}



\begin{figure*}[t]
    \centering
    \includegraphics[width=\textwidth]{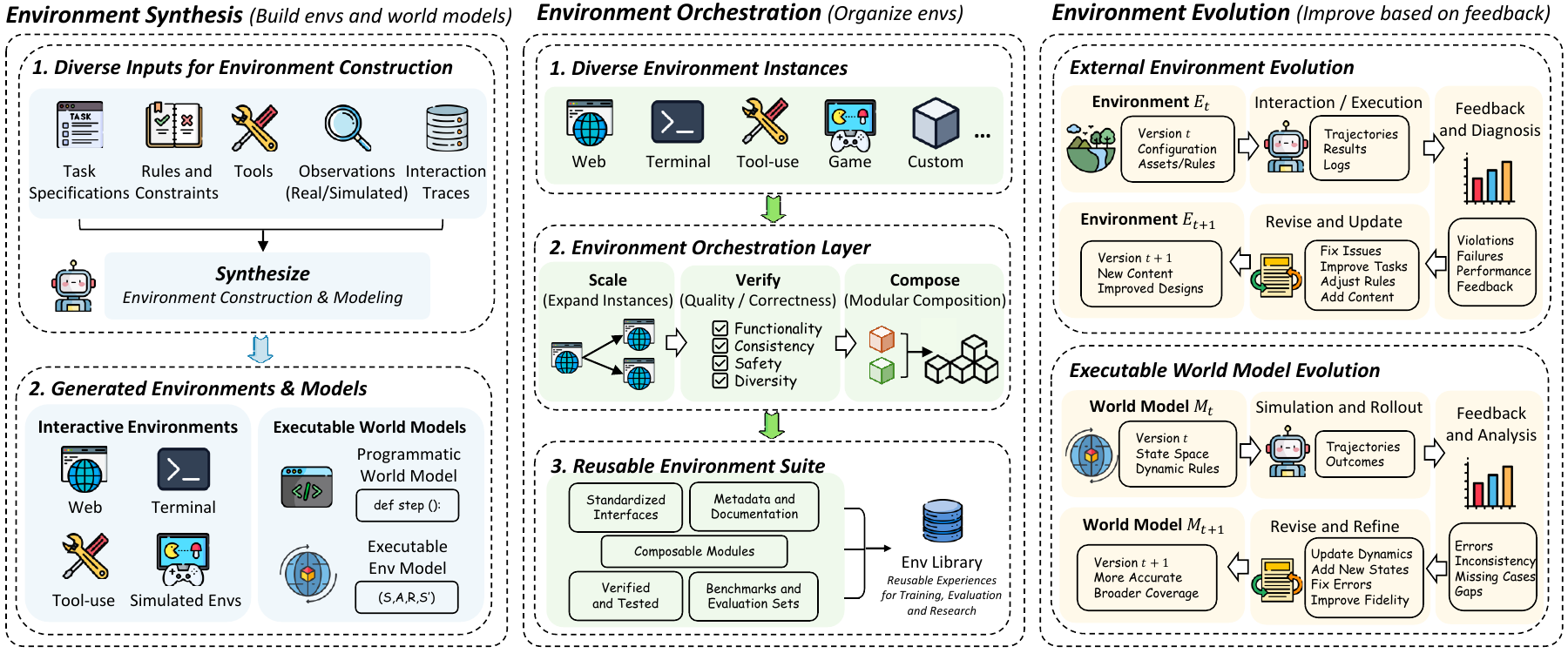}
    \caption{\textbf{Overview of Building Environment.} Environment Synthesis constructs interactive environments and executable world models from task requirements, interaction rules, and existing resources. Environment Orchestration scales, verifies, and composes individual environments into reusable environment resource systems. Environment Evolution continuously revises environments and world models through interaction trajectories and execution feedback.}
    \label{fig:environment_evolution}
\end{figure*}

Although advances in foundation models and agent harnesses continuously enhance agent capabilities, sustained agent learning also depends on diverse, reliable, and adaptable environments. Static or manually designed environments often fail to cover diverse task requirements or provide effective learning experiences as agent capabilities improve. Environments are therefore becoming not only external settings for agent execution but also objects of autonomous construction, organization, and continuous improvement.
Based on their development processes, we categorize existing studies into three complementary stages, as illustrated in Fig.~\ref{fig:environment_evolution}: Environment Synthesis constructs interactive environments and executable models from task specifications or existing resources; Environment Orchestration scales, verifies, and composes environments into reusable resource systems; and Environment Evolution continuously updates existing environments and their models through interaction experience and execution feedback.

\subsection{Environment Synthesis}

Environment Synthesis focuses on automatically constructing environments for agent learning and execution from task requirements, environmental rules, and interaction constraints. Early efforts explore using language models and programmatic generation to transform task descriptions, capability requirements, and interaction conditions into executable environments. EnvGen, Agent-World, Envs-FORGE, and AgentGen generate targeted training tasks and interaction conditions based on agent capability gaps, performance, or task constraints~\cite{envgen,agentworld,envsforge,agentgen}. 
AutoForge further automates the construction of complex environments and tasks~\cite{cai-etal-2026-autoforge}. 
In game and simulation settings, PCGRL+ and Agentic PCG employ reinforcement learning-based procedural generation and tool-using language models to construct controllable levels and interactive scenarios~\cite{earle2024pcgrlscalingcontrolgeneralization,jiang2026agentic}, while SimWorld Studio leverages an evolving coding agent to automatically generate simulated environments for embodied agent learning~\cite{kang2026simworldstudioautomaticenvironment}. These approaches extend environment construction beyond manual design, enabling task rules, state spaces, and interaction conditions to be instantiated according to different learning requirements.

Beyond constructing external interactive environments, another line of research builds executable environmental representations that enable agents to simulate, predict, and plan over environmental dynamics. WorldCoder, GIF-MCTS, PoE-World, and Code as Worlds construct executable world models through program generation, search, and compositional modeling, with interaction-based testing used to refine environmental dynamics~\cite{worldcoder,gifmcts,poeworld,codeasworlds}. Agent World Model further integrates tools, programmatic states, and environmental dependencies into interactive tool-use worlds~\cite{pmlr-v306-wang26jh}. WALL-E extracts environmental interaction rules from exploration trajectories and organizes them into reusable structured knowledge~\cite{walle2}. Unlike approaches that directly generate external environments, these methods explicitly encode environmental states, transitions, and interaction rules as programs or structured models, providing a foundation for subsequent simulation, prediction, and planning. Overall, Environment Synthesis establishes the capability to construct environments and their executable models, but individual environments alone cannot satisfy the need for large-scale, diverse, and reliable agent learning experiences.

\subsection{Environment Orchestration}

As the number and diversity of generated environments increase, organizing individual environments into scalable, reliable, and reusable resource systems becomes increasingly important. Environment Orchestration therefore addresses not only environment scaling but also reliability verification and effective composition of environmental components. First, several studies expand the scale and diversity of available environments. AgentScaler, EnvScaler, ScaleEnv, and AutoEnv construct heterogeneous tool-interactive environments through programmatic synthesis, tool-dependency modeling, and scenario instantiation~\cite{fang-etal-2026-towards,song-etal-2026-envscaler,pmlr-v306-tu26a,autoenv}. For different interaction scenarios, InfiniteWeb and AutoWebWorld expand executable web environments through automated construction and finite-state-machine representations~\cite{zhang-etal-2026-infiniteweb,pmlr-v306-wu26ar}, while Endless Terminals and SETA further develop scalable, containerized environments for terminal agents~\cite{gandhi2026endlessterminalsscalingrl,seta}. Meanwhile, some approaches organize learning experiences by dynamically adjusting task distributions and difficulty according to agent performance. WebRL, Self-Evolving Curriculum, RLVE, and SCALER adapt task selection, difficulty, or environment allocation based on failure trajectories and performance feedback~\cite{webrl,sec,rlve,scaler}. DreamGym and CuES further expand learning experiences through simulated interactions and curiosity-driven exploration~\cite{dreamgym,cues}. Together, these approaches expand the available environment space and interaction experiences for agent learning.

Environment scaling also introduces reliability challenges. Automatically generated environments may contain invalid states, inconsistent rules, or unreliable interactions, making systematic verification necessary to ensure functional and semantic correctness. For digital environments, SWE-Universe, EnvFactory, and Training Needs Trustworthy Worlds examine state consistency, task executability, and the reliability of stateful tool interfaces~\cite{chen2026sweuniversescalerealworldverifiable,xu2026envfactoryscalingtooluseagents,zhang2026trainingneedstrustworthyworlds}. High-Performance RL Environments further employs hierarchical testing to identify and repair simulator implementation errors while preserving their semantics~\cite{karten2026automaticgenerationhighperformancerl}. For executable environment models, Kepler emphasizes the auditability of model rules and supporting evidence~\cite{kepler}. These studies extend environment verification beyond simple post-generation checks toward systematic validation of functionality, states, and interactions.

Beyond scaling and verification, environment composition provides another means of constructing complex interactive scenarios. Rather than independently generating complete environments, modular composition reuses existing environmental structures, task rules, and interaction components to produce new instances through different combinations. RACES recursively composes verifiable environment modules to construct task variants with diverse reasoning structures~\cite{xiang2026verifiableenvironmentslegobricks}, while LOGIGEN generates verifiable agentic tasks through executable logical constraints~\cite{zeng2026logigenlogicdrivengenerationverifiable}. EnvScaler expands tool-interactive environments through environment skeletons and scenario instantiation~\cite{song-etal-2026-envscaler}, whereas ScaleEnv leverages tool dependencies and programmatic structures to support the construction of diverse task scenarios~\cite{pmlr-v306-tu26a}. These approaches advance environment construction from independent instance generation toward the systematic organization of reusable components. Overall, Environment Orchestration transforms scattered environment instances into resource systems with greater scale, diversity, reliability, and compositionality, providing a more comprehensive foundation for continuous agent training and evaluation.

\subsection{Environment Evolution}

Although environment synthesis and orchestration provide diverse interaction resources, static environments struggle to adapt to evolving agent capabilities and problems exposed during execution. Environment Evolution therefore continuously updates existing environments and their models through execution experience and feedback. For external interactive environments, Environment Evolution for Terminal Agents incrementally modifies environment lineages across generations to increase task complexity and difficulty~\cite{terminalenv_evolution}, while EnvHarness adapts environmental interfaces and task configurations through component updates guided by agent trajectories and training signals~\cite{envharness}. RSIGame and Play2Code diagnose environmental problems from execution outcomes and iteratively repair interactive game environments through code modifications and feedback-based verification~\cite{wu2026rsigameautonomousagenticgame,huang-etal-2026-gui-agents}. Furthermore, ProEvolve and Env-Rethink generate new environmental conditions by modifying coupled tool, schema, and data structures, as well as workspace states and evidence relationships, respectively~\cite{proevolve,envrethink}. These approaches transform environmental configurations, interaction rules, and task structures into persistently modifiable objects that adapt to evolving execution requirements.

Beyond modifying interactive environments, another line of research focuses on continuously updating World Models to better simulate environmental rules, state transitions, and action outcomes. WorldEvolver and ProPlay refine world models using prediction discrepancies and interaction feedback, enabling their environmental representations to adapt to newly observed states and dynamics~\cite{worldevolver,proplay}. OPINE-World and EvoBelief refine environmental rules and dynamic hypotheses through targeted exploration, counterexample discovery, and hypothesis falsification~\cite{opineworld,evobelief}. Twin, Memento 3, and D-World further employ observation discrepancies, interaction replay, and consistency checks to refine predicted state transitions or digital twin models~\cite{twin,memento3,dworld}. Rather than merely accumulating environmental knowledge or recording interaction experience, these approaches treat environmental rules and dynamics within world models as persistent improvement targets, enabling updated models to support subsequent simulation, prediction, and planning. Overall, Environment Evolution transforms environments from preconstructed static resources into dynamic systems that continuously update through interaction experience.

%% file: section/feedback_autonomy.tex
\section{Milestone V: Automating Feedback}
\label{sec::feedback_autonomy}

\begin{figure*}[t]
    \centering
    \includegraphics[width=\textwidth]{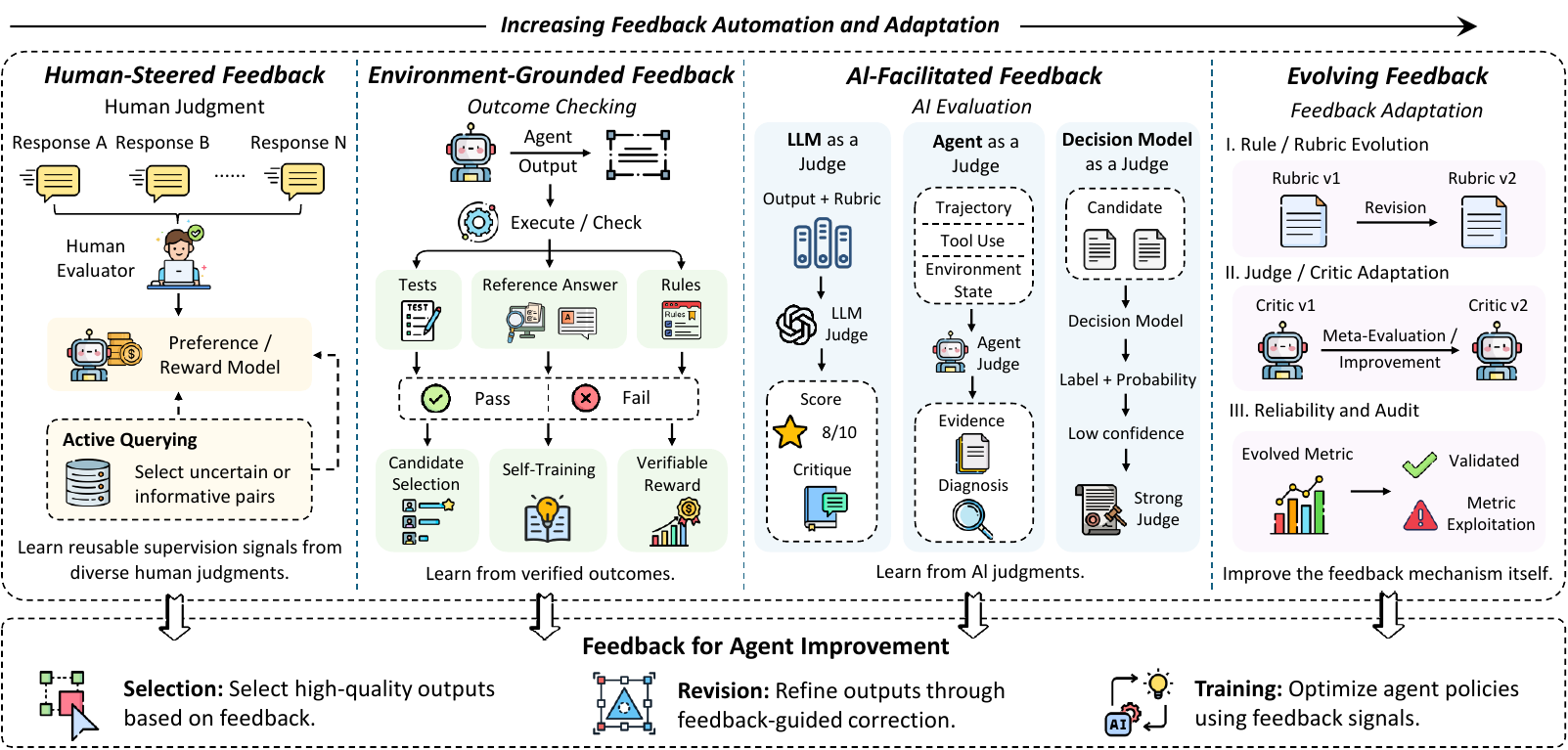}
    \caption{\textbf{Overview of Automating Feedback.} Human-Steered Feedback derives supervision from human evaluations and preferences, while Environment-Grounded Feedback provides verifiable signals through execution outcomes, tests, and rule-based checks. AI-Facilitated Feedback employs language models, agents, and decision models for evaluation and diagnosis. Evolving Feedback further adapts evaluation rules, judges, and feedback metrics.}
    \label{fig:feedback_autonomy}
\end{figure*}



As improvement extends from agent components to environments, identifying deficiencies, evaluating candidate modifications, and guiding subsequent updates become increasingly challenging. Although feedback underlies all preceding improvements, sustained autonomy requires reducing dependence on human judgments through increasingly automated feedback acquisition, interpretation, and evaluation. As illustrated in Fig.~\ref{fig:feedback_autonomy}, this section reviews four feedback mechanisms: Human-Steered Feedback, Environment-Grounded Feedback, AI-Facilitated Feedback, and Evolving Feedback, tracing the progression toward more autonomous and adaptive mechanisms for guiding agent improvement.

\subsection{Human-Steered Feedback}

Human-steered feedback provides the foundational supervision for agent improvement, allowing humans to evaluate outputs, compare behaviors, and express desired qualities. Preference learning translates comparisons of trajectory segments into reward functions that guide repeated behavior optimization~\cite{human_preferences}. The same principle extends to language: summary optimization and InstructGPT learn rewards from response comparisons, reusing limited annotations during subsequent training~\cite{summarize,instructgpt}. WebGPT applies preference supervision to answers supported by browser-collected sources, extending its use to tool-assisted tasks~\cite{webgpt}. Iterated training of helpful and harmless assistants refreshes preference models and policies with new human comparisons as behavior changes~\cite{helpful_harmless}. Together, these methods turn limited judgments into reusable supervision, extending human guidance across updates while reducing direct annotation at each step.

Continued preference collection makes annotation efficiency central to sustained improvement. Experience reuse and additional unlabeled data increase the value of collected judgments: PEBBLE relabels past experience when its reward model changes, while SURF combines confidence-based pseudo-labeling with trajectory augmentation~\cite{pebble,surf}. Active acquisition directs annotation toward useful comparisons, using uncertainty in preference learning or balancing information with representation coverage in active reward modeling~\cite{active_preference,active_reward}. PILAF further designs response sampling around the policy-improvement objective, connecting feedback collection to iterative optimization~\cite{pilaf}. These mechanisms reduce feedback acquisition and utilization costs while retaining human judgments as the supervisory basis. Directly verifiable outcomes provide a complementary source of automatic supervision.



\subsection{Environment-Grounded Feedback}

Human-steered feedback provides flexible supervision, but repeatedly collecting human judgments limits the scalability of agent improvement. Environment-grounded feedback instead derives automatic improvement signals from verifiable task outcomes, enabling candidate selection and subsequent training without requiring human evaluation for each attempt. Behavioral filtering in AlphaCode and dual execution agreement in CodeT support automatic candidate selection; CodeT additionally generates tests with a language model while grounding the ranking in execution results~\cite{alphacode,codet}. Beyond selection, correctness checks turn generated experience into reusable training supervision. STaR retains rationales leading to correct answers, including rationales regenerated with the known answer, while ReST-EM alternates generation, binary outcome filtering, and fine-tuning~\cite{STaR,ReSTEM}. DeepSeekMath's rule-reward experiments, Tulu-3's reinforcement learning with verifiable rewards, and DeepSeek-R1's rule-reward reasoning stages further connect checked outcomes to policy optimization~\cite{shao2024deepseekmath,tulu3,DeepSeekR1}. Automatic assessment thereby supports both current candidate selection and capability updates for future attempts.

Beyond outcome-based selection and rewards, execution feedback supplies finer-grained signals for learning from failures and guiding subsequent attempts. RLTF converts unit-test outcomes and error information into online reinforcement-learning signals at multiple granularities, while RLEF trains multi-step code generation to exploit execution feedback during successive revisions~\cite{rltf,rlef}. SWE-Gym extends executable supervision to repository-level tasks, coupling real codebases and tests with trajectories used to train software-engineering agents~\cite{swe_gym}. Feedback utility depends on test coverage, reference correctness, and the properties captured by assessment rules. Execution evidence can reveal failures within this coverage; interpreting their causes and assessing open-ended quality motivate complementary AI-based interpretation and diagnosis.



\subsection{AI-Facilitated Feedback}

Environment-grounded feedback automates assessment of verifiable outcomes but struggles with open-ended quality and complex failure diagnosis. AI-facilitated feedback addresses these settings through three complementary evaluation paradigms: generative semantic judgment, agentic evidence acquisition and verification, and specialized decision-oriented evaluation.


\subsubsection{LLM as a Judge}

Many open-ended tasks require semantic judgments that cannot be reliably obtained from executable checks alone. LLM-based judges automate these assessments through task-specific criteria, preference comparisons, and natural-language critiques, providing feedback for candidate selection and subsequent improvement. G-Eval organizes generated evaluation steps and structured scoring, while MT-Bench and Chatbot Arena examine agreement between LLM judgments and human preferences~\cite{geval,mtbench}. Dedicated training makes this evaluative function reusable: PandaLM compares instruction-tuning candidates to guide configuration selection, and JudgeLM distills generated judgments into scalable evaluators~\cite{pandalm,judgelm}. These judgments support candidate selection and, when accompanied by evaluative reasoning or critiques, can identify quality deficiencies beyond those captured by executable checks.

Actionable feedback requires specifying the dimensions behind an overall judgment. FLASK decomposes assessment into instruction-relevant skills, while Prometheus and Prometheus 2 train evaluators to use supplied rubrics and provide feedback across scoring and comparison formats~\cite{flask,prometheus,prometheus2}. These mechanisms make evaluation criteria explicit and provide finer-grained signals that can inform subsequent revision or optimization. Generated criticism further supports task-local correction in Self-Refine and code-defect detection for human oversight in CriticGPT~\cite{SelfRefine,criticgpt}. Persistent capability improvement requires incorporating such feedback into subsequent model or system updates.

AI judgments scale improvement when they become preference data or optimization signals. Constitutional AI links principle-guided criticism and revision to supervised training, then learns from AI-generated preferences, and UltraFeedback supplies diversified multidimensional judgments for selection and alignment~\cite{constitutional,ultrafeedback}. RLAIF experiments further demonstrate policy training with AI preferences or direct judge rewards~\cite{rlaif}. Position bias and other evaluation artifacts can distort model-generated preferences, as documented by empirical studies and broader judge surveys~\cite{fair_evaluators,judge_survey}. JETTS finds competitive reranking but weaker step-level search than process reward models, and ineffective critique-guided refinement in its tested settings~\cite{jetts}. Evaluation agreement, revision success, and training gains therefore provide complementary evidence of feedback utility.



\subsubsection{Agent as a Judge}

Evaluating complex agent tasks may require inspecting intermediate actions, execution trajectories, and external evidence beyond the initially supplied response. Agentic judges address this need through multi-step reasoning, active evidence acquisition, and tool-assisted verification. Requirement-guided inspection extends evaluation beyond final answers: Agent-as-a-Judge checks hierarchical requirements, and CodeVisionary combines contextual analysis with fine-grained code assessment~\cite{agent_judge,codevisionary}. Structured checks extend this approach across interactive settings, covering source attribution in Mind2Web 2, policy-sensitive dialogue diagnostics in IntellAgent, and decomposed task-completion requirements in Auto-Eval Judge~\cite{mind2web2,intellagent,auto_eval}. AJ-Bench evaluates information acquisition, state verification, and process verification as distinct capabilities needed for agentic assessment~\cite{ajbench}. Together, these methods connect requirements to execution evidence for finer-grained identification of deficiencies.

Evidence outside the recorded response motivates active verification during judgment. Themis invokes external tools for preference assessment, while VerifiAgent selects reasoning-specific verification tools and returns feedback for further reasoning~\cite{themis,verifiagent}. EvalAgent retrieves expert-authored guidance to identify implicit task-specific criteria, whereas FACT-AUDIT adapts fact-checking probes and assesses supporting justifications~\cite{evalagent,fact_audit}. GUI evaluation further requires inspecting visual interfaces and task-relevant environment states: GUI-PRA gathers criterion-guided visual evidence, while VAGEN combines trajectory inspection with targeted environment probing~\cite{gui_pra,vagen}. These approaches automate acquisition and verification of task-specific evidence and, in some cases, discovery of evaluation criteria, extending assessment beyond the original prompt.

Connecting evidence-grounded judgments to candidate selection and training gives agentic evaluation a direct role in subsequent improvement. RewardAgent combines preferences with factuality and instruction-following verification for candidate selection and preference-based training, while OpenReward trains the tool-augmented reward model OpenRM for response selection and training-data selection~\cite{rewardagent,openreward}. Specialized training also supplies reusable checking capabilities: AgentV-RL learns tool-interleaved verification, while competitive-coding verifiers learn to generate discriminative inputs exposing candidate discrepancies~\cite{agentv_rl,scaling_verifier}. Multi-step reasoning and tool interactions add evaluation costs, motivating compact judgments for frequent decisions.

\subsubsection{Decision Model as a Judge}

LLM-based and agentic judges provide flexible evaluations, but their reasoning and tool interactions can be costly for frequent assessments. Decision models enable efficient, structured judgments and support selective escalation to more capable judges when additional evaluation is needed. PairRanker learns joint candidate comparisons, MiniCheck verifies claims against documents, and BERT-as-a-Judge assesses answer correctness against references~\cite{pairranker,minicheck,bert_judge}. Luna-2 supplies single-token evaluation, while Skywork-Reward-V2 develops preference-trained scoring models~\cite{luna2,skywork}. RewardBench benchmarks preference-based reward models, supporting assessment under representative comparison tasks~\cite{rewardbench}. Jev provides a decision-only interface that directly returns typed verdicts and label probabilities~\cite{jev}. These outputs support candidate comparison and confidence-guided routing, which allocates further evaluation to selected cases.

Recent studies further investigate confidence-based routing between efficient decision models and more capable reasoning judges. JEV-as-a-Judge tests accepting confident verdicts and routing uncertain cases to reasoning judges, while Trust or Escalate calibrates selective acceptance to a specified level of human agreement~\cite{jev,trust_escalate}. Uncertainty-based reward-model routing also improves online alignment by calling a strong judge for uncertain preference pairs~\cite{ask_strong}. Rubric-judging experiments find correlated decision-model and generative-judge errors that limit cascade accuracy gains in the tested workloads~\cite{jev_rubric}. Broader Jev benchmarking distinguishes useful probability rankings from unsuitable fixed binary thresholds~\cite{jev_benchmark}. Routing benefits therefore depend on workload-specific threshold validation and the judges' joint error patterns. While these approaches automate judgment under largely fixed evaluation mechanisms, sustained agent improvement also motivates adapting the rules, criteria, and procedures that produce feedback across successive improvement cycles.

\subsection{Evolving Feedback}

Although automated feedback reduces reliance on human evaluation, fixed evaluation and diagnostic mechanisms may become less informative as agent capabilities evolve. Evolving Feedback addresses this limitation by persistently updating the mechanisms used to generate, interpret, diagnose, and evaluate feedback, including their underlying rules and criteria. Diagnostic-driven reward refinement revises reward functions using training evidence in sparse structured tasks~\cite{reward_refinement}. DynamicRubric and EvoRubric adapt rubric-generation mechanisms as response policies evolve, while GenRubric trains rubric generation using self-consistency-based signals~\cite{dynamic_rubric,evo_rubric,genrubric}. SERPO maintains discriminative criteria as response archives and policies evolve, while EvoRS expands revision to executable reward systems integrating scoring and signal composition~\cite{serpo,evors}. Beyond reward rules and rubrics, Who Grades the Grader evolves evaluation metrics by composing failure detectors against anchored references~\cite{zhang2026whogrades}. Its anchor-removal results further show that downstream skill gains do not necessarily establish the validity of an evolved evaluation metric.

Beyond explicit rules and rubrics, learned judges and critics become persistent targets of feedback-driven optimization. Self-Rewarding incorporates model-generated judgments into iterative preference optimization, while Self-Taught Evaluators repeatedly train judging capabilities using synthetic comparisons~\cite{self_rewarding,self_taught}. Meta-Rewarding trains judging capabilities using preferences over model judgments, whereas ECHO updates a trajectory critic according to the policy gains induced by its diagnoses~\cite{meta_rewarding,echo}. These methods extend persistent feedback adaptation to learned evaluative and diagnostic capabilities, using synthetic preferences, meta-evaluation, or downstream improvement outcomes as update signals. However, increasingly autonomous feedback does not by itself ensure that persistent changes to agents and environments remain coordinated as their capabilities and learning conditions evolve.

%% file: section/co_evolution.tex
\section{Milestone VI: Co-Evolving Agent and Environment}
\label{sec:co_evolution}

\begin{figure*}[t]
    \centering
    \includegraphics[width=\textwidth]{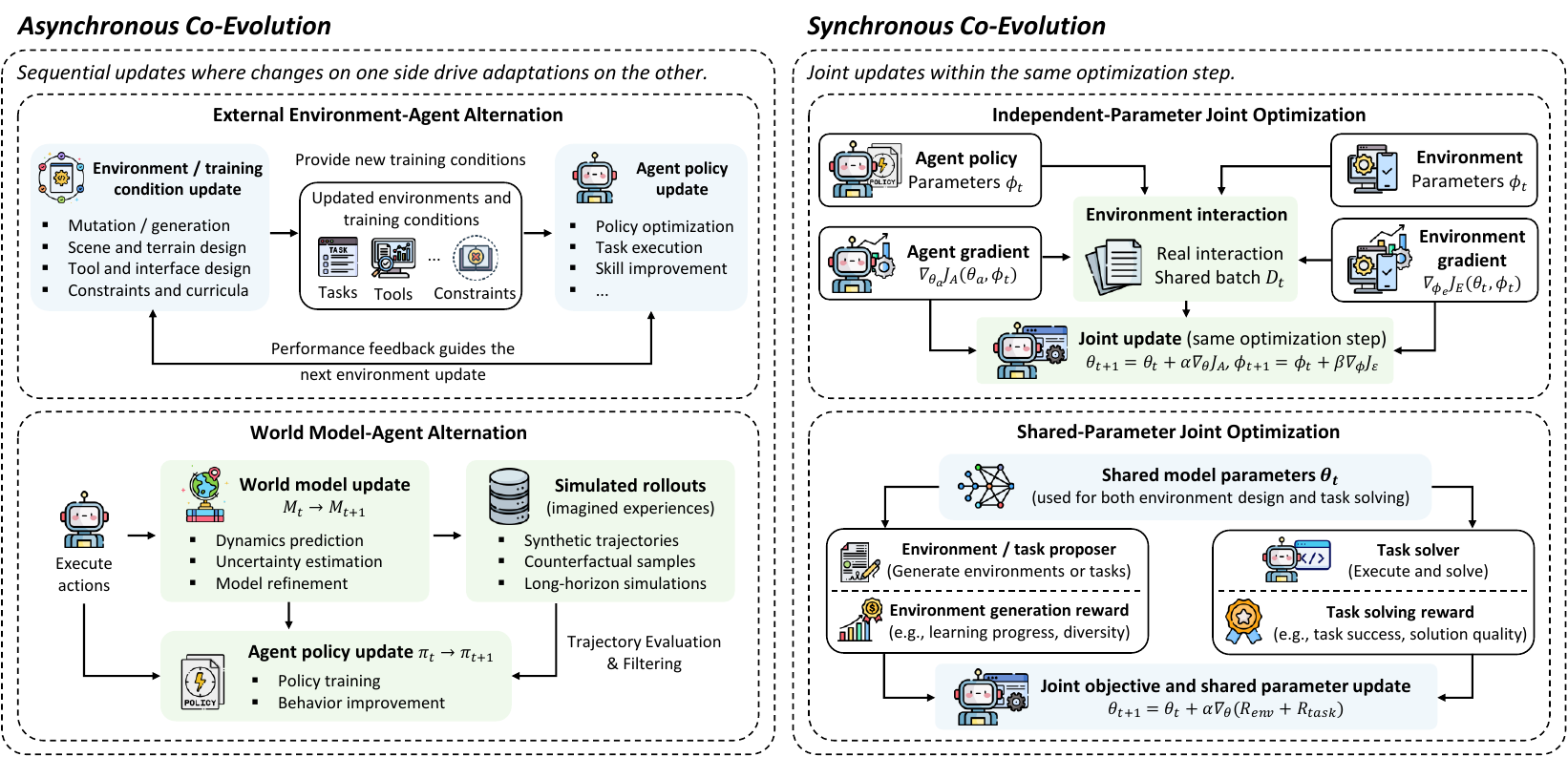}
    \caption{\textbf{Overview of Co-Evolving Agent and Environment.} Asynchronous Co-Evolution sequentially updates agents and environments, allowing changes on one side to drive subsequent adaptations on the other. Synchronous Co-Evolution jointly optimizes agents and environment-side mechanisms, using either independently parameterized or shared-parameter structures. Both mechanisms establish mutually reinforcing learning processes through updates on both sides.}
    \label{fig:co_evolution}
\end{figure*}

Automated feedback can identify deficiencies and guide the improvement of Agents and environments. However, agents and environments are not independent evolutionary targets: improvements in agent capabilities change the range of tasks they can handle, their interaction trajectories, and their state visitation distributions, thereby influencing how environment-side mechanisms generate tasks, adjust difficulty, and update environment models. Conversely, changes in environments and their models alter the interaction experiences, training conditions, and feedback signals available to agents, driving further adaptation. Agent--Environment Co-Evolution therefore concerns the mutually driven and persistent improvement of agents and environment-side mechanisms, rather than the unilateral enhancement of agent capabilities or environment updates alone. Based on how their updates are organized, existing approaches can be categorized into two types, as illustrated in Fig.~\ref{fig:co_evolution}: Asynchronous Co-Evolution, in which Agents and environments are updated sequentially and the outcome of one update guides the subsequent update on the other side; and Synchronous Co-Evolution, in which both sides are jointly updated within the same training cycle through shared interaction experiences or coupled optimization objectives.

\subsection{Asynchronous Co-Evolution}

Asynchronous Co-Evolution sequentially updates agents and environments, making changes on one side the basis for subsequent updates on the other. One approach first uses agent performance and execution experience to adapt environments, then trains agents under the updated conditions. CoEvolve identifies forgetting, capability boundaries, and task performance from interaction trajectories to generate new tasks and refine subsequent training distributions~\cite{yang-etal-2026-coevolve}. 
CoDA further translates policy failures and capability diagnoses into scenario-generation programs targeting current weaknesses~\cite{yu2026codacoevolutiondataengines}. 
R-Zero similarly couples task generation and reasoning through alternating challenger and solver optimization, where newly generated tasks adapt to the solver's evolving capabilities and guide subsequent training~\cite{huang2026rzero}. 
In interactive agent training, EAU and CoEvoLoop use execution trajectories, user feedback, and evaluation outcomes to revise user simulators, environment configurations, or reward mechanisms before subsequent Agent optimization~\cite{cai2026eauenvironmentagentusercoevolutiontraining,cai-etal-2026-coevoloop}. 
From Trainee to Trainer and RLAnything use policy performance to guide environment configurations and task generation, while SEAL leverages fine-grained failure diagnoses to adapt training interfaces and guide policy learning~\cite{traineetotrainer,wang2026rlanythingforgeenvironmentpolicy,seal}. 
This process also extends to world models and autonomous research. DRILL uses policy-learning gains to optimize world models for subsequent training, whereas JANUS incorporates research execution outcomes into environment prediction and Agent improvement~\cite{kang2026drilltrainingworldmodels,janus}. These methods turn evolving Agent capabilities into signals for environment adaptation, which subsequently provides new conditions for further Agent improvement.

Another approach begins with environment updates, allowing agents to learn from modified environments before their performance guides subsequent environmental changes. POET evolves environment–policy pairs through environment mutation, policy optimization, and cross-environment transfer, while LLM-POET and OMNI-EPIC extend learning opportunities through environment generation and task archives~\cite{poet,llmpoet,omniepic}. 
ACCEL and ReMiDi use regret signals to guide environment editing, selection, and training distributions, whereas regret-guided diffusion models generate candidate environments conditioned on policy feedback~\cite{accel,pmlr-v235-beukman24a}. 
In executable environments, COvolve evolves environment and policy code through cross-evaluation, while GenEnv adjusts task generation according to the alignment between task difficulty and solving capabilities~\cite{covolve,guo2025genenv}. 
Explorer-Zero constructs verifiable tasks from intermediate environment states, trains solvers on these tasks, and uses solving outcomes to guide subsequent task generation~\cite{liu2026explorerzeroimprovingselfplaylonghorizon}. 
In world-model-mediated learning, WebEvolver, VLAW, World-VLA-Loop, and WoVR use updated world models to generate simulated experiences for policy training, then refine the models using execution trajectories from improved policies~\cite{webevolver,vlaw,worldvlaloop,jiang2026wovrworldmodelsreliable}. CoMAP further couples future-aware policy reflection with on-policy world-model self-distillation, using predicted action outcomes to improve Agent decisions and subsequent interaction trajectories to update the world model~\cite{comap}. These methods alternate environment updates, Agent learning, and performance feedback, allowing each side to adapt to the other's evolving state.

\subsection{Synchronous Co-Evolution}

Synchronous Co-Evolution jointly updates agents and environment-side mechanisms through shared interaction feedback or coupled learning objectives, allowing both sides to adapt within a unified training process. When the two sides are independently parameterized, joint optimization enables them to adapt to each other through common interaction feedback. In unsupervised environment design, PAIRED constructs regret signals from the return gap between the protagonist and antagonist, guiding the environment generator to produce challenging yet solvable environments around the learner's capability boundary. NCC-UED further formulates the joint optimization of learners and environment adversaries from an optimization perspective~\cite{paired,monette2025optimisation}. In interactive dialogue training, Multi-Agent Dialog Policy Learning models the system and user as two interacting agents and jointly learns their policies through an actor and critic framework~\cite{takanobu-etal-2020-multi}. Cooperative multi-agent reinforcement learning further incorporates conversation knowledge into cooperative policy learning for dialogue management, while ONCE adopts concurrent joint updates of user simulators and dialogue systems to address the distribution shifts associated with conventional alternating training~\cite{lei2020cooperative,chi-etal-2022-offline}. More recently, SCOPE extends this mechanism to multi-turn LLM interactions by jointly training Agent policies against a population of trainable user simulators. Historical simulator checkpoints provide diverse interaction partners, reducing Agent overfitting to a single fixed simulator~\cite{yu2026frozensimulatorenoughsimulator}. SynCo similarly co-trains independently parameterized Synthesizer and Reasoner models, using task-solving outcomes to update both task generation and reasoning capabilities~\cite{yang2026syncodatasynthesiscotraining}.

Another approach jointly evolves agent and environment-side capabilities through shared model parameters, where a single model undertakes both environment construction and task solving, and the corresponding training signals contribute to the same parameter updates. SPADE and EvoEnv adopt this generator--solver paradigm: the shared model acts as an Environment Designer to construct executable training environments and as a Reasoning Agent to solve the generated tasks. In SPADE, hint-based regret guides environment design toward the learner's capability frontier, while task rewards improve the reasoning role; both signals jointly update the shared policy. EvoEnv similarly synthesizes executable Python environments and incorporates staged validation, semantic self-review, solver-relative difficulty calibration, and novelty checks to jointly improve environment construction and task-solving capabilities~\cite{spade,shi2026learningbuildenvironmentselfevolving}. Self-Play SWE-RL extends this mechanism to software engineering, where a single LLM alternates between injecting and repairing increasingly complex bugs in real code repositories. The outcomes of both roles provide reinforcement learning signals for the shared model, allowing generated challenges to evolve alongside software repair capabilities~\cite{pmlr-v306-wei26x}.  Absolute Zero and SPICE both jointly improve task-proposal and task-solving capabilities within a shared Foundation Model, while self-generated or corpus-grounded tasks continually reshape the training curriculum~\cite{zhao2025absolutezero,liu2025spiceselfplaycorpusenvironments}.

%% file: section/rsi.tex
\section{Milestone VII: Approaching Recursive Self-Improvement}
\label{sec::rsi}

\begin{figure*}[t]
    \centering
    \includegraphics[width=\textwidth]{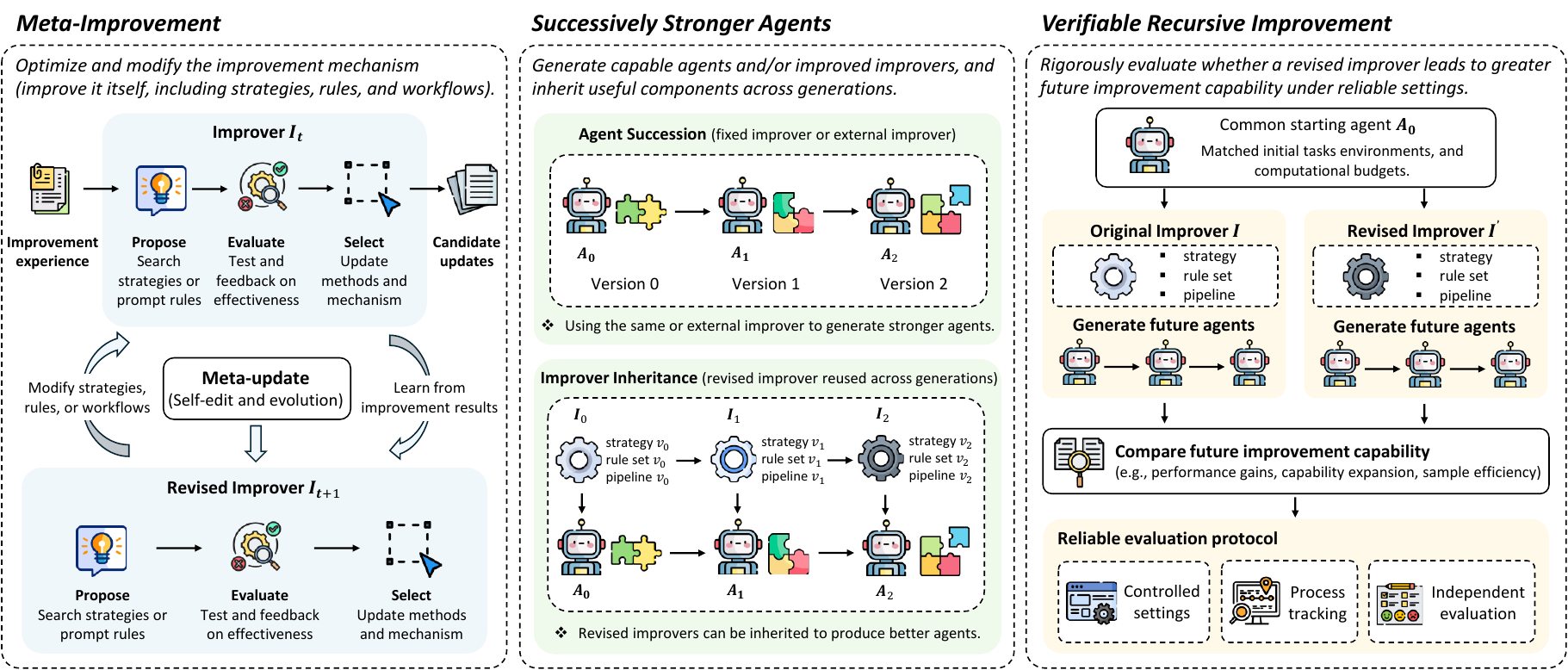}
    \caption{\textbf{Overview of Approaching Recursive Self-Improvement.} Meta-Improvement modifies search strategies, evaluation rules, and update procedures, making improvement mechanisms themselves targets of evolution. Successively Stronger Agents generates and inherits improved successor systems, potentially including revised improvers. Verifiable Recursive Improvement further compares original and revised improvers under common initial conditions and matched resource budgets to determine whether their capabilities to produce future improvements have genuinely increased.}
    \label{fig:rsi}
\end{figure*}

The persistent evolution of models, harnesses, feedback mechanisms, and environments enables agents to improve their capabilities and interaction conditions, yet the mechanisms responsible for producing subsequent improvements may remain unchanged. Recursive self-improvement (RSI) extends the scope of modification to these improvement mechanisms themselves, allowing revised mechanisms to be retained and reused in subsequent generations. As illustrated in Fig.~\ref{fig:rsi}, this section examines the construction, self-modification, and cross-generation evaluation of improvement mechanisms, clarifying the transition from generating stronger successor agents to achieving verifiable gains in recursive improvement capability.

\subsection{Meta-Improvement}

Meta-improvement targets the functions that propose, evaluate, select, and retain system modifications rather than only the components being changed. Frontis-MA1 learns execution-grounded drafting, debugging, improvement, and crossover operations, while POISE automates the proposal and testing of model-training algorithms~\cite{yang2026frontisma1,xia2026poise}. Both strengthen the production of changes without requiring the outer improvement procedure itself to change. Self-referential approaches go further: STOP edits its own code-improvement program, while Promptbreeder co-evolves mutation prompts with the task prompts they modify~\cite{zelikman2024stop,fernando2024promptbreeder}. These studies distinguish constructing an effective improver from making its modification rules editable.

Beyond individual modification rules, search strategies can adapt from accumulated outcomes. EvoX evolves selection and variation strategies, AEvo revises procedural guidance using evolution histories, and Bilevel Autoresearch uses an outer research loop to change an inner loop's search mechanism~\cite{liu2026evox,zhang2026aevo,qu2026bilevel}. Dream-RSI optimizes exploration policies using replayed discovery trees, whereas Mixture of Self-Improving Branches updates branch-specific proposal guidance and development subsets~\cite{zheng2026dreamrsi,dong2026mixture}. These approaches differ in which decisions are revised and which outer controls remain fixed: stronger search results alone do not establish that the search mechanism itself has evolved.

The editable scope can further extend from individual search decisions to the mechanism coordinating improvement. MetaSkill-Evolve updates guidance for analysis, retrieval, allocation, proposal, and editing on a slower timescale than task skills~\cite{wang2026metaskillevolve}. Hyperagents makes task-agent and meta-agent logic jointly editable, allowing the agent-generation procedure itself to be revised~\cite{zhang2026hyperagents}. The former evolves a structured improvement pipeline, while the latter exposes the meta-agent program; neither implies that every surrounding evaluation or acceptance rule is modifiable. Such revisions contribute to recursive improvement when they are retained and executed to produce later successors.

\subsection{Successively Stronger Agents}

Generating successive agent versions provides a foundation for recursive improvement, but inheriting stronger task-solving capabilities does not necessarily imply inheriting a stronger improvement mechanism. Existing approaches support successive improvement by retaining modified agent programs, archives of evaluated candidates, or reusable search experience for subsequent generations. G{\"o}del Agent revises its operating logic, SICA archives evaluated agent versions for further editing, and Ouroboros incorporates reviewed core-code changes into later runtime versions~\cite{yin2025goedelagent,robeyns2025sica,razzhigaev2026ouroboros}. Open-ended search broadens this inheritance path: DGM archives alternative agents as possible ancestors, MGM uses cross-task and cross-lineage evidence to guide modifications, and SelfSearch reuses records of earlier self-modification episodes~\cite{zhang2026darwin,liu2026mendel,yang2026selfsearch}. These systems retain executable candidates or search experience, even when their outer search and acceptance rules remain fixed. A useful lineage therefore need not improve monotonically at every intermediate generation.

Inheriting an improvement mechanism requires later generations to execute a revised procedure, not merely preserve a higher-scoring task agent. Hierarchical Self-Improvement makes an evolver's strategy editable by a meta-evolver while retaining a fixed outer anchor~\cite{zhou2026hsi}. Escher-Loop instead maintains an evolving optimizer population whose revised members help generate future task agents and optimizers~\cite{liu2026escher}. SpecRSI combines verification-aware successor search with updates to its improver from completed search outcomes, accepting only verified revisions~\cite{xu2026specrsi}. These designs preserve changes to controllers, optimizers, or search policies that shape subsequent modifications, while leaving explicit boundaries around the remaining fixed components.

In autonomous AI research, inheritance may span code, training choices, and policies coordinating multiple improvement operations. MetaRSI-v1 retains system versions together with policies governing data, harness, and model updates~\cite{tan2026metarsi}. AIDE$^{2}$ accumulates accepted inner research-agent rewrites, but its outer improver remains fixed during the main evolutionary run~\cite{srikanth2026aide2}. The contrast shows why successive gains in research-agent performance do not by themselves demonstrate the self-improvement of the mechanism producing them. Even when a revised improver is reused, whether it can generate greater future gains must be established separately.

\subsection{Verifiable Recursive Improvement}

Generating stronger successors does not establish that the revised improver is better at producing future gains. Verifiable RSI therefore evaluates improver capability through controlled comparisons, rather than relying only on current task performance. HGM shows that present agent scores can diverge from descendant productivity and uses estimates of lineage potential to guide exploration~\cite{wang2025huxley}. Such estimates depend on the search process that produces descendants. Hyperagents evaluates the improver more directly through improvement@k, measuring the test-performance gain of a validation-selected successor from a common starting agent within a fixed candidate budget and with a frozen meta-agent~\cite{zhang2026hyperagents}. This makes successor-generation effectiveness an explicit evaluation target, although equal candidate counts do not necessarily imply equal inference and verification costs.

Controlled continuations further test whether a revised improver produces greater future gains. From the same starting system and under matched continuation budgets, MetaRSI-v1 reports second-term improvements of 7.2 points with a meta-updated improver versus 4.9 without the update; five-term cumulative gains reach 26.2 versus 20.7, while marginal gains decline along both paths~\cite{tan2026metarsi}. Meta-update costs are counted separately. AIDE$^{2}$ compares a discovered research agent with the original outer improver from a common inner-agent starting point; three 50-step runs per arm end at approximately 0.780 and 0.782, yielding an inconclusive comparison~\cite{srikanth2026aide2}. These results provide setting-specific evidence, not proof of universally stronger improvers or indefinitely accelerating recursion.

Credible cross-generation comparison also requires reliable update acceptance, comparable resources, and independent assessment. SGM controls erroneous commitments under adaptive testing; SAGE applies paired, regression-aware statistical acceptance; and anytime-valid certificates support auditable acceptance of constrained self-modifications~\cite{wu2025sgm,sage2026,sea2026}. Analyses of harness self-evolution distinguish finding useful changes from reliably validating them~\cite{cai2026reliable}. FML-Bench examines search dynamics under comparable infrastructure, EvoPathBench tracks successive artifact checkpoints, and RRSI and ModularRSI test generalization outside evolution tasks~\cite{zou2026fmlbench,evopathbench2026,xia2026rrsi,wu2026modularrsi}. When internal evaluators themselves evolve, RQGM and Who Grades the Grader highlight the need for independent evaluation anchors~\cite{iacob2026redqueen,zhang2026whogrades}. These methods strengthen the reliability of measured gains, but verified task updates or generalization alone cannot establish sustained growth in future improvement capability.

%% file: section/benchmark.tex
\section{Benchmarks and Evaluation}
\label{sec:benchmarks}

\usetikzlibrary{svg.path}
\ifdefined\SIBenchmarkTableBox\else\newsavebox{\SIBenchmarkTableBox}\fi
\ifdefined\SIBenchmarkCaptionBox\else\newsavebox{\SIBenchmarkCaptionBox}\fi
\ifdefined\SIBenchmarkTableWidth\else\newlength{\SIBenchmarkTableWidth}\fi
\ifdefined\SIBenchmarkTableMaxHeight\else\newlength{\SIBenchmarkTableMaxHeight}\fi
\ifdefined\SIBenchmarkRowGap\else\newlength{\SIBenchmarkRowGap}\fi
\ifdefined\SIBenchmarkFontSize\else\newlength{\SIBenchmarkFontSize}\fi
\ifdefined\SIBenchmarkLeading\else\newlength{\SIBenchmarkLeading}\fi
\ifdefined\SIBenchmarkGitHubGlyph\else\newsavebox{\SIBenchmarkGitHubGlyph}\fi
\ifdefined\SIBenchmarkHFGlyph\else\newsavebox{\SIBenchmarkHFGlyph}\fi
\ifdefined\SIBenchmarkPaperGlyph\else\newsavebox{\SIBenchmarkPaperGlyph}\fi
\begin{table*}[!tp]
\centering
\begingroup
\setlength{\SIBenchmarkTableWidth}{\textwidth}
\normalfont\mdseries\upshape
\linespread{1}\selectfont
\setlength{\parskip}{0pt}
\begin{lrbox}{\SIBenchmarkCaptionBox}
\begin{minipage}{\SIBenchmarkTableWidth}
\normalfont\mdseries\upshape
\linespread{1}\selectfont
\setlength{\parskip}{0pt}
\caption{Representative benchmarks and evaluation protocols categorized by What Improves and Who Drives Improve. F1--F4 denote Human-Steered, Environment-Grounded, AI-Facilitated, and Evolving Feedback. T/C/G indicate task-level, cross-task, and cross-generation evaluation, respectively. Cross-generation evaluation does not necessarily establish stronger improvers. \textsuperscript{\ensuremath{\dagger}}denotes study-specific evaluation protocols. Links point to code, data, or primary papers.}
\label{tab:rsi-benchmarks}
\end{minipage}
\end{lrbox}
\setlength{\SIBenchmarkTableMaxHeight}{\dimexpr
  \textheight-\ht\SIBenchmarkCaptionBox-\dp\SIBenchmarkCaptionBox-14pt\relax}
\setlength{\SIBenchmarkRowGap}{0pt}
\providecommand{\SIGitHubIcon}{%
  \raisebox{-0.10ex}{\tikz[x=1pt,y=1pt,scale=0.30,yscale=-1,baseline=(current bounding box.south)]{%
    \path[use as bounding box] (0,0) rectangle (24,24);%
    \fill svg {M12 .297c-6.63 0-12 5.373-12 12 0 5.303 3.438 9.8 8.205 11.385.6.113.82-.258.82-.577 0-.285-.01-1.04-.015-2.04-3.338.724-4.042-1.61-4.042-1.61C4.422 18.07 3.633 17.7 3.633 17.7c-1.087-.744.084-.729.084-.729 1.205.084 1.838 1.236 1.838 1.236 1.07 1.835 2.809 1.305 3.495.998.108-.776.417-1.305.76-1.605-2.665-.3-5.466-1.332-5.466-5.93 0-1.31.465-2.38 1.235-3.22-.135-.303-.54-1.523.105-3.176 0 0 1.005-.322 3.3 1.23.96-.267 1.98-.399 3-.405 1.02.006 2.04.138 3 .405 2.28-1.552 3.285-1.23 3.285-1.23.645 1.653.24 2.873.12 3.176.765.84 1.23 1.91 1.23 3.22 0 4.61-2.805 5.625-5.475 5.92.42.36.81 1.096.81 2.22 0 1.606-.015 2.896-.015 3.286 0 .315.21.69.825.57C20.565 22.092 24 17.592 24 12.297c0-6.627-5.373-12-12-12};%
  }}%
}
\definecolor{SIHF3A3B45}{HTML}{3A3B45}
\definecolor{SIHFFF323D}{HTML}{FF323D}
\definecolor{SIHFFF9D0B}{HTML}{FF9D0B}
\definecolor{SIHFFFD21E}{HTML}{FFD21E}
\providecommand{\SIHFIcon}{%
  \raisebox{-0.10ex}{\tikz[x=1pt,y=1pt,scale=0.090,yscale=-1,baseline=(current bounding box.south)]{%
    \path[use as bounding box] (0,0) rectangle (95,88);%
    \fill[SIHFFFD21E] svg {M47.21 76.5a34.75 34.75 0 1 0 0-69.5 34.75 34.75 0 0 0 0 69.5Z};%
    \fill[SIHFFF9D0B] svg {M81.96 41.75a34.75 34.75 0 1 0-69.5 0 34.75 34.75 0 0 0 69.5 0Zm-73.5 0a38.75 38.75 0 1 1 77.5 0 38.75 38.75 0 0 1-77.5 0Z};%
    \fill[SIHF3A3B45] svg {M58.5 32.3c1.28.44 1.78 3.06 3.07 2.38a5 5 0 1 0-6.76-2.07c.61 1.15 2.55-.72 3.7-.32ZM34.95 32.3c-1.28.44-1.79 3.06-3.07 2.38a5 5 0 1 1 6.76-2.07c-.61 1.15-2.56-.72-3.7-.32Z};%
    \fill[SIHFFF323D] svg {M46.96 56.29c9.83 0 13-8.76 13-13.26 0-2.34-1.57-1.6-4.09-.36-2.33 1.15-5.46 2.74-8.9 2.74-7.19 0-13-6.88-13-2.38s3.16 13.26 13 13.26Z};%
    \fill[SIHF3A3B45,even odd rule] svg {M39.43 54a8.7 8.7 0 0 1 5.3-4.49c.4-.12.81.57 1.24 1.28.4.68.82 1.37 1.24 1.37.45 0 .9-.68 1.33-1.35.45-.7.89-1.38 1.32-1.25a8.61 8.61 0 0 1 5 4.17c3.73-2.94 5.1-7.74 5.1-10.7 0-2.34-1.57-1.6-4.09-.36l-.14.07c-2.31 1.15-5.39 2.67-8.77 2.67s-6.45-1.52-8.77-2.67c-2.6-1.29-4.23-2.1-4.23.29 0 3.05 1.46 8.06 5.47 10.97Z};%
    \fill[SIHFFF9D0B] svg {M70.71 37a3.25 3.25 0 1 0 0-6.5 3.25 3.25 0 0 0 0 6.5ZM24.21 37a3.25 3.25 0 1 0 0-6.5 3.25 3.25 0 0 0 0 6.5ZM17.52 48c-1.62 0-3.06.66-4.07 1.87a5.97 5.97 0 0 0-1.33 3.76 7.1 7.1 0 0 0-1.94-.3c-1.55 0-2.95.59-3.94 1.66a5.8 5.8 0 0 0-.8 7 5.3 5.3 0 0 0-1.79 2.82c-.24.9-.48 2.8.8 4.74a5.22 5.22 0 0 0-.37 5.02c1.02 2.32 3.57 4.14 8.52 6.1 3.07 1.22 5.89 2 5.91 2.01a44.33 44.33 0 0 0 10.93 1.6c5.86 0 10.05-1.8 12.46-5.34 3.88-5.69 3.33-10.9-1.7-15.92-2.77-2.78-4.62-6.87-5-7.77-.78-2.66-2.84-5.62-6.25-5.62a5.7 5.7 0 0 0-4.6 2.46c-1-1.26-1.98-2.25-2.86-2.82A7.4 7.4 0 0 0 17.52 48Zm0 4c.51 0 1.14.22 1.82.65 2.14 1.36 6.25 8.43 7.76 11.18.5.92 1.37 1.31 2.14 1.31 1.55 0 2.75-1.53.15-3.48-3.92-2.93-2.55-7.72-.68-8.01.08-.02.17-.02.24-.02 1.7 0 2.45 2.93 2.45 2.93s2.2 5.52 5.98 9.3c3.77 3.77 3.97 6.8 1.22 10.83-1.88 2.75-5.47 3.58-9.16 3.58-3.81 0-7.73-.9-9.92-1.46-.11-.03-13.45-3.8-11.76-7 .28-.54.75-.76 1.34-.76 2.38 0 6.7 3.54 8.57 3.54.41 0 .7-.17.83-.6.79-2.85-12.06-4.05-10.98-8.17.2-.73.71-1.02 1.44-1.02 3.14 0 10.2 5.53 11.68 5.53.11 0 .2-.03.24-.1.74-1.2.33-2.04-4.9-5.2-5.21-3.16-8.88-5.06-6.8-7.33.24-.26.58-.38 1-.38 3.17 0 10.66 6.82 10.66 6.82s2.02 2.1 3.25 2.1c.28 0 .52-.1.68-.38.86-1.46-8.06-8.22-8.56-11.01-.34-1.9.24-2.85 1.31-2.85Z};%
    \fill[SIHFFFD21E] svg {M38.6 76.69c2.75-4.04 2.55-7.07-1.22-10.84-3.78-3.77-5.98-9.3-5.98-9.3s-.82-3.2-2.69-2.9c-1.87.3-3.24 5.08.68 8.01 3.91 2.93-.78 4.92-2.29 2.17-1.5-2.75-5.62-9.82-7.76-11.18-2.13-1.35-3.63-.6-3.13 2.2.5 2.79 9.43 9.55 8.56 11-.87 1.47-3.93-1.71-3.93-1.71s-9.57-8.71-11.66-6.44c-2.08 2.27 1.59 4.17 6.8 7.33 5.23 3.16 5.64 4 4.9 5.2-.75 1.2-12.28-8.53-13.36-4.4-1.08 4.11 11.77 5.3 10.98 8.15-.8 2.85-9.06-5.38-10.74-2.18-1.7 3.21 11.65 6.98 11.76 7.01 4.3 1.12 15.25 3.49 19.08-2.12Z};%
    \fill[SIHFFF9D0B] svg {M77.4 48c1.62 0 3.07.66 4.07 1.87a5.97 5.97 0 0 1 1.33 3.76 7.1 7.1 0 0 1 1.95-.3c1.55 0 2.95.59 3.94 1.66a5.8 5.8 0 0 1 .8 7 5.3 5.3 0 0 1 1.78 2.82c.24.9.48 2.8-.8 4.74a5.22 5.22 0 0 1 .37 5.02c-1.02 2.32-3.57 4.14-8.51 6.1-3.08 1.22-5.9 2-5.92 2.01a44.33 44.33 0 0 1-10.93 1.6c-5.86 0-10.05-1.8-12.46-5.34-3.88-5.69-3.33-10.9 1.7-15.92 2.78-2.78 4.63-6.87 5.01-7.77.78-2.66 2.83-5.62 6.24-5.62a5.7 5.7 0 0 1 4.6 2.46c1-1.26 1.98-2.25 2.87-2.82A7.4 7.4 0 0 1 77.4 48Zm0 4c-.51 0-1.13.22-1.82.65-2.13 1.36-6.25 8.43-7.76 11.18a2.43 2.43 0 0 1-2.14 1.31c-1.54 0-2.75-1.53-.14-3.48 3.91-2.93 2.54-7.72.67-8.01a1.54 1.54 0 0 0-.24-.02c-1.7 0-2.45 2.93-2.45 2.93s-2.2 5.52-5.97 9.3c-3.78 3.77-3.98 6.8-1.22 10.83 1.87 2.75 5.47 3.58 9.15 3.58 3.82 0 7.73-.9 9.93-1.46.1-.03 13.45-3.8 11.76-7-.29-.54-.75-.76-1.34-.76-2.38 0-6.71 3.54-8.57 3.54-.42 0-.71-.17-.83-.6-.8-2.85 12.05-4.05 10.97-8.17-.19-.73-.7-1.02-1.44-1.02-3.14 0-10.2 5.53-11.68 5.53-.1 0-.19-.03-.23-.1-.74-1.2-.34-2.04 4.88-5.2 5.23-3.16 8.9-5.06 6.8-7.33-.23-.26-.57-.38-.98-.38-3.18 0-10.67 6.82-10.67 6.82s-2.02 2.1-3.24 2.1a.74.74 0 0 1-.68-.38c-.87-1.46 8.05-8.22 8.55-11.01.34-1.9-.24-2.85-1.31-2.85Z};%
    \fill[SIHFFFD21E] svg {M56.33 76.69c-2.75-4.04-2.56-7.07 1.22-10.84 3.77-3.77 5.97-9.3 5.97-9.3s.82-3.2 2.7-2.9c1.86.3 3.23 5.08-.68 8.01-3.92 2.93.78 4.92 2.28 2.17 1.51-2.75 5.63-9.82 7.76-11.18 2.13-1.35 3.64-.6 3.13 2.2-.5 2.79-9.42 9.55-8.55 11 .86 1.47 3.92-1.71 3.92-1.71s9.58-8.71 11.66-6.44c2.08 2.27-1.58 4.17-6.8 7.33-5.23 3.16-5.63 4-4.9 5.2.75 1.2 12.28-8.53 13.36-4.4 1.08 4.11-11.76 5.3-10.97 8.15.8 2.85 9.05-5.38 10.74-2.18 1.69 3.21-11.65 6.98-11.76 7.01-4.31 1.12-15.26 3.49-19.08-2.12Z};%
  }}%
}
\providecommand{\SIPaperIcon}{%
  \raisebox{-0.10ex}{\tikz[x=1pt,y=1pt,scale=0.30,yscale=-1,baseline=(current bounding box.south)]{%
    \path[use as bounding box] (0,0) rectangle (24,24);%
    \draw[line width=0.55pt,line join=round] (5,2)--(14,2)--(20,8)--(20,22)--(5,22)--cycle;%
    \draw[line width=0.55pt] (14,2)--(14,8)--(20,8);%
    \draw[line width=0.45pt] (8,12)--(17,12) (8,15)--(17,15) (8,18)--(15,18);%
  }}%
}
\providecommand{\SIGitHub}[1]{\href{#1}{\SIGitHubIcon}}
\providecommand{\SIHF}[1]{\href{#1}{\SIHFIcon}}
\providecommand{\SIPaper}[1]{\href{#1}{\SIPaperIcon}}

\sbox{\SIBenchmarkGitHubGlyph}{{\normalfont\fontsize{7.6}{8.6}\selectfont\SIGitHubIcon}}
\sbox{\SIBenchmarkHFGlyph}{{\normalfont\fontsize{7.6}{8.6}\selectfont\SIHFIcon}}
\sbox{\SIBenchmarkPaperGlyph}{{\normalfont\fontsize{7.6}{8.6}\selectfont\SIPaperIcon}}
\renewcommand{\SIGitHubIcon}{\usebox{\SIBenchmarkGitHubGlyph}}
\renewcommand{\SIHFIcon}{\usebox{\SIBenchmarkHFGlyph}}
\renewcommand{\SIPaperIcon}{\usebox{\SIBenchmarkPaperGlyph}}

\def\SIBenchmarkTypeset{%
\begin{lrbox}{\SIBenchmarkTableBox}
\begin{minipage}{\SIBenchmarkTableWidth}
\normalfont\mdseries\upshape
\linespread{1}\selectfont
\let\bffont\relax 
\fontsize{\the\SIBenchmarkFontSize}{\the\SIBenchmarkLeading}\selectfont
\setlength{\parskip}{0pt}
\setlength{\tabcolsep}{1.7pt}
\setlength{\extrarowheight}{0pt}
\setlength{\aboverulesep}{1.2pt}
\setlength{\belowrulesep}{1.2pt}
\setlength{\abovetopsep}{0pt}
\setlength{\belowbottomsep}{0pt}
\renewcommand{\arraystretch}{1.00}
\begin{tabularx}{\SIBenchmarkTableWidth}{@{}
>{\raggedright\arraybackslash}p{0.230\SIBenchmarkTableWidth}
>{\raggedright\arraybackslash}p{0.110\SIBenchmarkTableWidth}
>{\centering\arraybackslash}p{36pt}@{\hspace{6pt}}
>{\raggedright\arraybackslash}X
>{\centering\arraybackslash}p{24pt}
>{\centering\arraybackslash}p{22pt}
@{}}
\toprule
\textbf{Benchmark / Study} & \shortstack{\textbf{What}\\\textbf{Improves}} & \shortstack{\textbf{Who Drives} \\\textbf{Improve}} & \textbf{Evaluated Capabilities} & \textbf{Scope} & \textbf{Link} \\[\SIBenchmarkRowGap]
\midrule
MLAgentBench~\cite{huang2024mlagentbench}
& Model & F2 & ML experiment planning and optimization. & T & \SIGitHub{https://github.com/snap-stanford/MLAgentBench} \\[\SIBenchmarkRowGap]
MLE-bench~\cite{chan2025mlebench}
& Model & F2 & Competition performance on held-out data. & T & \SIGitHub{https://github.com/openai/mle-bench} \\[\SIBenchmarkRowGap]
MLGym~\cite{nathani2025mlgym}
& Model & F2 & Algorithm and training changes on ML tasks. & T & \SIGitHub{https://github.com/facebookresearch/MLGym} \\[\SIBenchmarkRowGap]
PaperBench~\cite{starace2025paperbench}
& Model & F3 & Research replication under AI-scored rubrics. & T & \SIPaper{https://arxiv.org/abs/2504.01848v3} \\[\SIBenchmarkRowGap]

PostTrainBench~\cite{rank2026posttrainbench}
& Model & F2 & Post-training gains within resource limits. & T & \SIGitHub{https://github.com/aisa-group/PostTrainBench}\,\,\SIHF{https://huggingface.co/datasets/aisa-group/PostTrainBench-Trajectories} \\[\SIBenchmarkRowGap]
RSIBench-Data~\cite{evolvent2026rsibenchdata}
& Model & F2 & Data-revision gains; shared selection/evaluation tasks. & T & \SIGitHub{https://github.com/evolvent-ai/RSIBench-Data} \\[\SIBenchmarkRowGap]
S\textsuperscript{3}Gym~\cite{shi2026s3gym}
& Model & F2/F3 & Self-tests, judgment reliability, experience reuse. & T/C & \SIPaper{https://arxiv.org/abs/2608.31100} \\[\SIBenchmarkRowGap]
InstructGPT\textsuperscript{\ensuremath{\dagger}}~\cite{instructgpt}
& Model & F1 & Alignment following human-preference training. & T & \SIPaper{https://proceedings.neurips.cc/paper_files/paper/2022/hash/b1efde53be364a73914f58805a001731-Abstract.html} \\[\SIBenchmarkRowGap]
EvoRubric\textsuperscript{\ensuremath{\dagger}}~\cite{evo_rubric}
& Model & F4 & Response quality and evolving criteria. & T & \SIPaper{https://arxiv.org/abs/2605.29847v2} \\[\SIBenchmarkRowGap]
\midrule
HarnessOpt-Bench~\cite{ursekar2026harnessoptbench}
& Harness & F2 & Held-out gains within optimization budgets. & T & \SIPaper{https://arxiv.org/abs/2608.06301} \\[\SIBenchmarkRowGap]
Evo-Bench~\cite{huang2026evobench}
& Harness & F2/F3 & Optimization gains and cross-setting transfer. & C & \SIGitHub{https://github.com/RUCAIBox/Evo-Bench}\,\,\SIHF{https://huggingface.co/datasets/RUC-AIBOX/Evo-Bench} \\[\SIBenchmarkRowGap]
HarnessDev~\cite{wu2026harnessdev}
& Harness & F2 & Evolution-split refinement; held-out task gains. & C & \SIPaper{https://arxiv.org/abs/2609.01437} \\[\SIBenchmarkRowGap]
Doc2Harness~\cite{ge2026doc2harness}
& Harness & F2 & Specification-to-harness executable conformance. & T & \SIPaper{https://arxiv.org/abs/2610.20344} \\[\SIBenchmarkRowGap]
EvoHarnessBench~\cite{ke2026evoharnessbench}
& Harness & F2 & Retention under externally expanded harnesses. & C & \SIHF{https://huggingface.co/collections/ZixuanKe/evoharnessbench} \\[\SIBenchmarkRowGap]
SkillsBench~\cite{li2026skillsbench}
& Harness & F2 & Paired skill-use versus no-skill performance. & T & \SIGitHub{https://github.com/benchflow-ai/skillsbench}\,\,\SIHF{https://huggingface.co/datasets/benchflow/skillsbench} \\[\SIBenchmarkRowGap]
SkillScriptBench~\cite{liu2026skillscriptbench}
& Harness & F2 & Script repair, documentation, behavior preservation. & T & \SIPaper{https://arxiv.org/abs/2610.04008v2} \\[\SIBenchmarkRowGap]

SkillLearnBench~\cite{zhong2026skilllearnbench}
& Harness & F2/F3 & Skill quality, execution, and task outcomes. & T & \SIGitHub{https://github.com/cxcscmu/SkillLearnBench} \\[\SIBenchmarkRowGap]
SkillEvolBench~\cite{lei2026skillevolbench}
& Harness & F2 & Procedural transfer to unseen tasks. & C & \SIGitHub{https://github.com/AIoT-MLSys-Lab/SkillEvolBench} \\[\SIBenchmarkRowGap]
SkillFlow~\cite{zhang2026skillflow}
& Harness & F2 & Verified skill revision and downstream reuse. & C & \SIGitHub{https://github.com/ZhangZi-a/SkillFlow}\,\,\SIHF{https://huggingface.co/datasets/zhang-ziao/SkillFlow-Task} \\[\SIBenchmarkRowGap]
ContinualSkillBench~\cite{guan2026continualskillbench}
& Harness & F2/F3 & Skill accumulation and retention across streams. & C & \SIGitHub{https://github.com/gtynnn060110-hash/continual-skill-bench-final} \\[\SIBenchmarkRowGap]
Tool-Genesis~\cite{xia2026toolgenesis}
& Harness & F2 & Generated-tool correctness and utility. & T & \SIGitHub{https://github.com/Tool-Genesis/Tool-Genesis}\,\,\SIHF{https://huggingface.co/datasets/tool-genesis/Tool-Genesis-Benchmark} \\[\SIBenchmarkRowGap]
EvoMemBench~\cite{wang2026evomembench}
& Harness & F2 & Memory use within and across interactions. & T/C & \SIGitHub{https://github.com/DSAIL-Memory/EvoMemBench} \\[\SIBenchmarkRowGap]
Evo-Memory~\cite{wei2025evomemory}
& Harness & F2/F3 & Memory updates and sequential-task reuse. & C & \SIPaper{https://arxiv.org/abs/2511.20857} \\[\SIBenchmarkRowGap]
MemoryArena~\cite{he2026memoryarena}
& Harness & F2 & Task success across interdependent sessions. & C & \SIGitHub{https://github.com/ZexueHe/MemoryArena}\,\,\SIHF{https://huggingface.co/datasets/ZexueHe/memoryarena} \\[\SIBenchmarkRowGap]
MemDerive~\cite{gao2026memderive}
& Harness & F2 & Derived-state dependencies and update propagation. & C & \SIPaper{https://arxiv.org/abs/2610.25373v1} \\[\SIBenchmarkRowGap]

SCLATE~\cite{jung2026sclate}
& Harness & F2/F3 & Cross-session performance under different memory configurations. & C & \SIPaper{https://arxiv.org/abs/2609.32391v2} \\[\SIBenchmarkRowGap]

PAST-Bench~\cite{xue2026pastbench}
& Harness & F2 & Paired retained-experience versus reset runs. & C & \SIGitHub{https://github.com/Gen-Verse/PAST-Bench} \\[\SIBenchmarkRowGap]
EvoAgentBench~\cite{gao2026evoagentbench}
& Harness & F2/F3 & Unseen-task transfer versus no-evolution control. & C & \SIGitHub{https://github.com/EverMind-AI/EvoAgentBench}\,\,\SIHF{https://huggingface.co/datasets/EverMind-AI/EvoAgentBench} \\[\SIBenchmarkRowGap]
FinEvo-Bench~\cite{deng2026finevo}
& Harness & F3 & Professional workflows versus paired state resets. & C & \SIPaper{https://arxiv.org/abs/2608.06144v2} \\[\SIBenchmarkRowGap]

FinEvolveBench~\cite{zhu2026finevolvebench}
& Harness & F2 & Delayed, noisy outcomes with limited experience reuse. & C & \SIPaper{https://arxiv.org/abs/2606.06960v4} \\[\SIBenchmarkRowGap]

SEAGym~\cite{zheng2026seagym}
& Harness & F2 & Checkpoint gains, forgetting, and costs. & C & \SIGitHub{https://github.com/antropy-research/SEAGym} \\[\SIBenchmarkRowGap]
SEABench~\cite{das2026seabench}
& Harness & F3 & Safety regression versus non-evolving controls. & C & \SIPaper{https://arxiv.org/abs/2609.35596v2} \\[\SIBenchmarkRowGap]

EvoPathBench~\cite{evopathbench2026}
& Harness & F2 & Transfer and retention under changing rules. & C & \SIGitHub{https://github.com/HQ-Lin/EvoPathBench} \\[\SIBenchmarkRowGap]
ServeLearnBench~\cite{zheng2026servelearnbench}
& Harness & F2/F3 & Outcome-guided adaptation to hidden rule changes. & C & \SIGitHub{https://github.com/Infini-AI-Lab/ServeLearnBench} \\[\SIBenchmarkRowGap]
RSE-Bench~\cite{yu2026rsebench}
& Harness & F2 & Skill resilience to corrupted learning signals. & C & \SIPaper{https://arxiv.org/abs/2610.19910} \\[\SIBenchmarkRowGap]
PAHF\textsuperscript{\ensuremath{\dagger}}~\cite{liang2026pahf}
& Harness & F3 & Adaptation to AI-simulated user preferences. & C & \SIGitHub{https://github.com/facebookresearch/PAHF} \\[\SIBenchmarkRowGap]
Meta-Agent Challenge~\cite{lu2026metaagentchallenge}
& Harness & F2 & Generated-agent quality and development cost. & T & \SIGitHub{https://github.com/ant-research/meta-agent-challenge} \\[\SIBenchmarkRowGap]
HGM\textsuperscript{\ensuremath{\dagger}}~\cite{wang2025huxley}
& Harness & F2 & Lineage potential via Clade-Metaproductivity. & C/G & \SIGitHub{https://github.com/metauto-ai/HGM} \\[\SIBenchmarkRowGap]
\midrule
EnvHarness\textsuperscript{\ensuremath{\dagger}}~\cite{envharness}
& Environment & F3 & Learning gains from AI-revised interfaces. & C & \SIGitHub{https://github.com/google-research/envharness} \\[\SIBenchmarkRowGap]
POET\textsuperscript{\ensuremath{\dagger}}~\cite{poet}
& Environment & F2 & Challenge diversity and post-adaptation transfer. & C & \SIGitHub{https://github.com/uber-research/poet} \\[\SIBenchmarkRowGap]
PAIRED\textsuperscript{\ensuremath{\dagger}}~\cite{paired}
& Environment & F2 & Generalization after regret-guided task design. & C & \SIGitHub{https://github.com/google-research/google-research/tree/master/social_rl/adversarial_env} \\[\SIBenchmarkRowGap]
ACCEL\textsuperscript{\ensuremath{\dagger}}~\cite{accel}
& Environment & F2 & Zero-shot transfer from evolving curricula. & C & \SIGitHub{https://github.com/facebookresearch/dcd} \\[\SIBenchmarkRowGap]
GenEnv\textsuperscript{\ensuremath{\dagger}}~\cite{guo2025genenv}
& Environment & F2 & Learning gains and generated-data efficiency. & C & \SIGitHub{https://github.com/Gen-Verse/GenEnv} \\[\SIBenchmarkRowGap]
\midrule
AI4AI-Bench~\cite{chi2026ai4aibench}
& Improver & F2 & Algorithm utility after independent retraining. & T & \SIGitHub{https://github.com/Einsia/AI4AI-Bench} \\[\SIBenchmarkRowGap]
RSI-Exam~\cite{rsiexam2026}
& Improver & F2 & Revised-method gains under sealed replay. & T & \SIGitHub{https://github.com/aiming-lab/RSI-Exam}\,\,\SIHF{https://huggingface.co/datasets/RSI-Exam/RSI-Exam} \\[\SIBenchmarkRowGap]
STOP\textsuperscript{\ensuremath{\dagger}}~\cite{zelikman2024stop}
& Improver & F2 & Downstream utility of revised improver code. & T/G & \SIGitHub{https://github.com/microsoft/stop} \\[\SIBenchmarkRowGap]
Hyperagents\textsuperscript{\ensuremath{\dagger}}~\cite{zhang2026hyperagents}
& Improver & F2 & Candidate-generation utility at fixed budgets. & C/G & \SIGitHub{https://github.com/facebookresearch/HyperAgents} \\[\SIBenchmarkRowGap]
AIDE\textsuperscript{2}\textsuperscript{\ensuremath{\dagger}}~\cite{srikanth2026aide2}
& Improver & F2 & Later gains from frozen research-agent revisions. & T/G & \SIPaper{https://arxiv.org/abs/2609.26457} \\[\SIBenchmarkRowGap]

MetaRSI\textsuperscript{\ensuremath{\dagger}}~\cite{tan2026metarsi}
& Improver & F2 & Later gains: original versus revised improver. & G & \SIPaper{https://arxiv.org/abs/2609.06396} \\[\SIBenchmarkRowGap]
RQGM\textsuperscript{\ensuremath{\dagger}}~\cite{iacob2026redqueen}
& Improver & F4 & Task gains and reliability of co-evolving judges. & G & \SIPaper{https://arxiv.org/abs/2606.26294} \\[\SIBenchmarkRowGap]
Double Ratchet\textsuperscript{\ensuremath{\dagger}}~\cite{zhang2026whogrades}
& Improver & F4 & Fixed-reference agreement and held-out gains. & C & \SIGitHub{https://github.com/amazon-science/Self-Evolving-Agents-Double-Ratchet} \\[\SIBenchmarkRowGap]
\bottomrule
\end{tabularx}

\end{minipage}
\end{lrbox}%
}
\def\SIBenchmarkSetFont#1#2{%
  \setlength{\SIBenchmarkFontSize}{#1pt}%
  \setlength{\SIBenchmarkLeading}{#2pt}%
}
\def\SIBenchmarkTryFont#1#2{%
  \ifdim\dimexpr\ht\SIBenchmarkTableBox+\dp\SIBenchmarkTableBox\relax>\SIBenchmarkTableMaxHeight
    \SIBenchmarkSetFont{#1}{#2}%
    \SIBenchmarkTypeset
  \fi
}
\SIBenchmarkSetFont{8.0}{9.0}
\SIBenchmarkTypeset
\ifdim\dimexpr\ht\SIBenchmarkTableBox+\dp\SIBenchmarkTableBox\relax<\SIBenchmarkTableMaxHeight
  \setlength{\SIBenchmarkRowGap}{\dimexpr
    (\SIBenchmarkTableMaxHeight-\ht\SIBenchmarkTableBox-\dp\SIBenchmarkTableBox)/52\relax}
  \ifdim\SIBenchmarkRowGap>0.1pt\SIBenchmarkTypeset\fi
\fi
\typeout{SITable layout: width=\the\SIBenchmarkTableWidth,
  font=\the\SIBenchmarkFontSize,
  height=\the\dimexpr\ht\SIBenchmarkTableBox+\dp\SIBenchmarkTableBox\relax,
  available=\the\SIBenchmarkTableMaxHeight.}
\noindent\begin{minipage}{\SIBenchmarkTableWidth}
\centering
\setlength{\parskip}{0pt}
\usebox{\SIBenchmarkCaptionBox}\par
\vspace{2pt}\nointerlineskip
\usebox{\SIBenchmarkTableBox}\par
\end{minipage}
\endgroup
\end{table*}

Conventional agent benchmarks primarily assess fixed-task performance. Evaluating improvement requires distinguishing revision quality, later-task benefits, and stronger future improvement capability. We examine two complementary questions: what is revised, and what evidence guides its revision. Table~\ref{tab:rsi-benchmarks} jointly records the What Improves and Who Drives Improve of each resource.

\subsection{What Improves: Target Evaluation}
\label{subsec:benchmark-target}

The revised object determines which capability changes an evaluation must measure.

Executable research suites evaluate experimentation and training~\cite{huang2024mlagentbench,chan2025mlebench,nathani2025mlgym}. RE-Bench compares research-engineering performance across time budgets~\cite{wijk2024rebench}. In model-training tasks, the target model may differ from the model executing the agent. PostTrainBench constrains post-training resources~\cite{rank2026posttrainbench}, while MLRC-Bench measures method performance~\cite{zhang2025mlrcbench}. These protocols assess research outputs or trained models under specified task and resource conditions.

Harness protocols evaluate retained instructions, code, skills, tools, and memory under a specified model~\cite{ursekar2026harnessoptbench,huang2026evobench,wu2026harnessdev}. Matched starting systems and search budgets separate revision gains from additional optimization~\cite{wang2026rethinking,gan2026harnessevolutionlearning}. The Meta-Agent Challenge supplies common agent-development constraints~\cite{lu2026metaagentchallenge}. MaD-RSI uses failures across heterogeneous models to optimize a shared harness; its metrics assess success coverage and joint success across models~\cite{cui2026madrsi}.

Repository, browser, desktop, and application suites supply executable task conditions~\cite{jimenez2023swebench,zhou2023webarena,xie2024osworld,liu2023agentbench,trivedi2024appworld,drouin2024workarena,dechezelles2024browsergym}. Paired runs with and without retained updates test persistence benefits. Component diagnostics further distinguish useful revisions from regressions: SkillsBench measures paired skill-use benefits~\cite{li2026skillsbench}; SkillScriptBench separates repair from behavior preservation~\cite{liu2026skillscriptbench}. EmbodiedCLBench uses offline, simulation-derived transition graphs~\cite{huang2026embodiedclbench}. LoCoMo and LongMemEval test conversational retention and temporal reasoning~\cite{maharana2024locomo,wu2024longmemeval}; MemoryAgentBench adds incremental learning and selective forgetting~\cite{hu2025memoryagentbench}. Dependency and interference diagnostics expose failures hidden by final-answer scores~\cite{gao2026memderive,ding2026memoryinterference}.

Generated tasks and curricula are evaluated through the resulting learner's performance, transfer, and data efficiency~\cite{poet,paired,accel,guo2025genenv}. Procgen, Meta-World, and MiniHack provide complementary test distributions~\cite{cobbe2019procgen,yu2019metaworld,samvelyan2021minihack}. MetaBench-Harness extends revision to task-construction and evaluation workflows~\cite{chen2026metabenchharness}. Generated-problem difficulty describes the evaluation resource; learner gains require a common external test. Versioned tasks and verifiers help separate generator changes from learner progress.

AI4AI-Bench and RSI-Exam evaluate revised methods through independent retraining or sealed evaluation~\cite{chi2026ai4aibench,rsiexam2026}; RSIGym supplies configurable improvement tasks~\cite{meng2026rsigymflexibleenvironmentrecursive}. Direct improver evaluation compares original and revised mechanisms through their subsequent outputs. STOP, Hyperagents, MetaRSI, and AIDE$^2$ examine later gains produced by revised improvement procedures~\cite{zelikman2024stop,zhang2026hyperagents,tan2026metarsi,srikanth2026aide2}. Lineage productivity provides complementary evidence whose interpretation depends on search allocation~\cite{wang2025huxley}. A full-text audit distinguishes research-product assessment from improver comparison within its literature sample~\cite{chen2026auditingrecursion}. Such comparisons require common starting systems, independent continuation tasks, and matched continuation budgets, with improver-construction costs reported separately.

\subsection{Who Drives Improve: Feedback Evaluation}
\label{subsec:benchmark-feedback}

Feedback determines which evidence guides revision and how its benefits are assessed. Protocols must distinguish signals available during improvement from measurements used only for final acceptance.

InstructGPT provides a study-specific protocol for human-preference-guided model updates~\cite{instructgpt}. Evaluation should distinguish preferences collected for improvement from held-out human judgments, while reporting annotation cost and agreement.

Verifiable outcomes support revision, but their timing and reuse conditions affect what gains mean. RSIBench-Data's checkpoint selection and official evaluation share a task subset~\cite{evolvent2026rsibenchdata}, limiting claims about held-out task generalization. FinEvolveBench exposes delayed, noisy outcomes with limited experience reuse~\cite{zhu2026finevolvebench}. Evaluations must therefore control outcome availability, temporal leakage, and access to prior experience.

In $\tau$-bench and $\tau^2$-bench, users are AI-simulated~\cite{yao2024taubench,barres2025tau2bench}; PAHF likewise studies adaptation to simulated preferences~\cite{liang2026pahf}. These settings differ from real human guidance. FinEvo-Bench compares retained experience with paired state-reset runs using fixed AI scoring and diagnostic feedback~\cite{deng2026finevo}. PaperBench uses AI judges to apply replication rubrics~\cite{starace2025paperbench}. MLR-Bench uses AI reviews to assess research products~\cite{chen2025mlrbench}. Human-authored rubrics do not make AI grading Human-Steered Feedback, and hidden final judgments may provide assessment without guiding revision. Reports should specify judge versions, prompts, and whether diagnoses are available during improvement.

Evaluation must assess changes to the feedback mechanism as well as resulting task gains. EvoRubric updates rubric generation and retained criteria~\cite{evo_rubric}. RQGM co-evolves agents and learned evaluators, using ground-truth anchors to govern evaluator replacement~\cite{iacob2026redqueen}. Double Ratchet evolves inspectable metrics against anchored references and couples metric and skill updates~\cite{zhang2026whogrades}. Comparing feedback versions requires fixed or independently anchored references; improved internal scores alone cannot establish more reliable evaluation.

FinEvo-Bench~\cite{deng2026finevo} and FinEvolveBench~\cite{zhu2026finevolvebench} evaluate harness evolution using AI-facilitated and environment-grounded feedback, respectively. Conversely, environment-grounded feedback supports both model and improver evaluation (Table~\ref{tab:rsi-benchmarks}). CHASE's counterfactual protocol tests show why fresh tasks alone can preserve protocol shortcuts~\cite{zhu2026chase}. Comparisons should report both the object changed and the feedback accessible at each revision, together with task exposure, retained state, repeated runs, and total development and deployment costs.

%% file: section/application.tex
\FloatBarrier
\section{Applications}
\label{sec:application}

\providecommand{\SIGitHubIcon}{%
  \raisebox{-0.10ex}{\tikz[x=1pt,y=1pt,scale=0.30,yscale=-1,baseline=(current bounding box.south)]{%
    \path[use as bounding box] (0,0) rectangle (24,24);%
    \fill svg {M12 .297c-6.63 0-12 5.373-12 12 0 5.303 3.438 9.8 8.205 11.385.6.113.82-.258.82-.577 0-.285-.01-1.04-.015-2.04-3.338.724-4.042-1.61-4.042-1.61C4.422 18.07 3.633 17.7 3.633 17.7c-1.087-.744.084-.729.084-.729 1.205.084 1.838 1.236 1.838 1.236 1.07 1.835 2.809 1.305 3.495.998.108-.776.417-1.305.76-1.605-2.665-.3-5.466-1.332-5.466-5.93 0-1.31.465-2.38 1.235-3.22-.135-.303-.54-1.523.105-3.176 0 0 1.005-.322 3.3 1.23.96-.267 1.98-.399 3-.405 1.02.006 2.04.138 3 .405 2.28-1.552 3.285-1.23 3.285-1.23.645 1.653.24 2.873.12 3.176.765.84 1.23 1.91 1.23 3.22 0 4.61-2.805 5.625-5.475 5.92.42.36.81 1.096.81 2.22 0 1.606-.015 2.896-.015 3.286 0 .315.21.69.825.57C20.565 22.092 24 17.592 24 12.297c0-6.627-5.373-12-12-12};%
  }}%
}
\providecommand{\SIPaperIcon}{%
  \raisebox{-0.10ex}{\tikz[x=1pt,y=1pt,scale=0.30,yscale=-1,baseline=(current bounding box.south)]{%
    \path[use as bounding box] (0,0) rectangle (24,24);%
    \draw[line width=0.55pt,line join=round] (5,2)--(14,2)--(20,8)--(20,22)--(5,22)--cycle;%
    \draw[line width=0.55pt] (14,2)--(14,8)--(20,8);%
    \draw[line width=0.45pt] (8,12)--(17,12) (8,15)--(17,15) (8,18)--(15,18);%
  }}%
}
\providecommand{\SIWebIcon}{%
  \raisebox{-0.10ex}{\tikz[x=1pt,y=1pt,scale=0.30,baseline=(current bounding box.south)]{%
    \path[use as bounding box] (0,0) rectangle (24,24);%
    \draw[line width=0.55pt] (12,12) circle (10);
    \draw[line width=0.45pt] (12,12) ellipse (4.5 and 10);
    \draw[line width=0.45pt] (2,12)--(22,12) (4,6)--(20,6) (4,18)--(20,18);
  }}%
}
\providecommand{\RSIAppPaper}[1]{\href{#1}{\SIPaperIcon}}
\providecommand{\RSIAppCode}[1]{\href{#1}{\SIGitHubIcon}}
\providecommand{\RSIAppWeb}[1]{\href{#1}{\SIWebIcon}}

\newsavebox{\RSIAppCaptionBox}
\newsavebox{\RSIAppBodyBox}
\newsavebox{\RSIAppBestBox}
\newlength{\RSIAppAvailableHeight}
\newdimen\RSIAppTableWidth
\newdimen\RSIAppLowerFont
\newdimen\RSIAppUpperFont
\newdimen\RSIAppFont
\newdimen\RSIAppLeading
\newcount\RSIAppFitIteration

\begin{table*}[!p]
\centering
\begingroup
\RSIAppTableWidth=\textwidth
\sbox{\RSIAppCaptionBox}{%
  \begin{minipage}{\RSIAppTableWidth}
  \caption[Representative applications and improvement mechanisms of AI agents.]{Representative applications and improvement mechanisms of AI agents. Targets: O = Output (task artifacts); M = Model; H = Harness; E = Environment; I = Improver. Multiple codes indicate updates spanning targets. Product capabilities depend on the documented configuration. Links: \SIPaperIcon\ paper; \SIGitHubIcon\ GitHub; \SIWebIcon\ official documentation.}
  \label{tab:rsi-applications}
  \end{minipage}%
}
\def\RSIAppBuildBody{%
\sbox{\RSIAppBodyBox}{%
\begin{minipage}{\RSIAppTableWidth}
\let\bffont\relax 
\RSIAppLeading=\dimexpr\RSIAppFont*112/100\relax
\fontsize{\the\RSIAppFont}{\the\RSIAppLeading}\selectfont
\setlength{\tabcolsep}{0.32\RSIAppFont}
\renewcommand{\arraystretch}{1.00}
\begin{tabularx}{\linewidth}{@{}
>{\raggedright\arraybackslash}p{0.215\linewidth}
>{\raggedright\arraybackslash}p{0.195\linewidth}
>{\raggedright\arraybackslash}p{0.085\linewidth}
>{\raggedright\arraybackslash}X
>{\centering\arraybackslash}p{0.035\linewidth}
@{}}
\toprule
\textbf{System} & \textbf{Application} & \shortstack{\textbf{Updated}\\\textbf{Object}} &
\textbf{Description} & \textbf{Link} \\
\midrule
\multicolumn{5}{c}{\textbf{Software Engineering}} \\
\midrule
SWE-agent~\cite{yang2024sweagent}
& Repository repair & O
& Inspect, edit, and test repository patches.
& \RSIAppPaper{https://arxiv.org/abs/2405.15793} \\

RuleEvolve~\cite{jiang2026ruleevolve}
& Coding rules & H & Retain coding rules from evaluated variants. & \RSIAppPaper{https://arxiv.org/abs/2610.00650v1} \\

SICA~\cite{robeyns2025sica}
& Agent development & H/I
& Reuse evaluated agents for later self-modification.
& \RSIAppPaper{https://arxiv.org/abs/2504.15228} \\

DGM~\cite{zhang2026darwin}
& Agent development & H/I
& Retain evaluated self-modifications in an archive.
& \RSIAppPaper{https://arxiv.org/abs/2505.22954} \\

Agentic Harness Engineering~\cite{lin2026agentic}
& Harness optimization & H
& Revise harnesses; test frozen-harness transfer.
& \RSIAppPaper{https://arxiv.org/abs/2604.25850} \\

ReTerm~\cite{xie2026reterm}
& Terminal recovery & M & Learn failure recovery after teacher warm-start. & \RSIAppPaper{https://arxiv.org/abs/2610.18484v2} \\

Codex CLI~\cite{openai_codex_cli}
& Software work & O
& Edit code and execute commands and tests.
& \RSIAppCode{https://github.com/openai/codex} \\

Claude Code~\cite{anthropic_claude_code_memory}
& Agentic coding & O/H
& Edit code; retain configurable project memory.
& \RSIAppWeb{https://code.claude.com/docs/en/memory} \\

Cursor~\cite{cursor_project_rules}
& Software work & O/H
& Edit code; create project rules on request.
& \RSIAppWeb{https://cursor.com/docs/rules} \\

OpenHands~\cite{openhands_persistent_memory}
& Software work & O/H
& Automate development; optionally retain memory.
& \RSIAppCode{https://github.com/OpenHands/OpenHands} \\

OpenCode~\cite{anomaly_opencode_rules}
& Terminal coding & O/H
& Edit code; initialize rules on user invocation.
& \RSIAppCode{https://github.com/anomalyco/opencode} \\

\midrule
\multicolumn{5}{c}{\textbf{Autonomous AI Research}} \\
\midrule
AlphaEvolve~\cite{novikov2025alphaevolve}
& Algorithm discovery & O
& Search algorithms using executable evaluations.
& \RSIAppPaper{https://arxiv.org/abs/2506.13131} \\

Athanor~\cite{zhou2026athanor}
& Post-training & O/M & Train separate target models; retain research logs. & \RSIAppPaper{https://arxiv.org/abs/2610.21055v1} \\

\'{E}pi~\cite{huang2026epi}
& Research memory & H & Retain hierarchical memory across model sessions. & \RSIAppPaper{https://arxiv.org/abs/2610.24944v1} \\

Frontis-MA1~\cite{yang2026frontisma1}
& ML engineering & M/I
& Train program-evolution operators from execution.
& \RSIAppPaper{https://arxiv.org/abs/2607.28568} \\

STOP~\cite{zelikman2024stop}
& Program optimization & I
& Apply the code improver to its own program.
& \RSIAppPaper{https://arxiv.org/abs/2310.02304} \\

EvoX~\cite{liu2026evox}
& Automated optimization & O/I
& Co-evolve solutions and variation/selection rules.
& \RSIAppPaper{https://arxiv.org/abs/2602.23413} \\

G{\"o}del Agent~\cite{yin2025goedelagent}
& Agent development & H/I
& Revise its own executable operating logic.
& \RSIAppPaper{https://aclanthology.org/2025.acl-long.1354/} \\

AIDE\textsuperscript{2}~\cite{srikanth2026aide2}
& Research agents & H/I
& Retain rewrites; test outer-loop improvement.
& \RSIAppPaper{https://arxiv.org/abs/2609.26457} \\

Hyperagents~\cite{zhang2026hyperagents}
& Agent development & H/I
& Evolve task-agent and meta-agent programs.
& \RSIAppPaper{https://arxiv.org/abs/2603.19461} \\

MetaRSI-v1~\cite{tan2026metarsi}
& Improver research & M/H/I
& Compare later gains from revised improvers.
& \RSIAppPaper{https://arxiv.org/abs/2609.06396} \\

autoresearch~\cite{karpathy2026autoresearch}
& LLM research & O
& Revise training code using validation metrics.
& \RSIAppCode{https://github.com/karpathy/autoresearch} \\

AIDE ML~\cite{jiang2025aide}
& ML engineering & O
& Search and debug ML programs using metrics.
& \RSIAppCode{https://github.com/WecoAI/aideml} \\

ShinkaEvolve~\cite{lange2025shinka}
& Program search & O
& Mutate and evaluate executable programs.
& \RSIAppCode{https://github.com/SakanaAI/ShinkaEvolve} \\
\midrule
\multicolumn{5}{c}{\textbf{Scientific Discovery}} \\
\midrule
Coscientist~\cite{boiko2023coscientist}
& Reaction design & O
& Revise reaction plans using experimental data.
& \RSIAppPaper{https://www.nature.com/articles/s41586-023-06792-0} \\

A-Lab~\cite{szymanski2023alab}
& Materials synthesis & O
& Revise recipes using physical characterization.
& \RSIAppPaper{https://www.nature.com/articles/s41586-023-06734-w} \\

Self-Building Labs~\cite{bai2026selfbuildinglabs}
& Lab automation & H/E & Build digital twins and robot skills; human approval. & \RSIAppPaper{https://arxiv.org/abs/2610.23826v1} \\

Co-Scientist~\cite{gottweis2026coscientist}
& Biomedical research & O
& Revise hypotheses for experimental validation.
& \RSIAppPaper{https://www.nature.com/articles/s41586-026-10644-y} \\

Virtual Lab~\cite{swanson2025virtuallab}
& Nanobody design & O
& Design nanobodies; validate them experimentally.
& \RSIAppPaper{https://www.nature.com/articles/s41586-025-09442-9} \\

Robin~\cite{ghareeb2026robin}
& Biology & O
& Update hypotheses from laboratory findings.
& \RSIAppPaper{https://www.nature.com/articles/s41586-026-10652-y} \\

\midrule
\multicolumn{5}{c}{\textbf{Web and Digital Agents}} \\
\midrule
ExpeL~\cite{zhao2024expel}
& Decision-making & H
& Extract and retrieve reusable task guidance.
& \RSIAppPaper{https://arxiv.org/abs/2308.10144} \\

ReasoningBank~\cite{ouyang2026reasoningbank}
& Digital tasks & H
& Distill trajectories into reusable strategies.
& \RSIAppPaper{https://arxiv.org/abs/2509.25140} \\

SkillWeaver~\cite{zheng2025skillweaver}
& Web automation & H
& Practice interactions; retain executable APIs.
& \RSIAppPaper{https://arxiv.org/abs/2504.07079} \\

Skill-Evo4GUI~\cite{hu2026skillevo4gui}
& GUI skills & H & Retain and revise versioned GUI skills across rounds. & \RSIAppPaper{https://arxiv.org/abs/2609.04869v1} \\

WebRL~\cite{webrl}
& Web navigation & M/E
& Train policies with failure-derived curricula.
& \RSIAppPaper{https://arxiv.org/abs/2411.02337} \\

WebEvolver~\cite{webevolver}
& Web interaction & M
& Co-train policies and predictive world models.
& \RSIAppPaper{https://aclanthology.org/2025.emnlp-main.454/} \\

GEPA~\cite{agrawal2025gepareflectivepromptevolution}
& Agent optimization & H
& Optimize instructions using trace reflection.
& \RSIAppCode{https://github.com/gepa-ai/gepa} \\

EvoAgentX~\cite{wang2025evoagentx}
& Workflow design & H
& Search evaluated prompts and workflows.
& \RSIAppCode{https://github.com/ANative-Lab/EvoAgentX} \\

\midrule
\multicolumn{5}{c}{\textbf{Open-World and Embodied Agents}} \\
\midrule
Voyager~\cite{wang2023voyager}
& Minecraft & H
& Accumulate executable exploration skills.
& \RSIAppPaper{https://arxiv.org/abs/2305.16291} \\

WorldCoder~\cite{worldcoder}
& Planning & H
& Revise world-model code from interactions.
& \RSIAppPaper{https://arxiv.org/abs/2402.12275} \\

EnvGen~\cite{envgen}
& RL training & E
& Generate practice targeting weak agent skills.
& \RSIAppPaper{https://arxiv.org/abs/2403.12014} \\

OMNI-EPIC~\cite{omniepic}
& Open-ended learning & E
& Retain generated environments and rewards.
& \RSIAppPaper{https://arxiv.org/abs/2405.15568} \\

EmbodiedEvo~\cite{li2026embodiedevo}
& Simulated control & H & Freeze learned code skills for unseen scenes. & \RSIAppPaper{https://arxiv.org/abs/2610.18867v2} \\

ReTrace~\cite{yu2026retrace}
& Robot skills & H & Retain validated corrections; four physical tasks. & \RSIAppPaper{https://arxiv.org/abs/2610.15814v1} \\

VLAW~\cite{vlaw}
& Robot control & M
& Use robot rollouts for world/policy learning.
& \RSIAppPaper{https://proceedings.mlr.press/v306/guo26i.html} \\

World-VLA-Loop~\cite{worldvlaloop}
& Embodied control & M
& Refine world models and VLA policies iteratively.
& \RSIAppPaper{https://arxiv.org/abs/2602.06508} \\
\bottomrule
\end{tabularx}
\end{minipage}%
}
}
\setlength{\RSIAppAvailableHeight}{%
  \dimexpr\textheight-\ht\RSIAppCaptionBox-\dp\RSIAppCaptionBox-2pt\relax}
\RSIAppLowerFont=8pt
\RSIAppUpperFont=8.5pt
\RSIAppFont=\RSIAppLowerFont
\RSIAppBuildBody
\setbox\RSIAppBestBox=\copy\RSIAppBodyBox
\RSIAppFitIteration=0
\loop
  \RSIAppFont=\dimexpr(\RSIAppLowerFont+\RSIAppUpperFont)/2\relax
  \RSIAppBuildBody
  \ifdim\dimexpr\ht\RSIAppBodyBox+\dp\RSIAppBodyBox\relax>\RSIAppAvailableHeight
    \RSIAppUpperFont=\RSIAppFont
  \else
    \RSIAppLowerFont=\RSIAppFont
    \setbox\RSIAppBestBox=\copy\RSIAppBodyBox
  \fi
  \advance\RSIAppFitIteration by 1
\ifnum\RSIAppFitIteration<6
\repeat
\typeout{SIApp layout: font=\the\RSIAppLowerFont, height=\the\dimexpr\ht\RSIAppBestBox+\dp\RSIAppBestBox\relax, available=\the\RSIAppAvailableHeight.}
\usebox{\RSIAppCaptionBox}\par\nointerlineskip
\makebox[\RSIAppTableWidth][c]{\usebox{\RSIAppBestBox}}
\endgroup
\end{table*}

The applications in Table~\ref{tab:rsi-applications} retain different products of feedback: task artifacts, reusable agent components, and improvement procedures. The five domains below compare their practical benefits and validation.

\subsection{Software Engineering.}
Executable tests support both immediate repair and reusable updates. SWE-agent, Agentless, and SWE-Search use interaction, localization, and search for current repairs~\cite{yang2024sweagent,xia2024agentless,antoniades2024swesearch}. Later-task gains require retained rules, experience, harness revisions, or learned parameters. SWE-Exp distills repair trajectories, while HarnessPro selects performance--cost trade-offs using generated tests~\cite{chen2025sweexp,pan2026harnesspro}. Table~\ref{tab:rsi-applications} contrasts rule and agent-code revision with ReTerm's parameter learning after teacher warm-start. For parameter learning, SWE-RL trains repair policies from software-evolution records~\cite{wei2025swerl}, while R2E-Gym and SWE-smith supply executable training tasks~\cite{jain2025r2egym,yang2025swesmith}. Repository overlap, supervision sources, regression behavior, and subsequent reuse qualify these gains. Project-memory retention depends on tool configuration.

\subsection{Autonomous AI Research.}
Research agents propose experiments, modify code, interpret results, and preserve useful findings. Research systems iterate over executable code candidates~\cite{novikov2025alphaevolve,romeraparedes2024funsearch,jiang2025aide,lange2025shinka,karpathy2026autoresearch}; the AI Scientist series extends this workflow to manuscripts and review~\cite{lu2024aiscientist,yamada2025aiscientistv2}. Athanor trains separate target models~\cite{zhou2026athanor}, whereas \'{E}pi retains research memory with fixed model weights~\cite{huang2026epi}. \'{E}pi-turbo's FrontierOR comparison uses one run per task and system, limiting statistical conclusions~\cite{huang2026epi}. The Agentic Scientist includes human collaboration, while ScientistTwo uses simulated review~\cite{lange2026agenticscientist,nam2026scientisttwo}. STOP, Hyperagents, AIDE$^2$, and MetaRSI assess revised improvement procedures through subsequent improvement outcomes~\cite{zelikman2024stop,zhang2026hyperagents,srikanth2026aide2,tan2026metarsi}.

\subsection{Scientific Discovery.}
Experimental outcomes revise hypotheses and subsequent measurements. Coscientist, A-Lab, and ChemCrow connect chemical or materials decisions to tools and experimental evidence~\cite{boiko2023coscientist,szymanski2023alab,bran2023chemcrow}. LLM-RDF uses analytical measurements in Bayesian optimization, with human-reviewed plans and agent connections~\cite{ruan2024llmrdf}; Self-Building Labs extends modification to laboratory capabilities, as summarized in Table~\ref{tab:rsi-applications}. Biological systems illustrate different validation levels: HypoForge and The Little Scientist use computational critique and testing~\cite{qian2026hypoforge,smith2026littlescientist}, whereas Co-Scientist, Virtual Lab, and Robin connect proposals to physical experiments under their reported human-participation conditions~\cite{gottweis2026coscientist,swanson2025virtuallab,ghareeb2026robin}.

\subsection{Web and Digital Agents.}
Repeated digital work creates reusable interface knowledge, which changes in sites and permissions can invalidate. Mind2Web and WebLINX provide demonstrations~\cite{deng2023mind2web,lu2024weblinx}, while WebVoyager tests live execution~\cite{he2024webvoyager}. ExpeL, ReasoningBank, and SkillWeaver retain guidance or executable procedures~\cite{zhao2024expel,ouyang2026reasoningbank,zheng2025skillweaver}; Skill-Evo4GUI evaluates repeated-task reuse against an empty-library control~\cite{hu2026skillevo4gui}. Agent S4 revises interaction resources under a fixed meta-agent model and harness~\cite{agashe2026agents4}. Adaptation includes policy learning and environment revision~\cite{webrl,webevolver,pandya2026echoverse}, alongside outcome-grounded forecasting in EvoForesight~\cite{chen2026evoforesight}. Practical benefit depends on later-task success and recovery from obsolete skills as interfaces, evaluators, or news distributions change.

\subsection{Open-World and Embodied Agents.}
Exploration requires reusable capabilities and suitable practice. Voyager and WorldCoder retain skills or predictive programs, while EnvGen and OMNI-EPIC generate training conditions~\cite{wang2023voyager,worldcoder,envgen,omniepic}. Grounded action selection and programmatic control provide foundations~\cite{ahn2022saycan,liang2022codeaspolicies}. Table~\ref{tab:rsi-applications} contrasts EmbodiedEvo's frozen-skill transfer to unseen scenes~\cite{li2026embodiedevo} with ReTrace's corrections validated on four physical tasks~\cite{yu2026retrace}. EvolvingBrain tests frozen-experience transfer, whereas ASENA combines repeated simulated tasks with a robot demonstration~\cite{liu2026evolvingbrain,cheng2026asena}. Pretrained policies support subsequent adaptation~\cite{brohan2023rt2,kim2024openvla,octomodelteam2024octo,bousmalis2023robocat}; SEES targets failure-specific adaptation of a pretrained VLA~\cite{li2026sees}. Coupled policy/world-model learning adds another route~\cite{vlaw,worldvlaloop}. These protocols distinguish reuse within an adaptation setting from transfer to new conditions.

Across domains, controlled subsequent use and independent evaluation establish the value of retained changes. Recursive growth requires comparing the later gains produced by original and revised improvers.

\FloatBarrier

%% file: section/future_directions.tex
\section{Challenges and Future Directions}
\label{sec:challenges-future}


The preceding sections have traced the progression of agent improvement from persistent system evolution and increasingly autonomous feedback to agent–environment co-evolution and recursive improvement of the Improver itself. Building on these advances, several promising research directions are emerging. Trustworthy RSI focuses on maintaining the reliability and verifiability of recursive improvement across generations, while Scientific Agent RSI and Multi-Agent RSI explore its potential in autonomous scientific research and collaborative agent systems. Meanwhile, RSI Infrastructure provides a foundation for persistent experimentation, experience accumulation, and cross-generational evolution. Looking further ahead, Intelligence Explosion raises the fundamental question of whether improvements in improvement mechanisms can compound across generations and accelerate the growth of AI capabilities. This section discusses these emerging directions and their research opportunities toward reliable RSI.

\subsection{Trustworthy RSI}

Trustworthy RSI has recently received dedicated attention, with a comprehensive survey organizing its requirements into four interconnected themes: genuine and sustained improvement, behavioral constraints and alignment, system security and privacy, and transparency and controllability~\cite{pu2026trustworthyrecursiveselfimprovementera}. As agent capabilities, task distributions, and interaction environments evolve, fixed evaluators and diagnostic procedures may gradually become inadequate. No More Stale Feedback co-evolves critics to adapt to open-world agent learning, while Who Grades the Grader? explores the co-evolution of evaluation metrics and agent skills, using anchored reference standards to assess updated evaluators \cite{echo,zhang2026whogrades}. Meanwhile, self-authored verification procedures may fail to reflect actual task performance \cite{guo2026selfauthoredverification}, whereas statistical acceptance mechanisms and risk-controlled self-modification provide initial safeguards for candidate updates \cite{sea2026,sage2026,wu2025sgm}. However, the continued evolution of feedback mechanisms may also alter evaluation criteria, causing internal assessments to diverge from genuine capability gains. When both evaluators and Improvers are modifiable, maintaining the independence of evaluation criteria and ensuring the validity of updated feedback mechanisms remain challenging. Future research should establish independent evaluation references and verification mechanisms to continuously assess feedback mechanisms and candidate modifications, enabling adaptation to changing task conditions without undermining the credibility of improvement outcomes.

Beyond the validity of improvement outcomes, continued evolution must preserve behavioral constraints and safety properties. Your Agent May Misevolve identifies risks across different evolution pathways, including models, memory, tools, and workflows, revealing that memory accumulation may degrade safety alignment, while tool creation and reuse may introduce unexpected vulnerabilities \cite{shao2026agentmisevolveemergentrisks}. SEABench further shows that locally beneficial updates may lead to unsafe behaviors in subsequent tasks, even in the absence of direct adversarial inputs \cite{das2026seabench}. In addition, When Self-Evolution Backfires explores pre-commit gating to prevent skill contamination, while SkillJack reveals that contaminated interaction experiences can be transformed into persistent malicious behaviors through skill extraction, continuing to affect subsequent tasks even after the original experiences are removed \cite{shang2026backfires,ying2026skilljackpersistentskillbackdoors}. These studies indicate that improvements in local task performance do not necessarily ensure the preservation of existing behavioral constraints, as accumulated experiences and reusable components may propagate erroneous behaviors across tasks and system versions. Future research should incorporate safety constraints and permission requirements into candidate screening and acceptance, while validating experiences before they are transformed into persistent memories, skills, and tools. Such mechanisms should enable beneficial experiences to be reused without allowing unreliable behaviors and sensitive information to propagate throughout system evolution.

As the evolution target further extends to the Improver itself, trustworthy RSI also requires effective oversight of the entire recursive improvement process. Modifying candidate generation, evaluation, and acceptance mechanisms may simultaneously change the oversight conditions on which subsequent updates rely, rendering existing verification results and control mechanisms inapplicable to new system versions. Reflections on Trusting Trust, Revisited demonstrates that poisoning evaluation benchmarks can induce self-modifying coding agents to retain unsafe instructions in subsequent versions, even when evolution continues using clean benchmarks \cite{roesner2026trustingtrust}. RSI Control further investigates how agents participating in next-generation model training may covertly introduce behaviors or objectives that deviate from intended specifications while maintaining performance on primary training tasks \cite{zhou2026rsicontrolaimodels}. These findings indicate that single-update verification and rollback of the original modification are insufficient to eliminate cross-generational risks. Future research should establish traceability across the entire improvement process by recording modification provenance, version dependencies, and validation evidence, while integrating independent auditing, authorization controls, execution replay, and rollback mechanisms to supervise systems throughout continued evolution. In particular, identifying and repairing erroneous modifications propagated to successor versions, and verifying that repaired systems preserve their original constraints after recursive evolution resumes, remain important challenges for preventing unreliable modifications from becoming the foundation for subsequent improvement.

\subsection{Scientific Agent RSI}

With the advancement of AI4AI, researchers have begun to explore how AI systems can autonomously improve models, algorithms, and their development processes, further extending recursive self-improvement to scientific research. AIDE$^2$ explores the recursive improvement of research agents by modifying their search strategies and memory mechanisms, while Frontis-MA1 trains a dedicated AI4AI model with execution feedback to autonomously generate, debug, and optimize programs for machine learning engineering \cite{srikanth2026aide2,yang2026frontisma1}. Recent studies have further extended this paradigm to scientific agents. ScienceBuddy introduces a nested Model--Harness update mechanism, enabling scientific agents to continuously improve through researcher interactions and task execution experiences \cite{xue2026sciencebuddyrecursiveinrecursiveselfimprovementinteractive}. DiscoveryRSI leverages hypothesis evaluation, research contribution attribution, and historical trajectory replay to learn from the evolution of scientific hypotheses and improve subsequent research strategies \cite{zhong2026discoveryrsirecursivelylearningdiscover}. ScienceClaw further transforms verified execution experiences into reusable Skills and Operators, enabling scientific agents to accumulate and transfer research experience across sequential tasks \cite{zhang2026scienceclawbenchmarkingcontinualselfevolution}. In addition, ScientistOne establishes traceable links between research claims and their supporting evidence through a Chain-of-Evidence, while EviGraph represents dependencies among research questions, hypotheses, experiments, and findings through structured evidence graphs, providing complementary foundations for accumulating and reusing research knowledge \cite{meng2026scientistonehumanlevelautonomousresearch,ren2026evigraphevidenceguidedautonomousresearch}. Together, these studies demonstrate an emerging transition from automating scientific research toward the continuous self-improvement of scientific agents, making research processes and their accumulated experiences increasingly important sources for agent evolution.

However, existing studies have yet to establish a comprehensive recursive improvement mechanism that transforms progress in scientific research tasks into sustained enhancements of research capabilities and the mechanisms underlying their improvement. First, scientific discovery involves multiple interdependent stages, including literature retrieval, research gap identification, hypothesis formulation, experimental design, and result analysis. Consequently, research outcomes cannot be readily attributed to individual decisions or actions, making it challenging to identify the specific stages that contribute to scientific progress and determine which research capabilities should be improved. Second, the accumulation and reuse of research experiences remain limited by the challenges of experience abstraction and transfer. How to extract generalizable research strategies from individual scientific tasks while preserving their applicability conditions and domain-specific contexts, so that these strategies can be effectively transferred across tasks and disciplines, remains an important challenge. Furthermore, existing approaches primarily focus on improving task execution capabilities and their associated components, while how to leverage accumulated research experiences to improve the mechanisms for generating, selecting, and optimizing research strategies remains largely unexplored. Future research should establish a recursive process that connects research experience accumulation, strategy abstraction, and improvement mechanism updates, enabling scientific agents not only to leverage existing knowledge and experiences to accomplish research tasks, but also to continuously improve their own scientific research methodologies and the mechanisms through which these methodologies are developed and refined.

\subsection{Multi-Agent RSI}

Existing RSI approaches primarily focus on individual agents or shared models, while recent studies have begun to extend evolution to multi-agent systems and their collaborative mechanisms. GEA treats an agent group as the fundamental evolutionary unit, facilitating collective capability improvement through experience sharing across evolutionary branches~\cite{weng2026groupevolvingagentsopenendedselfimprovement}. Meta-Team transforms distributed execution experience into reusable improvements in agent behaviors, inter-agent coordination, and team organization, while EVOCHAMBER further explores test-time co-evolution at individual, team, and population levels~\cite{hao2026evolveteamcollaborativeselfevolution, zhang2026evochambertesttimecoevolutionmultiagent}. 
 MASS alternates multi-agent workflow optimization with model training on collectively generated trajectories, whereas Multi-Agent Evolve jointly optimizes Proposer, Solver, and Judge roles to enable model self-evolution through collaborative learning~\cite{lee2026recursiveselfimprovementmultiagentselfsupervision, chen-etal-2026-multi-agent-evolve}. 
 Beyond evolving agent capabilities, Skill-MAS evolves Meta-Skills for automatic MAS generation, while EvoMAS and Learning to Evolve investigate the adaptation of evolutionary strategies and optimization mechanisms to improve subsequent system modifications~\cite{lin-etal-2026-skill, wei2026evomas,he-etal-2026-learning}. 
 SwarmRSI further promotes the accumulation and recombination of collective research experience through cross-agent experimental feedback, hypothesis evolution, and program components, while MetaShare enables the reuse of successful and failed modification experiences across evolutionary lineages~\cite{meng2026swarmrsirecursiveselfimprovementevolving, bhutani2026metasharefailureawarecollectiveselfimprovement}. 
 Together, these developments indicate a shift from individual agent improvement toward the collective evolution of agent capabilities, collaboration mechanisms, and improvement processes.

However, performance gains in multi-agent systems do not necessarily indicate stronger collective improvement capabilities. Although existing approaches increasingly optimize agent populations, collaboration structures, and even evolutionary strategies, whether these improvements can be progressively accumulated to strengthen the collective improver itself remains insufficiently explored. Recent studies on recursive social improvement further show that social learning among agents does not yet consistently outperform independent learning under matched resource budgets, suggesting that collective interaction and knowledge sharing do not necessarily translate into more effective collective improvement~\cite{jha2026solosociallearningcharacterizing}. 
Future research should therefore extend RSI from individual agents to entire multi-agent systems, enabling agents not only to improve collective task-solving capabilities through collaboration, but also to jointly enhance the mechanisms responsible for generating subsequent improvements. Such a transition could establish a collective improver whose improvement capabilities progressively accumulate across generations, extending recursive self-improvement from individual agents to multi-agent systems.

\subsection{RSI Infrastructure}

Recent studies have begun to extend recursive self-improvement to AI infrastructure, exploring how continuous experimentation and accumulated experience can improve complex computing systems. OpenRSI establishes an agent-driven optimization loop for real-world model serving, where an external harness manages iterative experimentation and independent verification, allowing validated modifications to accumulate across successive optimization rounds~\cite{li2026openrsisglangomni}. 
InfraRSI further applies this paradigm to distributed model training by jointly optimizing training configurations and code patches, calibrating search strategies through execution feedback, and transferring accumulated optimization experience across hardware platforms~\cite{infrarsiteam2026infrarsiharnessingrecursiveselfimprovement}. 
Meanwhile, infrastructure for autonomous improvement is also advancing. RSIGym exposes training, inference, rollout, and evaluation through reusable services, while Athanor supports long-horizon autonomous research through managed computation and persistent experimental records~\cite{meng2026rsigymflexibleenvironmentrecursive, zhou2026athanor}. 
RSI-Env further supports the inheritance of model, harness, and research states across generations, enabling updated systems to continue subsequent improvement~\cite{qu2026rsienvopenendedenvironmentrecursive}. Together, these efforts demonstrate the potential of RSI for real-world AI infrastructure and the transition from isolated experimental tools toward persistent improvement platforms.

However, existing approaches remain largely tailored to specific systems, fixed optimization objectives, or predefined improvement workflows, lacking a general architecture that supports the joint evolution of infrastructure and improvement mechanisms. Although OpenRSI and InfraRSI demonstrate the feasibility of accumulating verified infrastructure optimizations, successive performance gains and experience reuse do not necessarily indicate that the underlying improver has become more effective. Future research should therefore develop unified RSI infrastructure that integrates training and inference services, experimental execution, resource scheduling, version management, and improvement processes into a persistent system. Such infrastructure should enable validated optimizations to inform subsequent search strategies, experimental orchestration, and resource allocation, rather than merely improving the current system configuration. More importantly, it should support controlled modifications and cross-generational inheritance of the improver itself, while maintaining independent verification and rollback mechanisms. This progression could transform AI infrastructure from a passive substrate for model training and agent execution into a self-improving system foundation that supports sustained and verifiable recursive improvement.

\subsection{Intelligence Explosion}

If gains in improver capability can be retained and reused across generations, a longer-term question is whether they can compound and accelerate the growth of AI capabilities. The classical intelligence explosion hypothesis suggests that increasingly capable systems may contribute to designing stronger successors~\cite{good1966ultraintelligent}. Recursive self-modification provides a potential mechanism for this progression by making improvement mechanisms themselves modifiable~\cite{schmidhuber2003goedel}. However, recursive modification and even demonstrated gains in improver capability over limited generations do not necessarily imply sustained acceleration. An important future direction is therefore to determine whether improvement capabilities can compound across generations and progressively increase the efficiency of subsequent improvements.

Sustained acceleration further depends on self-reinforcing progress in AI research, the time required for successive improvement generations, and available resources. Larger gains in one generation do not necessarily shorten the next, while diminishing returns, computational bottlenecks, and increasing validation costs may constrain long-term growth. Theoretical analyses also distinguish accelerating growth from stronger forms of intelligence explosion, including finite-time singularities~\cite{ord2026dynamics,cunningham2026economics,whitfill2025compute}. Future research should jointly examine cross-generational gains, improvement time, resource requirements, and transfer across domains to determine whether recursive improvement can overcome these constraints and whether advances in AI research can translate into broader capability growth.

%% file: section/conclusion.tex
\section{Conclusion}
\label{sec:conclusion}

As AI agents evolve beyond task-local refinement, their improvement increasingly involves persistent modifications to foundation models, agent harnesses, and environments, supported by increasingly autonomous feedback mechanisms. However, expanding the scope of modification and automating feedback do not necessarily guarantee reliable improvement. Achieving reliable RSI requires identifying appropriate evolution targets, obtaining valid feedback signals, and ensuring that successive modifications produce genuine and sustained capability gains.

To systematically examine these developments, this survey introduces two complementary views, What Improves and Who Drives Improve, and organizes existing research into seven interrelated milestones. Through these perspectives, we analyze the relationships among evolution targets, feedback mechanisms, and improvement capabilities, clarifying the technological pathways from task-local optimization and persistent system evolution toward recursive self-improvement. We further review representative benchmarks, practical applications, and open challenges, highlighting the capabilities established by existing approaches and their limitations.

A central insight of this survey is that producing stronger agents does not necessarily imply developing stronger improvement mechanisms. Reliable RSI further requires demonstrating that revised improvers can be retained, reused, and shown to generate greater future improvements under comparable evaluation conditions. Future research should therefore advance accurate improvement diagnosis, reliable feedback guidance, coordinated system evolution, and cross-generational verification, enabling agent improvement to progress toward reliable, sustainable, and verifiable recursive self-improvement.

%% file: reference.bib
@string{iclr = {International Conference on Learning Representations}}

@string{acl = {Association for Computational Linguistics}}

@string{emnlp = {Empirical Methods in Natural Language Processing}}

@string{naacl = {North American Chapter of the Association for Computational Linguistics}}

@article{lin2026agentic,
  author = {J. Lin and others},
  title = {Agentic Harness Engineering: Observability-Driven Automatic Evolution of Coding-Agent Harnesses},
  journal = {arXiv preprint arXiv:2604.25850},
  year = {2026},
  doi = {10.48550/arXiv.2604.25850},
  url = {https://arxiv.org/abs/2604.25850},
  eprint = {2604.25850},
  archiveprefix = {arXiv}
}

@inproceedings{zhang2024aflow,
  author = {Jiayi Zhang and Jinyu Xiang and Zhaoyang Yu and Fengwei Teng and Xionghui Chen and Jiaqi Chen and Mingchen Zhuge and Xin Cheng and Sirui Hong and Jinlin Wang and Bingnan Zheng and Bang Liu and Yuyu Luo and Chenglin Wu},
  title = {{AFlow}: Automating Agentic Workflow Generation},
  booktitle = {International Conference on Learning Representations},
  year = {2025},
  doi = {10.48550/arXiv.2410.10762},
  url = {https://arxiv.org/abs/2410.10762}
}

@inproceedings{worldevolver,
  author = {Xuan Zhang and Wenxuan Zhang and See-Kiong Ng and Yang Deng},
  title = {Self-Evolving World Models for {LLM} Agent Planning},
  booktitle = {Findings of the Association for Computational Linguistics: EMNLP 2026},
  pages = {47553--47577},
  publisher = {Association for Computational Linguistics},
  year = {2026},
  doi = {10.18653/2026.findings-emnlp.2221},
  url = {https://aclanthology.org/2026.findings-emnlp.2221/},
  eprint = {2606.30639},
  archiveprefix = {arXiv}
}

@article{wang2026rethinking,
  author = {Yike Wang and Huaisheng Zhu and Zhengyu Hu and Yige Yuan and Zhengyu Chen and Shakti Senthil and Hannaneh Hajishirzi and Yulia Tsvetkov and Noah A. Smith and Pradeep Dasigi and Teng Xiao},
  title = {Rethinking the Evaluation of Harness Evolution for Agents},
  journal = {arXiv preprint arXiv:2607.12227},
  year = {2026},
  doi = {10.48550/arXiv.2607.12227},
  url = {https://arxiv.org/abs/2607.12227},
  eprint = {2607.12227},
  archiveprefix = {arXiv}
}

@article{wang2026compound,
  author = {Wenxiao Wang and Priyatham Kattakinda and Soheil Feizi},
  title = {Do Agent Optimizers Compound? A Continual-Learning Evaluation on {Terminal-Bench} 2.0},
  journal = {arXiv preprint arXiv:2607.14004},
  year = {2026},
  doi = {10.48550/arXiv.2607.14004},
  url = {https://arxiv.org/abs/2607.14004},
  eprint = {2607.14004},
  archiveprefix = {arXiv}
}

@inproceedings{ICLR2024_8b4add8b,
  author = {Huang, Jie and Chen, Xinyun and Mishra, Swaroop and Zheng, Huaixiu Steven and Yu, Adams Wei and Song, Xinying and Zhou, Denny},
  title = {Large Language Models Cannot Self-Correct Reasoning Yet},
  booktitle = {International Conference on Learning Representations},
  year = {2024},
  url = {https://openreview.net/forum?id=IkmD3fKBPQ}
}

@inproceedings{instructgpt,
  author = {Long Ouyang and Jeff Wu and Xu Jiang and Diogo Almeida and Carroll L. Wainwright and Pamela Mishkin and Chong Zhang and Sandhini Agarwal and Katarina Slama and Alex Ray and John Schulman and Jacob Hilton and Fraser Kelton and Luke Miller and Maddie Simens and Amanda Askell and Peter Welinder and Paul Christiano and Jan Leike and Ryan Lowe},
  title = {Training language models to follow instructions with human feedback},
  booktitle = {Advances in Neural Information Processing Systems},
  volume = {35},
  pages = {27730--27744},
  year = {2022},
  doi = {10.52202/068431-2011},
  url = {https://proceedings.neurips.cc/paper_files/paper/2022/hash/b1efde53be364a73914f58805a001731-Abstract.html},
  eprint = {2203.02155},
  archiveprefix = {arXiv},
  primaryclass = {cs.CL}
}

@article{kamoi-etal-2024-llms,
  author = {Kamoi, Ryo  and
      Zhang, Yusen  and
      Zhang, Nan  and
      Han, Jiawei  and
      Zhang, Rui},
  title = {When Can {LLM}s Actually Correct Their Own Mistakes? A Critical Survey of Self-Correction of {LLM}s},
  journal = {Transactions of the Association for Computational Linguistics},
  volume = {12},
  pages = {1417--1440},
  publisher = {MIT Press},
  year = {2024},
  doi = {10.1162/tacl_a_00713},
  url = {https://aclanthology.org/2024.tacl-1.78/},
  address = {Cambridge, MA}
}

@article{tan2026metarsi,
  author = {Zihan Tan and Leixin Sun and Zitong Shi and Yitao Liu and Jiajun Wu and Nathaniel Brooks and Jiaru Qian and Xiaoran Shang and Suyuan Huang and Yi Ding and Yangxu Liao and Mukai Li and Qiushi Sun and Shudong Liu and Xuankun Rong and Xiaohang Yu and Zhuo Chen and Hejia Geng and Chenxin Li and Aozhou Wang and Zengji Tu and Robert Tang and Yuxin Zhan and Eric Jiang and Yuxin Wu and Jianqing Zhang and Xiao Liang and Fang Wu and Haochi Zhang and Alexander Marlow and Guancheng Wan},
  title = {{MetaRSI} / {RSI}\textsuperscript{2}: A Meta-Recursive Self-Improving System for Recursive Self-Improving Systems Themselves},
  journal = {arXiv preprint arXiv:2609.06396},
  year = {2026},
  doi = {10.48550/arXiv.2609.06396},
  url = {https://arxiv.org/abs/2609.06396},
  eprint = {2609.06396},
  archiveprefix = {arXiv},
  primaryclass = {cs.LG}
}

@inproceedings{xu2025amem,
  author = {Wujiang Xu and Zujie Liang and Kai Mei and Hang Gao and Juntao Tan and Yongfeng Zhang},
  title = {{A-MEM}: Agentic Memory for {LLM} Agents},
  booktitle = {Advances in Neural Information Processing Systems},
  volume = {38},
  year = {2025},
  doi = {10.52202/085713-0593},
  url = {https://proceedings.neurips.cc/paper_files/paper/2025/hash/19909c36f51abc4856b4560aff3d36d6-Abstract-Conference.html}
}

@inproceedings{schick2023toolformer,
  author = {Timo Schick and Jane Dwivedi-Yu and Roberto Dess{\`i} and Roberta Raileanu and Maria Lomeli and Eric Hambro and Luke Zettlemoyer and Nicola Cancedda and Thomas Scialom},
  title = {{Toolformer}: Language Models Can Teach Themselves to Use Tools},
  booktitle = {Advances in Neural Information Processing Systems},
  volume = {36},
  year = {2023},
  doi = {10.52202/075280-2997},
  url = {https://proceedings.neurips.cc/paper/2023/hash/d842425e4bf79ba039352da0f658a906-Abstract-Conference.html},
  eprint = {2302.04761},
  archiveprefix = {arXiv}
}

@inproceedings{qin2024toolllm,
  author = {Yujia Qin and Shihao Liang and Yining Ye and Kunlun Zhu and Lan Yan and Yaxi Lu and Yankai Lin and Xin Cong and Xiangru Tang and Bill Qian and Sihan Zhao and Lauren Hong and Runchu Tian and Ruobing Xie and Jie Zhou and Mark Gerstein and Dahai Li and Zhiyuan Liu and Maosong Sun},
  title = {{ToolLLM}: Facilitating Large Language Models to Master 16000+ Real-world {APIs}},
  booktitle = {International Conference on Learning Representations},
  year = {2024},
  url = {https://openreview.net/forum?id=dHng2O0Jjr},
  eprint = {2307.16789},
  archiveprefix = {arXiv}
}

@inproceedings{yang2024sweagent,
  author = {J. Yang and C. E. Jimenez and A. Wettig and K. Lieret and S. Yao and K. Narasimhan and O. Press},
  title = {{SWE-agent}: Agent-Computer Interfaces Enable Automated Software Engineering},
  booktitle = {Advances in Neural Information Processing Systems},
  volume = {37},
  year = {2024},
  doi = {10.52202/079017-1601},
  url = {https://proceedings.neurips.cc/paper_files/paper/2024/hash/5a7c947568c1b1328ccc5230172e1e7c-Abstract-Conference.html},
  eprint = {2405.15793},
  archiveprefix = {arXiv}
}

@inproceedings{CRITIC,
  author = {Zhibin Gou and Zhihong Shao and Yeyun Gong and Yelong Shen and Yujiu Yang and Nan Duan and Weizhu Chen},
  title = {{CRITIC}: Large Language Models Can Self-Correct with Tool-Interactive Critiquing},
  booktitle = {International Conference on Learning Representations},
  year = {2024},
  doi = {10.48550/arXiv.2305.11738},
  url = {https://arxiv.org/abs/2305.11738}
}

@inproceedings{mtbench,
  author = {Lianmin Zheng and Wei-Lin Chiang and Ying Sheng and Siyuan Zhuang and Zhanghao Wu and Yonghao Zhuang and Zi Lin and Zhuohan Li and Dacheng Li and Eric P. Xing and Hao Zhang and Joseph E. Gonzalez and Ion Stoica},
  title = {Judging {LLM-as-a-Judge} with {MT-Bench} and {Chatbot Arena}},
  booktitle = {Advances in Neural Information Processing Systems},
  year = {2023},
  doi = {10.48550/arXiv.2306.05685},
  url = {https://arxiv.org/abs/2306.05685v4}
}

@inproceedings{shinn2023reflexion,
  author = {N. Shinn and F. Cassano and A. Gopinath and K. Narasimhan and S. Yao},
  title = {{Reflexion}: Language Agents with Verbal Reinforcement Learning},
  booktitle = {Advances in Neural Information Processing Systems},
  volume = {36},
  year = {2023},
  url = {https://proceedings.neurips.cc/paper_files/paper/2023/hash/1b44b878bb782e6954cd888628510e90-Abstract-Conference.html}
}

@inproceedings{SelfRefine,
  author = {Aman Madaan and Niket Tandon and Prakhar Gupta and Skyler Hallinan and Luyu Gao and Sarah Wiegreffe and Uri Alon and Nouha Dziri and Shrimai Prabhumoye and Yiming Yang and Shashank Gupta and Bodhisattwa Prasad Majumder and Katherine Hermann and Sean Welleck and Amir Yazdanbakhsh and Peter Clark},
  title = {{Self-Refine}: Iterative Refinement with Self-Feedback},
  booktitle = {Advances in Neural Information Processing Systems},
  volume = {36},
  year = {2023},
  doi = {10.52202/075280-2019},
  url = {https://proceedings.neurips.cc/paper_files/paper/2023/hash/91edff07232fb1b55a505a9e9f6c0ff3-Abstract-Conference.html}
}

@article{wang2023voyager,
  author = {Guanzhi Wang and Yuqi Xie and Yunfan Jiang and Ajay Mandlekar and Chaowei Xiao and Yuke Zhu and Linxi Fan and Anima Anandkumar},
  title = {{Voyager}: An Open-Ended Embodied Agent with Large Language Models},
  journal = {Transactions on Machine Learning Research},
  year = {2024},
  doi = {10.48550/arXiv.2305.16291},
  url = {https://openreview.net/pdf?id=ehfRiF0R3a},
  eprint = {2305.16291},
  archiveprefix = {arXiv},
  month = {March}
}

@inproceedings{guo2025genenv,
  author = {Guo, Jiacheng and Yang, Ling and Chen, Peter and Xiao, Qixin and Wang, Yinjie and Juan, Xinzhe and Qiu, Jiahao and Shen, Ke and Wang, Mengdi},
  title = {{GenEnv}: Difficulty-Aligned Co-Evolution Between {LLM} Agents and Environment Simulators},
  booktitle = {The Fourteenth International Conference on Learning Representations},
  year = {2026},
  url = {https://openreview.net/forum?id=5vsYRklFJf},
  eprint = {2512.19682},
  archiveprefix = {arXiv},
  primaryclass = {cs.CL}
}

@inproceedings{scaler,
  author = {Caijun Xu and Changyi Xiao and Zhongyuan Peng and Xinrun Wang and Yixin Cao},
  title = {{SCALER}: Synthetic Scalable Adaptive Learning Environment for Reasoning},
  booktitle = {Findings of the Association for Computational Linguistics: {ACL} 2026},
  pages = {31905--31923},
  publisher = {Association for Computational Linguistics},
  year = {2026},
  doi = {10.18653/v1/2026.findings-acl.1596},
  url = {https://aclanthology.org/2026.findings-acl.1596/},
  eprint = {2601.04809},
  archiveprefix = {arXiv}
}

@article{spade,
  author = {Liu, Bo and Yu, Simon and Jiang, Yiding and Qu, Ao and Zhao, Andrew and Liu, Zichen and Kim, Junsu and Zhou, Zijian and Kim, Seungone and Ren, Tongzheng and Liu, Mickel and Yu, Hanfei and Chen, Zhaorun and Shi, Weiyan and Liang, Paul Pu and Zettlemoyer, Luke and Choi, Yejin and Jaques, Natasha},
  title = {{SPADE}: Self-Play in Adaptive Synthetic Executable Environments},
  journal = {arXiv preprint arXiv:2608.19197},
  year = {2026},
  doi = {10.48550/arXiv.2608.19197},
  url = {https://arxiv.org/abs/2608.19197},
  eprint = {2608.19197},
  archiveprefix = {arXiv},
  primaryclass = {cs.CL}
}

@inproceedings{envgen,
  author = {Abhay Zala and Jaemin Cho and Han Lin and Jaehong Yoon and Mohit Bansal},
  title = {{EnvGen}: Generating and Adapting Environments via {LLMs} for Training Embodied Agents},
  booktitle = {First Conference on Language Modeling},
  year = {2024},
  doi = {10.48550/arXiv.2403.12014},
  url = {https://arxiv.org/abs/2403.12014},
  eprint = {2403.12014},
  archiveprefix = {arXiv}
}

@article{envsforge,
  author = {Wu, Xiaojun and Yang, Cehao and Liu, Honghao and Lin, Xueyuan and Shi, Zhichao and Zhou, Hao and Jiang, Xuhui and Xu, Chengjin and Li, Jia and Guo, Jian},
  title = {{Envs-FORGE}: Frontier-Optimized Reward-Grounded Environment Synthesis for Agent {RL}},
  journal = {arXiv preprint arXiv:2608.14312},
  year = {2026},
  doi = {10.48550/arXiv.2608.14312},
  url = {https://arxiv.org/abs/2608.14312},
  eprint = {2608.14312},
  archiveprefix = {arXiv}
}

@article{agentworld,
  author = {Dong, Guanting and Lu, Junting and Huang, Junjie and Zhong, Wanjun and Liu, Longxiang and Huang, Shijue and Li, Zhenyu and Zhao, Yang and Song, Xiaoshuai and Li, Xiaoxi and Jin, Jiajie and Zhu, Yutao and Wang, Hanbin and Lei, Fangyu and Luo, Qinyu and Chen, Mingyang and Chen, Zehui and Feng, Jiazhan and Wen, Ji-Rong and Dou, Zhicheng},
  title = {{Agent-World}: Scaling Real-World Environment Synthesis for Evolving General Agent Intelligence},
  journal = {arXiv preprint arXiv:2604.18292},
  year = {2026},
  doi = {10.48550/arXiv.2604.18292},
  url = {https://arxiv.org/abs/2604.18292},
  eprint = {2604.18292},
  archiveprefix = {arXiv},
  primaryclass = {cs.AI}
}

@inproceedings{zhao2025absolutezero,
  author = {Zhao, Andrew and Wu, Yiran and Wu, Tong and Xu, Quentin and Yue, Yang and Lin, Matthieu and Wang, Shenzhi and Wu, Qingyun and Zheng, Zilong and Huang, Gao},
  title = {{Absolute Zero}: Reinforced Self-play Reasoning with Zero Data},
  booktitle = {Advances in Neural Information Processing Systems},
  volume = {38},
  pages = {117235--117298},
  year = {2025},
  doi = {10.52202/085713-3534},
  url = {https://papers.nips.cc/paper_files/paper/2025/hash/9837dc00ff67d176373268ed48042d49-Abstract-Conference.html},
  eprint = {2505.03335},
  archiveprefix = {arXiv}
}

@inproceedings{agentgen,
  author = {Mengkang Hu and Pu Zhao and Can Xu and Qingfeng Sun and Jianguang Lou and Qingwei Lin and Ping Luo and Saravan Rajmohan},
  title = {{AgentGen}: Enhancing Planning Abilities for Large Language Model based Agent via Environment and Task Generation},
  booktitle = {Proceedings of the 31st ACM SIGKDD Conference on Knowledge Discovery and Data Mining},
  year = {2025},
  url = {https://arxiv.org/abs/2408.00764},
  eprint = {2408.00764},
  archiveprefix = {arXiv}
}

@inproceedings{sec,
  author = {Chen, Xiaoyin and Lu, Jiarui and Kim, Minsu and Zhang, Dinghuai and Tang, Jian and Pich{\'e}, Alexandre and Gontier, Nicolas and Bengio, Yoshua and Kamalloo, Ehsan},
  title = {Self-Evolving Curriculum for {LLM} Reasoning},
  booktitle = {Third Conference on Language Modeling},
  year = {2026},
  url = {https://arxiv.org/abs/2505.14970},
  eprint = {2505.14970},
  archiveprefix = {arXiv},
  primaryclass = {cs.AI}
}

@inproceedings{dreamgym,
  author = {Zhaorun Chen and Zhuokai Zhao and Kai Zhang and Bo Liu and Qi Qi and Yifan Wu and Tarun Kalluri and Sara Cao and Yuanhao Xiong and Haibo Tong and Huaxiu Yao and Hengduo Li and Jiacheng Zhu and Xian Li and Dawn Song and Bo Li and Jason Weston and Dat Huynh},
  title = {Scaling Agent Learning via Experience Synthesis},
  booktitle = {The Fourteenth International Conference on Learning Representations},
  year = {2026},
  url = {https://openreview.net/pdf?id=cf7qpBwttr},
  eprint = {2511.03773},
  archiveprefix = {arXiv}
}

@article{reward_refinement,
  author = {Youting Wang and Yuan Tang and Bowen Liu and Xuan Liu and Dingyan Shang},
  title = {When {LLM} Reward Design Fails: Diagnostic-Driven Refinement for Sparse Structured {RL}},
  journal = {arXiv preprint arXiv:2605.28918},
  year = {2026},
  doi = {10.48550/arXiv.2605.28918},
  url = {https://arxiv.org/abs/2605.28918v2},
  eprint = {2605.28918},
  archiveprefix = {arXiv}
}

@article{seal,
  author = {Hu, Yihao and Wen, Zhihao and Liu, Xiujin and Wang, Pan and Zhang, Xin and Wu, Wei},
  title = {{SEAL}: Synergistic Co-Evolution of Agents and Learning Environments},
  journal = {arXiv preprint arXiv:2605.24426},
  year = {2026},
  doi = {10.48550/arXiv.2605.24426},
  url = {https://arxiv.org/abs/2605.24426},
  eprint = {2605.24426},
  archiveprefix = {arXiv},
  primaryclass = {cs.CL}
}

@article{guo2026selfauthoredverification,
  author = {Diandian Guo and Cong Cao and Fangfang Yuan and Yingqi Wang and Yueshan Wang and Dakui Wang},
  title = {Self-Authored Verification Is Unreliable in Heuristic Self-Improving Agents},
  journal = {arXiv preprint arXiv:2607.24300},
  year = {2026},
  doi = {10.48550/arXiv.2607.24300},
  url = {https://arxiv.org/abs/2607.24300},
  eprint = {2607.24300},
  archiveprefix = {arXiv}
}

@incollection{good1966ultraintelligent,
  author = {Irving John Good},
  title = {Speculations Concerning the First Ultraintelligent Machine},
  booktitle = {Advances in Computers},
  volume = {6},
  pages = {31--88},
  publisher = {Academic Press},
  year = {1966},
  doi = {10.1016/S0065-2458(08)60418-0},
  url = {https://doi.org/10.1016/S0065-2458(08)60418-0}
}

@article{schmidhuber2003goedel,
  author = {J{\"u}rgen Schmidhuber},
  title = {{Goedel} Machines: Self-Referential Universal Problem Solvers Making Provably Optimal Self-Improvements},
  journal = {arXiv preprint arXiv:cs/0309048},
  year = {2003},
  doi = {10.48550/arXiv.cs/0309048},
  url = {https://arxiv.org/abs/cs/0309048v5},
  eprint = {cs/0309048},
  archiveprefix = {arXiv},
  note = {Version 5, revised 2006}
}

@inproceedings{hu2024adas,
  author = {Shengran Hu and Cong Lu and Jeff Clune},
  title = {Automated Design of Agentic Systems},
  booktitle = {International Conference on Learning Representations},
  year = {2025},
  doi = {10.48550/arXiv.2408.08435},
  url = {https://openreview.net/pdf?id=t9U3LW7JVX},
  eprint = {2408.08435},
  archiveprefix = {arXiv}
}

@article{lee2026metaharness,
  author = {Yoonho Lee and Roshen Nair and Qizheng Zhang and Kangwook Lee and Omar Khattab and Chelsea Finn},
  title = {{Meta-Harness}: End-to-End Optimization of Model Harnesses},
  journal = {arXiv preprint arXiv:2603.28052},
  year = {2026},
  doi = {10.48550/arXiv.2603.28052},
  url = {https://arxiv.org/abs/2603.28052},
  eprint = {2603.28052},
  archiveprefix = {arXiv}
}

@inproceedings{zelikman2024stop,
  author = {Zelikman, Eric and Lorch, Eliana and Mackey, Lester and Kalai, Adam Tauman},
  title = {Self-Taught Optimizer ({STOP}): Recursively Self-Improving Code Generation},
  booktitle = {First Conference on Language Modeling},
  year = {2024},
  doi = {10.48550/arXiv.2310.02304},
  url = {https://arxiv.org/abs/2310.02304},
  eprint = {2310.02304},
  archiveprefix = {arXiv}
}

@inproceedings{fernando2024promptbreeder,
  author = {Fernando, Chrisantha and Banarse, Dylan Sunil and Michalewski, Henryk and Osindero, Simon and Rockt{\"a}schel, Tim},
  title = {{Promptbreeder}: Self-Referential Self-Improvement via Prompt Evolution},
  booktitle = {Proceedings of the 41st International Conference on Machine Learning},
  series = {Proceedings of Machine Learning Research},
  volume = {235},
  pages = {13481--13544},
  publisher = {PMLR},
  year = {2024},
  url = {https://proceedings.mlr.press/v235/fernando24a.html}
}

@inproceedings{yin2025goedelagent,
  author = {Yin, Xunjian and Wang, Xinyi and Pan, Liangming and Lin, Li and Wan, Xiaojun and Wang, William Yang},
  title = {{G{\"o}del Agent}: A Self-Referential Agent Framework for Recursively Self-Improvement},
  booktitle = {Proceedings of the 63rd Annual Meeting of the Association for Computational Linguistics (Volume 1: Long Papers)},
  pages = {27890--27913},
  publisher = {Association for Computational Linguistics},
  year = {2025},
  doi = {10.18653/v1/2025.acl-long.1354},
  url = {https://aclanthology.org/2025.acl-long.1354/}
}

@article{robeyns2025sica,
  author = {Robeyns, Maxime and Szummer, Martin and Aitchison, Laurence},
  title = {A Self-Improving Coding Agent},
  journal = {arXiv preprint arXiv:2504.15228},
  year = {2025},
  doi = {10.48550/arXiv.2504.15228},
  url = {https://arxiv.org/abs/2504.15228},
  eprint = {2504.15228},
  archiveprefix = {arXiv}
}

@inproceedings{zhang2026darwin,
  author = {Jenny Zhang and Shengran Hu and Cong Lu and Robert Lange and Jeff Clune},
  title = {{Darwin G{\"{o}}del Machine}: Open-Ended Evolution of Self-Improving Agents},
  booktitle = {International Conference on Learning Representations (ICLR)},
  year = {2026},
  doi = {10.48550/arXiv.2505.22954},
  url = {https://iclr.cc/virtual/2026/poster/10007327},
  eprint = {2505.22954},
  archiveprefix = {arXiv}
}

@inproceedings{wang2025huxley,
  author = {Wenyi Wang and Piotr Pi{\k{e}}kos and Li, Nanbo and Firas Laakom and Yimeng Chen and Mateusz Ostaszewski and Mingchen Zhuge and J{\"{u}}rgen Schmidhuber},
  title = {{Huxley-G{\"{o}}del Machine}: Human-Level Coding Agent Development by an Approximation of the Optimal Self-Improving Machine},
  booktitle = {International Conference on Learning Representations (ICLR)},
  year = {2026},
  doi = {10.48550/arXiv.2510.21614},
  url = {https://arxiv.org/abs/2510.21614},
  eprint = {2510.21614},
  archiveprefix = {arXiv}
}

@article{zhang2026hyperagents,
  author = {Zhang, Jenny and Zhao, Bingchen and Yang, Wannan and Foerster, Jakob and Clune, Jeff and Jiang, Minqi and Devlin, Sam and Shavrina, Tatiana},
  title = {Hyperagents},
  journal = {arXiv preprint arXiv:2603.19461},
  year = {2026},
  doi = {10.48550/arXiv.2603.19461},
  url = {https://arxiv.org/abs/2603.19461},
  eprint = {2603.19461},
  archiveprefix = {arXiv}
}

@article{razzhigaev2026ouroboros,
  author = {Razzhigaev, Anton and Gritsaev, Andrei and Kaznacheev, Andrei and Dragunov, Nikita and Yampolskiy, Roman and Kuznetsov, Andrei},
  title = {{Ouroboros}: A Self-Developing Frontier Coding Agent with Reviewed Core Evolution},
  journal = {arXiv preprint arXiv:2608.08311},
  year = {2026},
  doi = {10.48550/arXiv.2608.08311},
  url = {https://arxiv.org/abs/2608.08311},
  eprint = {2608.08311},
  archiveprefix = {arXiv},
  primaryclass = {cs.SE}
}

@article{iacob2026redqueen,
  author = {Alex Iacob and Andrej Jovanovi{\'{c}} and William F. Shen and Daniel Burkhardt and Meghdad Kurmanji and Nurbek Tastan and Lorenzo Sani and Niccol{\`{o}} Alberto Elia Venanzi and Jiayi Nie and Ambroise Odonnat and Zeyu Cao and Bill Marino and Xinchi Qiu and Rika Antonova and Nicholas D. Lane},
  title = {The {Red Queen G{\"o}del Machine}: Co-Evolving Agents and Their Evaluators},
  journal = {arXiv preprint arXiv:2606.26294},
  year = {2026},
  doi = {10.48550/arXiv.2606.26294},
  url = {https://arxiv.org/abs/2606.26294},
  eprint = {2606.26294},
  archiveprefix = {arXiv}
}

@article{zhang2026whogrades,
  author = {Zhang, Xing and Wang, Guanghui and Cui, Yanwei and Li, Ziyuan and Qiu, Wei and Zhu, Bing and He, Peiyang},
  title = {Who Grades the Grader? Co-Evolving Evaluation Metrics and Skills for Self-Improving {LLM} Agents},
  journal = {arXiv preprint arXiv:2607.12790},
  year = {2026},
  doi = {10.48550/arXiv.2607.12790},
  url = {https://arxiv.org/abs/2607.12790},
  eprint = {2607.12790},
  archiveprefix = {arXiv}
}

@article{boiko2023coscientist,
  author = {Boiko, Daniil A. and MacKnight, Robert and Kline, Ben and Gomes, Gabe},
  title = {Autonomous Chemical Research with Large Language Models},
  journal = {Nature},
  volume = {624},
  number = {7992},
  pages = {570--578},
  year = {2023},
  doi = {10.1038/s41586-023-06792-0},
  url = {https://doi.org/10.1038/s41586-023-06792-0}
}

@article{szymanski2023alab,
  author = {Szymanski, Nathan J. and Rendy, Bernardus and Fei, Yuxing and Kumar, Rishi E. and He, Tanjin and Milsted, David and McDermott, Matthew J. and Gallant, Max and Cubuk, Ekin Dogus and Merchant, Amil and Kim, Haegyeom and Jain, Anubhav and Bartel, Christopher J. and Persson, Kristin and Zeng, Yan and Ceder, Gerbrand},
  title = {An Autonomous Laboratory for the Accelerated Synthesis of Inorganic Materials},
  journal = {Nature},
  volume = {624},
  number = {7990},
  pages = {86--91},
  year = {2023},
  doi = {10.1038/s41586-023-06734-w},
  url = {https://doi.org/10.1038/s41586-023-06734-w}
}

@article{swanson2025virtuallab,
  author = {Swanson, Kyle and Wu, Wesley and Bulaong, Nash L. and Pak, John E. and Zou, James},
  title = {The Virtual Lab of {AI} Agents Designs New {SARS-CoV-2} Nanobodies},
  journal = {Nature},
  volume = {646},
  pages = {716--723},
  year = {2025},
  doi = {10.1038/s41586-025-09442-9},
  url = {https://doi.org/10.1038/s41586-025-09442-9}
}

@article{huang2026evobench,
  author = {Huang, Lisheng and Yang, Chen and Zhou, Hao and Song, Huatong and Chen, Zongchao and Le, Ran and Song, Yang and Zhao, Wayne Xin and Zhang, Tao},
  title = {{Evo-Bench}: Can Language Models Improve Agent Harness?},
  journal = {arXiv preprint arXiv:2608.09096},
  year = {2026},
  doi = {10.48550/arXiv.2608.09096},
  url = {https://arxiv.org/abs/2608.09096},
  eprint = {2608.09096},
  archiveprefix = {arXiv}
}

@article{ursekar2026harnessoptbench,
  author = {Ursekar, Varun and Shanker, Apaar and Maurya, Yash and Yasser, Shehab and Kalmath, Vijay S. and Chatrath, Veronica and Xue, Yuan},
  title = {{HarnessOpt-Bench}: Evaluating {LLMs} at Harness Optimization},
  journal = {arXiv preprint arXiv:2608.06301},
  year = {2026},
  doi = {10.48550/arXiv.2608.06301},
  url = {https://arxiv.org/abs/2608.06301},
  eprint = {2608.06301},
  archiveprefix = {arXiv}
}

@article{wei2025evomemory,
  author = {Wei, Tianxin and Sachdeva, Noveen and Coleman, Benjamin and He, Zhankui and Bei, Yuanchen and Ning, Xuying and Ai, Mengting and Li, Yunzhe and He, Jingrui and Chi, Ed H. and Wang, Chi and Chen, Shuo and Pereira, Fernando and Kang, Wang-Cheng and Cheng, Derek Zhiyuan},
  title = {{Evo-Memory}: Benchmarking {LLM} Agent Test-time Learning with Self-Evolving Memory},
  journal = {arXiv preprint arXiv:2511.20857},
  year = {2025},
  doi = {10.48550/arXiv.2511.20857},
  url = {https://arxiv.org/abs/2511.20857},
  eprint = {2511.20857},
  archiveprefix = {arXiv}
}

@article{zhang2026skillflow,
  author = {Zhang, Ziao and Shi, Kou and Huang, Shiting and Nie, Avery and Zeng, Yu and Zhao, Yiming and Fang, Zhen and Su, Qishen and Qiu, Haibo and Yang, Wei and Ren, Qingnan and Zou, Shun and Huang, Wenxuan and Chen, Lin and Chen, Zehui and Zhao, Feng},
  title = {{SkillFlow}: Benchmarking Lifelong Skill Discovery and Evolution for Autonomous Agents},
  journal = {arXiv preprint arXiv:2604.17308},
  year = {2026},
  doi = {10.48550/arXiv.2604.17308},
  url = {https://arxiv.org/abs/2604.17308},
  eprint = {2604.17308},
  archiveprefix = {arXiv}
}

@article{gao2026evoagentbench,
  author = {Xingze Gao and Chuanrui Hu and Hongda Chen and Pengfei Yao and Zhao Wang and Yi Bai and Zhengwei Wu and Yunyun Han and Xiaofeng Cong and Jie Gui and Yafeng Deng and Teng Li},
  title = {{EvoAgentBench}: Benchmarking Agent Self-Evolution via Ability Transfer},
  journal = {arXiv preprint arXiv:2607.05202},
  year = {2026},
  doi = {10.48550/arXiv.2607.05202},
  url = {https://arxiv.org/abs/2607.05202},
  eprint = {2607.05202},
  archiveprefix = {arXiv}
}

@inproceedings{he2026memoryarena,
  author = {He, Zexue and Wang, Yu and Zhi, Churan and Hu, Yuanzhe and Chen, Tzu-Ping and Yin, Lang and Chen, Ze and Wu, Tong Arthur and Ouyang, Siru and Wang, Zihan and Pei, Jiaxin and McAuley, Julian and Choi, Yejin and Pentland, Alex},
  title = {{MemoryArena}: Benchmarking Agent Memory in Interdependent Multi-Session Agentic Tasks},
  booktitle = {Proceedings of the 43rd International Conference on Machine Learning},
  year = {2026},
  url = {https://arxiv.org/abs/2602.16313},
  eprint = {2602.16313},
  archiveprefix = {arXiv}
}

@article{wang2026evomembench,
  author = {Wang, Yuyao and Zhang, Zhongjian and Chi, Mo and Yu, Kaichi and Li, Yuhan and Peng, Miao and Tong, Bing and Zhang, Chen and Zhou, Yan and Li, Jia},
  title = {{EvoMemBench}: Benchmarking Agent Memory from a Self-Evolving Perspective},
  journal = {arXiv preprint arXiv:2605.18421},
  year = {2026},
  doi = {10.48550/arXiv.2605.18421},
  url = {https://arxiv.org/abs/2605.18421},
  eprint = {2605.18421},
  archiveprefix = {arXiv}
}

@article{gottweis2026coscientist,
  author = {Gottweis, Juraj and Weng, Wei-Hung and Daryin, Alexander
             and Tu, Tao and Sirkovic, Petar and Myaskovsky, Artiom
             and Glowaty, Grzegorz and Weissenberger, Felix
             and Orlandi, Alessio and others},
  title = {Accelerating Scientific Discovery with {Co-Scientist}},
  journal = {Nature},
  volume = {655},
  pages = {487--496},
  year = {2026},
  doi = {10.1038/s41586-026-10644-y},
  url = {https://doi.org/10.1038/s41586-026-10644-y}
}

@article{ghareeb2026robin,
  author = {Ghareeb, Ali E. and Chang, Benjamin and Mitchener, Ludovico
             and Yiu, Angela and Szostkiewicz, Caralyn J.
             and Shved, Dmytro and Gyimesi, Gavin J.
             and Laurent, Jon M. and Wright, Samantha M.
             and Razzak, Muhammed T. and White, Andrew D.
             and Finnemann, Silvia C. and Hinks, Michaela M.
             and Rodriques, Samuel G.},
  title = {A Multi-Agent System for Automating Scientific Discovery},
  journal = {Nature},
  volume = {655},
  pages = {497--505},
  year = {2026},
  doi = {10.1038/s41586-026-10652-y},
  url = {https://doi.org/10.1038/s41586-026-10652-y}
}

@article{xue2026pastbench,
  author = {Xue, Shuhan and Ding, Zixin and Shen, Yichen and Wang, Yinjie and Yin, Zhenfei and Wu, Yingcheng and Chen, Yuxin and Wang, Mengdi and Yang, Ling},
  title = {{PAST-Bench}: Benchmarking the Foundations of Recursive Self-Improvement in Personal Agents},
  journal = {arXiv preprint arXiv:2608.04003},
  year = {2026},
  doi = {10.48550/arXiv.2608.04003},
  url = {https://arxiv.org/abs/2608.04003},
  eprint = {2608.04003},
  archiveprefix = {arXiv}
}

@article{chi2026ai4aibench,
  author = {Chi, Yizhe and Li, Wenyi and Hong, Deyao and Wang, Xiaoqiu and Gao, Mingju and Yang, Kaisen and He, Bingxiang and Zheng, Youjie and Xiao, Calvin and Na, Qinhuai},
  title = {{AI4AI-Bench}: Benchmarking {LLM} Agents in Algorithmic Design for Recursive Self-Improvement},
  journal = {arXiv preprint arXiv:2608.20318},
  year = {2026},
  doi = {10.48550/arXiv.2608.20318},
  url = {https://arxiv.org/abs/2608.20318},
  eprint = {2608.20318},
  archiveprefix = {arXiv}
}

@misc{rsiexam2026,
  author = {{RSI-Exam Team}},
  title = {{RSI-Exam}: Benchmarking Recursive Self-Improvement through Executable Research},
  year = {2026},
  url = {https://github.com/aiming-lab/RSI-Exam},
  howpublished = {Official software repository and evaluation protocol},
  note = {Accessed: 2026-10-19}
}

@article{zhang2025evoflow,
  author = {Guibin Zhang and Kaijie Chen and Guancheng Wan and Heng Chang and Hong Cheng and Kun Wang and Shuyue Hu and Lei Bai},
  title = {{EvoFlow}: Evolving Diverse Agentic Workflows On The Fly},
  journal = {arXiv preprint arXiv:2502.07373},
  year = {2025},
  doi = {10.48550/arXiv.2502.07373},
  url = {https://arxiv.org/abs/2502.07373},
  eprint = {2502.07373},
  archiveprefix = {arXiv}
}

@article{yang2026autoskill,
  author = {Yutao Yang and Junsong Li and Qianjun Pan and Bihao Zhan and Yuxuan Cai and Lin Du and Jie Zhou and Kai Chen and Qin Chen and Xin Li and Bo Zhang and Liang He},
  title = {{AutoSkill}: Experience-Driven Lifelong Learning via Skill Self-Evolution},
  journal = {arXiv preprint arXiv:2603.01145},
  year = {2026},
  doi = {10.48550/arXiv.2603.01145},
  url = {https://arxiv.org/abs/2603.01145},
  eprint = {2603.01145},
  archiveprefix = {arXiv}
}

@inproceedings{liu2026skillforge,
  author = {Xingyan Liu and Xiyue Luo and Linyu Li and Ganghong Huang and Jianfeng Liu and Honglin Qiao},
  title = {{SkillForge}: Forging Domain-Specific, Self-Evolving Agent Skills in Cloud Technical Support},
  booktitle = {ACM SIGIR 2026 Industry Track},
  publisher = {Association for Computing Machinery},
  year = {2026},
  doi = {10.1145/3805712.3808466},
  url = {https://arxiv.org/abs/2604.08618},
  eprint = {2604.08618},
  archiveprefix = {arXiv}
}

@article{ma2026skillclaw,
  author = {Ziyu Ma and Shidong Yang and Yuxiang Ji and Xucong Wang and Yong Wang and Yiming Hu and Tongwen Huang and Xiangxiang Chu},
  title = {{SkillClaw}: Let Skills Evolve Collectively with Agentic Evolver},
  journal = {arXiv preprint arXiv:2604.08377},
  year = {2026},
  doi = {10.48550/arXiv.2604.08377},
  url = {https://arxiv.org/abs/2604.08377},
  eprint = {2604.08377},
  archiveprefix = {arXiv}
}

@inproceedings{wang2026sage,
  author = {Wang, Jiongxiao and Yan, Qiaojing and Wang, Yawei and Tian, Yijun and Mishra, Soumya Smruti and Xu, Zhichao and Gandhi, Megha and Xu, Panpan and Cheong, Lin Lee},
  title = {Reinforcement Learning for Self-Improving Agent with Skill Library},
  booktitle = {Proceedings of the 64th Annual Meeting of the {A}ssociation for {C}omputational {L}inguistics (Volume 1: Long Papers)},
  pages = {1529--1550},
  publisher = {Association for Computational Linguistics},
  year = {2026},
  doi = {10.18653/v1/2026.acl-long.69},
  url = {https://aclanthology.org/2026.acl-long.69/}
}

@article{shi2026skill1,
  author = {Yaorui Shi and Yuxin Chen and Zhengxi Lu and Yuchun Miao and Shugui Liu and Qi Gu and Xunliang Cai and Xiang Wang and An Zhang},
  title = {{Skill1}: Unified Evolution of Skill-Augmented Agents via Reinforcement Learning},
  journal = {arXiv preprint arXiv:2605.06130},
  year = {2026},
  doi = {10.48550/arXiv.2605.06130},
  url = {https://arxiv.org/abs/2605.06130},
  eprint = {2605.06130},
  archiveprefix = {arXiv}
}

@article{xia2026skillrl,
  author = {Peng Xia and Jianwen Chen and Hanyang Wang and Jiaqi Liu and Kaide Zeng and Yu Wang and Siwei Han and Yiyang Zhou and Xujiang Zhao and Haifeng Chen and Zeyu Zheng and Cihang Xie and Huaxiu Yao},
  title = {{SkillRL}: Evolving Agents via Recursive Skill-Augmented Reinforcement Learning},
  journal = {arXiv preprint arXiv:2602.08234},
  year = {2026},
  doi = {10.48550/arXiv.2602.08234},
  url = {https://arxiv.org/abs/2602.08234},
  eprint = {2602.08234},
  archiveprefix = {arXiv}
}

@inproceedings{zhang2026coevoskills,
  author = {Hanrong Zhang and Shicheng Fan and Henry Peng Zou and Yankai Chen and Zhenting Wang and Jiayu Zhou and Chengze Li and Wei-Chieh Huang and Yifei Yao and Kening Zheng and Xue (Steve) Liu and Xiaoxiao Li and Philip S. Yu},
  title = {{CoEvoSkills}: Self-Evolving Agent Skills via Co-Evolutionary Verification},
  booktitle = {Conference on Language Modeling},
  year = {2026},
  url = {https://arxiv.org/abs/2604.01687},
  eprint = {2604.01687},
  archiveprefix = {arXiv},
  note = {Accepted at COLM 2026; bibliographic metadata from the authors' arXiv record}
}

@article{zhang2026memskill,
  author = {Haozhen Zhang and Quanyu Long and Jianzhu Bao and Tao Feng and Weizhi Zhang and Haodong Yue and Wenya Wang},
  title = {{MemSkill}: Learning and Evolving Memory Skills for Self-Evolving Agents},
  journal = {arXiv preprint arXiv:2602.02474},
  year = {2026},
  doi = {10.48550/arXiv.2602.02474},
  url = {https://arxiv.org/abs/2602.02474},
  eprint = {2602.02474},
  archiveprefix = {arXiv}
}

@inproceedings{yu2026agemem,
  author = {Yu, Yi and Yao, Liuyi and Xie, Yuexiang and Tan, Qingquan and Feng, Jiaqi and Li, Yaliang and Wu, Libing},
  title = {Agentic Memory: Learning Unified Long-Term and Short-Term Memory Management for Large Language Model Agents},
  booktitle = {Proceedings of the 64th Annual Meeting of the {A}ssociation for {C}omputational {L}inguistics (Volume 1: Long Papers)},
  pages = {21457--21483},
  publisher = {Association for Computational Linguistics},
  year = {2026},
  doi = {10.18653/v1/2026.acl-long.981},
  url = {https://aclanthology.org/2026.acl-long.981/}
}

@article{zhang2026memrl,
  author = {Shengtao Zhang and Jiaqian Wang and Ruiwen Zhou and Junwei Liao and Yuchen Feng and Zhuo Li and Yujie Zheng and Weinan Zhang and Ying Wen and Zhiyu Li and Feiyu Xiong and Yutao Qi and Bo Tang and Muning Wen},
  title = {{MemRL}: Self-Evolving Agents via Runtime Reinforcement Learning on Episodic Memory},
  journal = {arXiv preprint arXiv:2601.03192},
  year = {2026},
  doi = {10.48550/arXiv.2601.03192},
  url = {https://arxiv.org/abs/2601.03192},
  eprint = {2601.03192},
  archiveprefix = {arXiv}
}

@article{karten2026continualharness,
  author = {Seth Karten and Joel Zhang and Upaa, Jr., Tersoo and Ruirong Feng and Wenzhe Li and Chengshuai Shi and Chi Jin and Kiran Vodrahalli},
  title = {Continual Harness: Online Adaptation for Self-Improving Foundation Agents},
  journal = {arXiv preprint arXiv:2605.09998},
  year = {2026},
  doi = {10.48550/arXiv.2605.09998},
  url = {https://arxiv.org/abs/2605.09998},
  eprint = {2605.09998},
  archiveprefix = {arXiv}
}

@article{yang2026frontisma1,
  author = {Yang, Junlin and Jiang, Che and Fu, Yu and Luo, Tianwei and Ren, Can and Wang, Weizhi and Zhao, Kaikai and Liu, Hongyi and Zuo, Yuxin and Wang, Yuru and Fan, Yuchen and Tian, Kai and Yuan, Zhenzhao and Lin, Xiaojian and Sheng, Li and Qiang, Rushi and Jia, Guoli and Lv, Xingtai and Hua, Ermo and Lei, Dianqiao and Sun, Youbang and Ding, Ning and Zhou, Bowen and Zhang, Kaiyan},
  title = {{Frontis-MA1}: Training an {AI4AI} Model towards Recursive Self-Improvement in Machine Learning Engineering},
  journal = {arXiv preprint arXiv:2607.28568},
  year = {2026},
  doi = {10.48550/arXiv.2607.28568},
  url = {https://arxiv.org/abs/2607.28568},
  eprint = {2607.28568},
  archiveprefix = {arXiv}
}

@inproceedings{meta_rewarding,
  author = {Wu, Tianhao and Yuan, Weizhe and Golovneva, Olga and Xu, Jing and Tian, Yuandong and Jiao, Jiantao and Weston, Jason E. and Sukhbaatar, Sainbayar},
  title = {Meta-Rewarding Language Models: Self-Improving Alignment with {LLM-as-a-Meta-Judge}},
  booktitle = {Proceedings of the 2025 Conference on Empirical Methods in Natural Language Processing},
  pages = {11537--11554},
  publisher = {Association for Computational Linguistics},
  year = {2025},
  doi = {10.18653/v1/2025.emnlp-main.583},
  url = {https://aclanthology.org/2025.emnlp-main.583/}
}

@article{liu2026mendel,
  author = {Changzhi Liu and Yilun Liu and Sikuan Yan and Volker Tresp and Yunpu Ma},
  title = {{Mendel G{\"{o}}del Machine}: Recursive Self-Improving Coding Agents via Comparative Evolution},
  journal = {arXiv preprint arXiv:2608.07645},
  year = {2026},
  doi = {10.48550/arXiv.2608.07645},
  url = {https://arxiv.org/abs/2608.07645},
  eprint = {2608.07645},
  archiveprefix = {arXiv}
}

@article{liu2026evox,
  author = {Liu, Shu and Agarwal, Shubham and Maheswaran, Monishwaran and Cemri, Mert and Li, Zhifei and Mang, Qiuyang and Naren, Ashwin and Boneh, Ethan and Cheng, Audrey and Pan, Melissa Z. and Du, Alexander and Keutzer, Kurt and Cheung, Alvin and Dimakis, Alexandros G. and Sen, Koushik and Zaharia, Matei and Stoica, Ion},
  title = {{EvoX}: Meta-Evolution for Automated Discovery},
  journal = {arXiv preprint arXiv:2602.23413},
  year = {2026},
  doi = {10.48550/arXiv.2602.23413},
  url = {https://arxiv.org/abs/2602.23413},
  eprint = {2602.23413},
  archiveprefix = {arXiv}
}

@article{zhou2026hsi,
  author = {Tailin Zhou},
  title = {Hierarchical Self-Improvement: A Framework for Task-Specific Evolvable Agent Harnesses},
  journal = {arXiv preprint arXiv:2608.08466},
  year = {2026},
  doi = {10.48550/arXiv.2608.08466},
  url = {https://arxiv.org/abs/2608.08466},
  eprint = {2608.08466},
  archiveprefix = {arXiv}
}

@article{wang2026metaskillevolve,
  author = {Zefeng Wang and Minxi Yan and Jinhe Bi and Sikuan Yan and Volker Tresp and Yunpu Ma},
  title = {{MetaSkill-Evolve}: Recursive Self-Improvement of {LLM} Agents via Two-Timescale Meta-Skill Evolution},
  journal = {arXiv preprint arXiv:2607.05297},
  year = {2026},
  doi = {10.48550/arXiv.2607.05297},
  url = {https://arxiv.org/abs/2607.05297},
  eprint = {2607.05297},
  archiveprefix = {arXiv}
}

@article{liu2026escher,
  author = {Ziyang Liu and Xinyan Guo and Xuchen Wei and Han Hao and Liu Yang},
  title = {{Escher-Loop}: Mutual Evolution by Closed-Loop Self-Referential Optimization},
  journal = {arXiv preprint arXiv:2604.23472},
  year = {2026},
  doi = {10.48550/arXiv.2604.23472},
  url = {https://arxiv.org/abs/2604.23472},
  eprint = {2604.23472},
  archiveprefix = {arXiv}
}

@article{novikov2025alphaevolve,
  author = {Novikov, Alexander and V{\~u}, Ng{\^a}n and Eisenberger, Marvin and Dupont, Emilien and Huang, Po-Sen and Wagner, Adam Zsolt and Shirobokov, Sergey and Kozlovskii, Borislav and Ruiz, Francisco J. R. and Mehrabian, Abbas and Kumar, M. Pawan and See, Abigail and Chaudhuri, Swarat and Holland, George and Davies, Alex and Nowozin, Sebastian and Kohli, Pushmeet and Balog, Matej},
  title = {{AlphaEvolve}: A Coding Agent for Scientific and Algorithmic Discovery},
  journal = {arXiv preprint arXiv:2506.13131},
  year = {2025},
  doi = {10.48550/arXiv.2506.13131},
  url = {https://arxiv.org/abs/2506.13131},
  eprint = {2506.13131},
  archiveprefix = {arXiv}
}

@article{jiang2025aide,
  author = {Jiang, Zhengyao and Schmidt, Dominik and Srikanth, Dhruv and Xu, Dixing and Kaplan, Ian and Jacenko, Deniss and Wu, Yuxiang},
  title = {{AIDE}: {AI-Driven} Exploration in the Space of Code},
  journal = {arXiv preprint arXiv:2502.13138},
  year = {2025},
  doi = {10.48550/arXiv.2502.13138},
  url = {https://arxiv.org/abs/2502.13138},
  eprint = {2502.13138},
  archiveprefix = {arXiv}
}

@article{xia2026poise,
  author = {Sirui Xia and Yikai Zhang and Aili Chen and Siye Wu and Siyu Yuan and Yanghua Xiao},
  title = {From {AI} Assistant to {AI} Scientist: Autonomous Discovery of {LLM-RL} Algorithms with {LLM} Agents},
  journal = {arXiv preprint arXiv:2603.23951},
  year = {2026},
  doi = {10.48550/arXiv.2603.23951},
  url = {https://arxiv.org/abs/2603.23951},
  eprint = {2603.23951},
  archiveprefix = {arXiv}
}

@article{li2026skillsbench,
  author = {Xiangyi Li and Yimin Liu and Wenbo Chen and Bingran You and Zonglin Di and Yifeng He and Shenghan Zheng and Kyoung Whan Choe and Jiankai Sun and Shuyi Wang and Chujun Tao and Binxu Li and Xuandong Zhao and Hejia Geng and Xiaojun Wu and Junwei Zhou and Xiaokun Chen and Hanwen Xing and Yubo Li and Qunhong Zeng and Di Wang and Yuanli Wang and Roey Ben Chaim and Penghao Jiang and Haotian Shen and Luyang Kong and Xinyi Liu and Runhui Wang and Xuanqing Liu and Jiachen Li and Xin Lan and Yueqian Lin and Wengao Ye and Junwei He and Songlin Li and Yue Zhang and Yipeng Gao and Yijiang Li and Ze Ma and Liqiang Jing and Tianyu Wang and Kaixin Li and Yiqi Xue and Haoran Lyu and Yizhuo He and Yuchen Tian and Shutong Wu and Bowei Wang and Yixuan Gao and Bo Chen and Litong Liu and Sikai Cheng and Jiajun Bao and Shuaicheng Tong and Shuwen Xu and Terry Yue Zhuo and Tinghan Ye and Qi Qi and Miao Li and Longtai Liao and Zelin Tan and Chang Shi and Xilin Tang and Srinath Tankasala and Boqin Yuan and Yaoyao Qian and Jianhong Tu and Chenguang Wang and Yizhou Sun and Wei Wang and Aaron Taylor and Ziyue Yang and Changkun Guan and Zhikang Dong and Xinyu Zhang and Steven Dillmann and Lee, {Han-chung} and Dawn Song},
  title = {{SkillsBench}: Benchmarking How Well Agent Skills Work Across Diverse Tasks},
  journal = {arXiv preprint arXiv:2602.12670},
  year = {2026},
  doi = {10.48550/arXiv.2602.12670},
  url = {https://arxiv.org/abs/2602.12670},
  eprint = {2602.12670},
  archiveprefix = {arXiv}
}

@inproceedings{swe_gym,
  author = {Jiayi Pan and Xingyao Wang and Graham Neubig and Navdeep Jaitly and Heng Ji and Alane Suhr and Yizhe Zhang},
  title = {Training Software Engineering Agents and Verifiers with {SWE-Gym}},
  booktitle = {Proceedings of the 42nd International Conference on Machine Learning},
  volume = {267},
  pages = {47717--47737},
  year = {2025},
  url = {https://proceedings.mlr.press/v267/pan25g.html}
}

@inproceedings{cai2024large,
  author = {Tianle Cai and Xuezhi Wang and Tengyu Ma and Xinyun Chen and Denny Zhou},
  title = {Large Language Models as Tool Makers},
  booktitle = {International Conference on Learning Representations},
  year = {2024},
  url = {https://openreview.net/forum?id=qV83K9d5WB},
  eprint = {2305.17126},
  archiveprefix = {arXiv}
}

@inproceedings{yuan2024craft,
  author = {Lifan Yuan and Yangyi Chen and Xingyao Wang and Yi R. Fung and Hao Peng and Heng Ji},
  title = {{CRAFT}: Customizing {LLMs} by Creating and Retrieving from Specialized Toolsets},
  booktitle = {International Conference on Learning Representations},
  year = {2024},
  url = {https://openreview.net/forum?id=G0vdDSt9XM},
  eprint = {2309.17428},
  archiveprefix = {arXiv}
}

@article{ReSTEM,
  author = {Avi Singh and John D. Co-Reyes and Rishabh Agarwal and Ankesh Anand and Piyush Patil and Xavier Garcia and Peter J. Liu and James Harrison and Jaehoon Lee and Kelvin Xu and Aaron Parisi and Abhishek Kumar and Alex Alemi and Alex Rizkowsky and Azade Nova and Ben Adlam and Bernd Bohnet and Gamaleldin Elsayed and Hanie Sedghi and Igor Mordatch and Isabelle Simpson and Izzeddin Gur and Jasper Snoek and Jeffrey Pennington and Jiri Hron and Kathleen Kenealy and Kevin Swersky and Kshiteej Mahajan and Laura Culp and Lechao Xiao and Maxwell L. Bileschi and Noah Constant and Roman Novak and Rosanne Liu and Tris Warkentin and Yundi Qian and Yamini Bansal and Ethan Dyer and Behnam Neyshabur and Jascha Sohl-Dickstein and Noah Fiedel},
  title = {Beyond Human Data: Scaling Self-Training for Problem-Solving with Language Models},
  journal = {Transactions on Machine Learning Research},
  year = {2024},
  doi = {10.48550/arXiv.2312.06585},
  url = {https://arxiv.org/abs/2312.06585},
  eprint = {2312.06585},
  archiveprefix = {arXiv}
}

@inproceedings{SCoRe,
  author = {Aviral Kumar and Vincent Zhuang and Rishabh Agarwal and Yi Su and John D. Co-Reyes and Avi Singh and Kate Baumli and Shariq Iqbal and Colton Bishop and Rebecca Roelofs and Lei M. Zhang and Kay McKinney and Disha Shrivastava and Cosmin Paduraru and George Tucker and Doina Precup and Feryal Behbahani and Aleksandra Faust},
  title = {Training Language Models to Self-Correct via Reinforcement Learning},
  booktitle = {International Conference on Learning Representations},
  year = {2025},
  doi = {10.48550/arXiv.2409.12917},
  url = {https://arxiv.org/abs/2409.12917}
}

@inproceedings{SearchR1,
  author = {Bowen Jin and Hansi Zeng and Zhenrui Yue and Jinsung Yoon and Sercan {\"O}. Ar{\i}k and Dong Wang and Hamed Zamani and Jiawei Han},
  title = {{Search-R1}: Training {LLMs} to Reason and Leverage Search Engines with Reinforcement Learning},
  booktitle = {Conference on Language Modeling},
  year = {2025},
  url = {https://openreview.net/forum?id=Rwhi91ideu},
  eprint = {2503.09516},
  archiveprefix = {arXiv}
}

@inproceedings{rlve,
  author = {Zhiyuan Zeng and Hamish Ivison and Yiping Wang and Lifan Yuan and Shuyue Stella Li and Zhuorui Ye and Siting Li and Jacqueline He and Runlong Zhou and Tong Chen and Chenyang Zhao and Yulia Tsvetkov and Simon Shaolei Du and Natasha Jaques and Hao Peng and Pang Wei Koh and Hannaneh Hajishirzi},
  title = {{RLVE}: Scaling Up Reinforcement Learning for Language Models with Adaptive Verifiable Environments},
  booktitle = {Proceedings of the 43rd International Conference on Machine Learning},
  series = {Proceedings of Machine Learning Research},
  volume = {306},
  pages = {153406--153428},
  publisher = {PMLR},
  year = {2026},
  url = {https://proceedings.mlr.press/v306/zeng26j.html},
  eprint = {2511.07317},
  archiveprefix = {arXiv}
}

@inproceedings{webevolver,
  author = {Tianqing Fang and Hongming Zhang and Zhisong Zhang and Kaixin Ma and Wenhao Yu and Haitao Mi and Dong Yu},
  title = {{WebEvolver}: Enhancing Web Agent Self-Improvement with Co-evolving World Model},
  booktitle = {Proceedings of the 2025 Conference on Empirical Methods in Natural Language Processing},
  pages = {8959--8975},
  publisher = {Association for Computational Linguistics},
  year = {2025},
  doi = {10.18653/v1/2025.emnlp-main.454},
  url = {https://aclanthology.org/2025.emnlp-main.454/},
  eprint = {2504.21024},
  archiveprefix = {arXiv}
}

@inproceedings{huang2026rzero,
  author = {Chengsong Huang and Wenhao Yu and Xiaoyang Wang and Hongming Zhang and Zongxia Li and Ruosen Li and Jiaxin Huang and Haitao Mi and Dong Yu},
  title = {{R-Zero}: Self-Evolving Reasoning {LLM} from Zero Data},
  booktitle = {International Conference on Learning Representations},
  year = {2026},
  url = {https://openreview.net/forum?id=96apU6YzSO},
  eprint = {2508.05004},
  archiveprefix = {arXiv}
}

@inproceedings{covolve,
  author = {Sygkounas, Alkis and Hazra, Rishi and Persson, Andreas and {Zuidberg dos Martires}, Pedro and Loutfi, Amy},
  title = {{COvolve}: Adversarial Co-Evolution of Large-Language-Model-Generated Policies and Environments via Two-Player Zero-Sum Game},
  booktitle = {Proceedings of the Genetic and Evolutionary Computation Conference},
  series = {GECCO '26},
  pages = {338--347},
  publisher = {ACM},
  year = {2026},
  doi = {10.1145/3795095.3805144},
  url = {https://arxiv.org/abs/2603.28386},
  eprint = {2603.28386},
  archiveprefix = {arXiv}
}

@article{comap,
  author = {Liu, Youwei and Wang, Jian and Wang, Hanlin and Li, Wenjie},
  title = {{CoMAP}: Co-Evolving World Models and Agent Policies for {LLM} Agents},
  journal = {arXiv preprint arXiv:2606.02372},
  year = {2026},
  doi = {10.48550/arXiv.2606.02372},
  url = {https://arxiv.org/abs/2606.02372},
  eprint = {2606.02372},
  archiveprefix = {arXiv},
  primaryclass = {cs.AI}
}

@article{lu2026metaagentchallenge,
  author = {Xinyu Lu and Tianshu Wang and Pengbo Wang and Zujie Wen and Zhiqiang Zhang and Jun Zhou and Boxi Cao and Yaojie Lu and Hongyu Lin and Xianpei Han and Le Sun},
  title = {The Meta-Agent Challenge: Are Current Agents Capable of Autonomous Agent Development?},
  journal = {arXiv preprint arXiv:2606.04455},
  year = {2026},
  doi = {10.48550/arXiv.2606.04455},
  url = {https://arxiv.org/abs/2606.04455},
  eprint = {2606.04455},
  archiveprefix = {arXiv}
}

@inproceedings{huang2024mlagentbench,
  author = {Qian Huang and Jian Vora and Percy Liang and Jure Leskovec},
  title = {{MLAgentBench}: Evaluating Language Agents on Machine Learning Experimentation},
  booktitle = {Proceedings of the 41st International Conference on Machine Learning},
  series = {Proceedings of Machine Learning Research},
  volume = {235},
  pages = {20271--20309},
  publisher = {PMLR},
  year = {2024},
  url = {https://proceedings.mlr.press/v235/huang24y.html}
}

@inproceedings{chan2025mlebench,
  author = {Jun Shern Chan and Neil Chowdhury and Oliver Jaffe and James Aung and Dane Sherburn and Evan Mays and Giulio Starace and Kevin Liu and Leon Maksin and Tejal Patwardhan and Lilian Weng and Aleksander M{\k{a}}dry},
  title = {{MLE-bench}: Evaluating Machine Learning Agents on Machine Learning Engineering},
  booktitle = {International Conference on Learning Representations},
  year = {2025},
  url = {https://arxiv.org/abs/2410.07095},
  eprint = {2410.07095},
  archiveprefix = {arXiv}
}

@article{nathani2025mlgym,
  author = {Deepak Nathani and Lovish Madaan and Nicholas Roberts and Nikolay Bashlykov and Ajay Menon and Vincent Moens and Amar Budhiraja and Despoina Magka and Vladislav Vorotilov and Gaurav Chaurasia and Dieuwke Hupkes and Ricardo Silveira Cabral and Tatiana Shavrina and Jakob Foerster and Yoram Bachrach and William Yang Wang and Roberta Raileanu},
  title = {{MLGym}: A New Framework and Benchmark for Advancing {AI} Research Agents},
  journal = {arXiv preprint arXiv:2502.14499},
  year = {2025},
  doi = {10.48550/arXiv.2502.14499},
  url = {https://arxiv.org/abs/2502.14499},
  eprint = {2502.14499},
  archiveprefix = {arXiv}
}

@article{rank2026posttrainbench,
  author = {Ben Rank and Hardik Bhatnagar and Ameya Prabhu and Shira Eisenberg and Karina Nguyen and Matthias Bethge and Maksym Andriushchenko},
  title = {{PostTrainBench}: Can {LLM} Agents Automate {LLM} Post-Training?},
  journal = {arXiv preprint arXiv:2603.08640},
  year = {2026},
  doi = {10.48550/arXiv.2603.08640},
  url = {https://arxiv.org/abs/2603.08640},
  eprint = {2603.08640},
  archiveprefix = {arXiv}
}

@inproceedings{NEURIPS2023_7cc1005e,
  author = {Kim, Geunwoo and Baldi, Pierre and McAleer, Stephen},
  title = {Language Models can Solve Computer Tasks},
  booktitle = {Advances in Neural Information Processing Systems},
  volume = {36},
  pages = {39648--39677},
  publisher = {Curran Associates, Inc.},
  year = {2023},
  doi = {10.52202/075280-1723},
  url = {https://proceedings.neurips.cc/paper_files/paper/2023/file/7cc1005ec73cfbaac9fa21192b622507-Paper-Conference.pdf}
}

@inproceedings{dhuliawala-etal-2024-chain,
  author = {Dhuliawala, Shehzaad  and
      Komeili, Mojtaba  and
      Xu, Jing  and
      Raileanu, Roberta  and
      Li, Xian  and
      Celikyilmaz, Asli  and
      Weston, Jason},
  title = {Chain-of-Verification Reduces Hallucination in Large Language Models},
  booktitle = {Findings of the {A}ssociation for {C}omputational {L}inguistics: {ACL} 2024},
  pages = {3563--3578},
  publisher = {Association for Computational Linguistics},
  year = {2024},
  doi = {10.18653/v1/2024.findings-acl.212},
  url = {https://aclanthology.org/2024.findings-acl.212/}
}

@article{jiang2023selfevolvecodeevolutionframework,
  author = {Shuyang Jiang and Yuhao Wang and Yu Wang},
  title = {{SelfEvolve}: A Code Evolution Framework via Large Language Models},
  journal = {arXiv preprint arXiv:2306.02907},
  year = {2023},
  url = {https://arxiv.org/abs/2306.02907},
  eprint = {2306.02907},
  archiveprefix = {arXiv},
  primaryclass = {cs.CL}
}

@inproceedings{SelfDebugging,
  author = {Xinyun Chen and Maxwell Lin and Nathanael Sch{\"{a}}rli and Denny Zhou},
  title = {Teaching Large Language Models to Self-Debug},
  booktitle = {International Conference on Learning Representations},
  year = {2024},
  doi = {10.48550/arXiv.2304.05128},
  url = {https://arxiv.org/abs/2304.05128}
}

@inproceedings{zhong-etal-2024-debug,
  author = {Zhong, Li  and
      Wang, Zilong  and
      Shang, Jingbo},
  title = {Debug like a Human: A Large Language Model Debugger via Verifying Runtime Execution Step by Step},
  booktitle = {Findings of the {A}ssociation for {C}omputational {L}inguistics: {ACL} 2024},
  pages = {851--870},
  publisher = {Association for Computational Linguistics},
  year = {2024},
  doi = {10.18653/v1/2024.findings-acl.49},
  url = {https://aclanthology.org/2024.findings-acl.49/}
}

@inproceedings{tyen-etal-2024-llms,
  author = {Tyen, Gladys  and
      Mansoor, Hassan  and
      Carbune, Victor  and
      Chen, Peter  and
      Mak, Tony},
  title = {{LLM}s cannot find reasoning errors, but can correct them given the error location},
  booktitle = {Findings of the {A}ssociation for {C}omputational {L}inguistics: {ACL} 2024},
  pages = {13894--13908},
  publisher = {Association for Computational Linguistics},
  year = {2024},
  doi = {10.18653/v1/2024.findings-acl.826},
  url = {https://aclanthology.org/2024.findings-acl.826/}
}

@inproceedings{zhuge2024gptswarm,
  author = {Mingchen Zhuge and Wenyi Wang and Louis Kirsch and Francesco Faccio and Dmitrii Khizbullin and J{\"u}rgen Schmidhuber},
  title = {{GPTSwarm}: Language Agents as Optimizable Graphs},
  booktitle = {Proceedings of the 41st International Conference on Machine Learning},
  series = {Proceedings of Machine Learning Research},
  volume = {235},
  pages = {62743--62767},
  publisher = {PMLR},
  year = {2024},
  url = {https://proceedings.mlr.press/v235/zhuge24a.html}
}

@article{li2024autoflow,
  author = {Zelong Li and Shuyuan Xu and Kai Mei and Wenyue Hua and Balaji Rama and Om Raheja and Hao Wang and He Zhu and Yongfeng Zhang},
  title = {{AutoFlow}: Automated Workflow Generation for Large Language Model Agents},
  journal = {arXiv preprint arXiv:2407.12821},
  year = {2024},
  doi = {10.48550/arXiv.2407.12821},
  url = {https://arxiv.org/abs/2407.12821},
  eprint = {2407.12821},
  archiveprefix = {arXiv}
}

@inproceedings{zhou2026mass,
  author = {Han Zhou and Xingchen Wan and Ruoxi Sun and Hamid Palangi and Shariq Iqbal and Ivan Vuli{\'c} and Anna Korhonen and Sercan {\"O}. Ar{\i}k},
  title = {Multi-Agent Design: Optimizing Agents with Better Prompts and Topologies},
  booktitle = {International Conference on Learning Representations},
  year = {2026},
  url = {https://arxiv.org/abs/2502.02533},
  eprint = {2502.02533},
  archiveprefix = {arXiv}
}

@article{zhang2024agentprune,
  author = {Guibin Zhang and Yanwei Yue and Zhixun Li and Sukwon Yun and Guancheng Wan and Kun Wang and Dawei Cheng and Jeffrey Xu Yu and Tianlong Chen},
  title = {Cut the Crap: An Economical Communication Pipeline for {LLM-based} Multi-Agent Systems},
  journal = {arXiv preprint arXiv:2410.02506},
  year = {2024},
  doi = {10.48550/arXiv.2410.02506},
  url = {https://arxiv.org/abs/2410.02506},
  eprint = {2410.02506},
  archiveprefix = {arXiv}
}

@article{zhao2026a2flow,
  author = {Mingming Zhao and Xiaokang Wei and Yuanqi Shao and Kaiwen Zhou and Lin Yang and Siwei Rao and Junhui Zhan and Zhitang Chen},
  title = {{A$^{2}$Flow}: Automating Agentic Workflow Generation via Self-Adaptive Abstraction Operators},
  journal = {Proceedings of the AAAI Conference on Artificial Intelligence},
  volume = {40},
  number = {35},
  pages = {29930--29938},
  year = {2026},
  doi = {10.1609/aaai.v40i35.40240},
  url = {https://ojs.aaai.org/index.php/AAAI/article/view/40240}
}

@article{zhao2024expel,
  author = {Andrew Zhao and Daniel Huang and Quentin Xu and Matthieu Lin and Yong-Jin Liu and Gao Huang},
  title = {{ExpeL}: {LLM} Agents Are Experiential Learners},
  journal = {Proceedings of the AAAI Conference on Artificial Intelligence},
  volume = {38},
  number = {17},
  pages = {19632--19642},
  year = {2024},
  doi = {10.1609/aaai.v38i17.29936},
  url = {https://ojs.aaai.org/index.php/AAAI/article/view/29936},
  eprint = {2308.10144},
  archiveprefix = {arXiv}
}

@inproceedings{ouyang2026reasoningbank,
  author = {Siru Ouyang and Jun Yan and I-Hung Hsu and Yanfei Chen and Ke Jiang and Zifeng Wang and Rujun Han and Long T. Le and Samira Daruki and Xiangru Tang and Vishy Tirumalashetty and George Lee and Mahsan Rofouei and Hangfei Lin and Jiawei Han and Chen-Yu Lee and Tomas Pfister},
  title = {{ReasoningBank}: Scaling Agent Self-Evolving with Reasoning Memory},
  booktitle = {International Conference on Learning Representations},
  year = {2026},
  url = {https://arxiv.org/abs/2509.25140},
  eprint = {2509.25140},
  archiveprefix = {arXiv}
}

@inproceedings{suzgun2026dynamiccheatsheet,
  author = {Mirac Suzgun and Mert Yuksekgonul and Federico Bianchi and Dan Jurafsky and James Zou},
  title = {Dynamic Cheatsheet: Test-Time Learning with Adaptive Memory},
  booktitle = {Proceedings of the 19th Conference of the European Chapter of the Association for Computational Linguistics (Volume 1: Long Papers)},
  pages = {7080--7106},
  publisher = {Association for Computational Linguistics},
  year = {2026},
  doi = {10.18653/v1/2026.eacl-long.333},
  url = {https://aclanthology.org/2026.eacl-long.333/}
}

@inproceedings{cao2026reme,
  author = {Zouying Cao and Jiaji Deng and Li Yu and Weikang Zhou and Zhaoyang Liu and Bolin Ding and Hai Zhao},
  title = {Remember Me, Refine Me: A Dynamic Procedural Memory Framework for Experience-Driven Agent Evolution},
  booktitle = {Findings of the Association for Computational Linguistics: ACL 2026},
  pages = {16803--16822},
  publisher = {Association for Computational Linguistics},
  year = {2026},
  doi = {10.18653/v1/2026.findings-acl.829},
  url = {https://aclanthology.org/2026.findings-acl.829/}
}

@inproceedings{mi2026darwinianmemory,
  author = {Hongze Mi and Yibo Feng and Wenjie Lu and Song Cao and Jinyuan Li and Yanming Li and Xuelin Zhang and Haotian Luo and Songyang Peng and He Cui and Tengfei Tian and Jun Fang and Hua Chai and Naiqiang Tan},
  title = {Darwinian Memory: A Training-Free Self-Regulating Memory System for {GUI} Agent Evolution},
  booktitle = {Proceedings of the 43rd International Conference on Machine Learning},
  series = {Proceedings of Machine Learning Research},
  volume = {306},
  pages = {88535--88553},
  publisher = {PMLR},
  year = {2026},
  url = {https://proceedings.mlr.press/v306/mi26b.html}
}

@inproceedings{zhang2026memevolve,
  author = {Guibin Zhang and Haotian Ren and Chong Zhan and Junhao Wang and He Zhu and Wangchunshu Zhou and Shuicheng Yan},
  title = {{MemEvolve}: Meta-Evolution of Agent Memory Systems},
  booktitle = {Proceedings of the 43rd International Conference on Machine Learning},
  series = {Proceedings of Machine Learning Research},
  volume = {306},
  pages = {158215--158234},
  publisher = {PMLR},
  year = {2026},
  url = {https://proceedings.mlr.press/v306/zhang26fa.html}
}

@article{liu2026evolvemem,
  author = {Jiaqi Liu and Xinyu Ye and Peng Xia and Zeyu Zheng and Cihang Xie and Mingyu Ding and Huaxiu Yao},
  title = {{EvolveMem}: Self-Evolving Memory Architecture via {AutoResearch} for {LLM} Agents},
  journal = {arXiv preprint arXiv:2605.13941},
  year = {2026},
  doi = {10.48550/arXiv.2605.13941},
  url = {https://arxiv.org/abs/2605.13941},
  eprint = {2605.13941},
  archiveprefix = {arXiv}
}

@inproceedings{wang2025asi,
  author = {Zora Zhiruo Wang and Apurva Gandhi and Graham Neubig and Daniel Fried},
  title = {Inducing Programmatic Skills for Agentic Tasks},
  booktitle = {Conference on Language Modeling},
  year = {2025},
  url = {https://openreview.net/forum?id=lsAY6fWsog},
  eprint = {2504.06821},
  archiveprefix = {arXiv}
}

@article{zheng2025skillweaver,
  author = {Boyuan Zheng and Michael Y. Fatemi and Xiaolong Jin and Zora Zhiruo Wang and Apurva Gandhi and Yueqi Song and Yu Gu and Jayanth Srinivasa and Gaowen Liu and Graham Neubig and Yu Su},
  title = {{SkillWeaver}: Web Agents can Self-Improve by Discovering and Honing Skills},
  journal = {arXiv preprint arXiv:2504.07079},
  year = {2025},
  doi = {10.48550/arXiv.2504.07079},
  url = {https://arxiv.org/abs/2504.07079},
  eprint = {2504.07079},
  archiveprefix = {arXiv}
}

@article{ni2026trace2skill,
  author = {Jingwei Ni and Yihao Liu and Xinpeng Liu and Yutao Sun and Mengyu Zhou and Pengyu Cheng and Dexin Wang and Erchao Zhao and Xiaoxi Jiang and Guanjun Jiang},
  title = {{Trace2Skill}: Distill Trajectory-Local Lessons into Transferable Agent Skills},
  journal = {arXiv preprint arXiv:2603.25158},
  year = {2026},
  doi = {10.48550/arXiv.2603.25158},
  url = {https://arxiv.org/abs/2603.25158},
  eprint = {2603.25158},
  archiveprefix = {arXiv}
}

@article{wang2026skillx,
  author = {Chenxi Wang and Zhuoyun Yu and Xin Xie and Wuguannan Yao and Runnan Fang and Shuofei Qiao and Kexin Cao and Guozhou Zheng and Xiang Qi and Peng Zhang and Shumin Deng},
  title = {{SkillX}: Automatically Constructing Skill Knowledge Bases for Agents},
  journal = {arXiv preprint arXiv:2604.04804},
  year = {2026},
  doi = {10.48550/arXiv.2604.04804},
  url = {https://arxiv.org/abs/2604.04804},
  eprint = {2604.04804},
  archiveprefix = {arXiv}
}

@article{shi2026psn,
  author = {Haochen Shi and Xingdi Yuan and Bang Liu},
  title = {Evolving Programmatic Skill Networks},
  journal = {arXiv preprint arXiv:2601.03509},
  year = {2026},
  doi = {10.48550/arXiv.2601.03509},
  url = {https://arxiv.org/abs/2601.03509},
  eprint = {2601.03509},
  archiveprefix = {arXiv}
}

@article{lu2026contractskill,
  author = {Zijian Lu and Yiping Zuo and Yupeng Nie and Xin He and Weibei Fan and Lianyong Qi and Shi Jin},
  title = {{ContractSkill}: Repairable Contract-Based Skills for Multimodal Web Agents},
  journal = {arXiv preprint arXiv:2603.20340},
  year = {2026},
  doi = {10.48550/arXiv.2603.20340},
  url = {https://arxiv.org/abs/2603.20340},
  eprint = {2603.20340},
  archiveprefix = {arXiv}
}

@article{vishe2026skillr1,
  author = {Yash Vishe and Rohan Surana and Xunyi Jiang and Zihan Huang and Xintong Li and Nikki Lijing Kuang and Tong Yu and Ryan A. Rossi and Jingbo Shang and Julian McAuley and Junda Wu},
  title = {{Skill-R1}: Agent Skill Evolution via Reinforcement Learning},
  journal = {arXiv preprint arXiv:2605.09359},
  year = {2026},
  doi = {10.48550/arXiv.2605.09359},
  url = {https://arxiv.org/abs/2605.09359},
  eprint = {2605.09359},
  archiveprefix = {arXiv}
}

@inproceedings{wang2024trove,
  author = {Zhiruo Wang and Graham Neubig and Daniel Fried},
  title = {{TroVE}: Inducing Verifiable and Efficient Toolboxes for Solving Programmatic Tasks},
  booktitle = {Proceedings of the 41st International Conference on Machine Learning},
  series = {Proceedings of Machine Learning Research},
  volume = {235},
  pages = {51177--51191},
  publisher = {PMLR},
  year = {2024},
  url = {https://proceedings.mlr.press/v235/wang24az.html}
}

@inproceedings{stengeleskin2024regal,
  author = {Elias Stengel-Eskin and Archiki Prasad and Mohit Bansal},
  title = {{ReGAL}: Refactoring Programs to Discover Generalizable Abstractions},
  booktitle = {Proceedings of the 41st International Conference on Machine Learning},
  series = {Proceedings of Machine Learning Research},
  volume = {235},
  pages = {46605--46624},
  publisher = {PMLR},
  year = {2024},
  url = {https://proceedings.mlr.press/v235/stengel-eskin24a.html}
}

@article{yue2025toollibgen,
  author = {Murong Yue and Zhiwei Liu and Liangwei Yang and Jianguo Zhang and Zuxin Liu and Haolin Chen and Ziyu Yao and Silvio Savarese and Caiming Xiong and Shelby Heinecke and Huan Wang},
  title = {{ToolLibGen}: Scalable Automatic Tool Creation and Aggregation for {LLM} Reasoning},
  journal = {arXiv preprint arXiv:2510.07768},
  year = {2025},
  doi = {10.48550/arXiv.2510.07768},
  url = {https://arxiv.org/abs/2510.07768},
  eprint = {2510.07768},
  archiveprefix = {arXiv}
}

@inproceedings{luo2026gate,
  author = {Jianwen Luo and Yiming Huang and Jinxiang Meng and Fangyu Lei and Shizhu He and Xiao Liu and Shanshan Jiang and Bin Dong and Jun Zhao and Kang Liu},
  title = {{GATE}: Graph-based Adaptive Tool Evolution Across Diverse Tasks},
  booktitle = {Proceedings of the 64th Annual Meeting of the Association for Computational Linguistics (Volume 1: Long Papers)},
  pages = {1934--1961},
  publisher = {Association for Computational Linguistics},
  year = {2026},
  doi = {10.18653/v1/2026.acl-long.87},
  url = {https://aclanthology.org/2026.acl-long.87/}
}

@inproceedings{wu2026contda,
  author = {Bin Wu and Edgar Meij and Emine Yilmaz},
  title = {Beyond Static Toolsets: Self-Evolving {LLM} Tool Agents via Continual Documentation Adaptation},
  booktitle = {Findings of the Association for Computational Linguistics: ACL 2026},
  pages = {21519--21539},
  publisher = {Association for Computational Linguistics},
  year = {2026},
  doi = {10.18653/v1/2026.findings-acl.1082},
  url = {https://aclanthology.org/2026.findings-acl.1082/}
}

@inproceedings{shang2025agentsquare,
  author = {Yu Shang and Yu Li and Keyu Zhao and Likai Ma and Jiahe Liu and Fengli Xu and Yong Li},
  title = {{AgentSquare}: Automatic {LLM} Agent Search in Modular Design Space},
  booktitle = {International Conference on Learning Representations},
  year = {2025},
  url = {https://arxiv.org/abs/2410.06153},
  eprint = {2410.06153},
  archiveprefix = {arXiv}
}

@article{zhang2026selfharness,
  author = {Hangfan Zhang and Shao Zhang and Kangcong Li and Chen Zhang and Yang Chen and Yiqun Zhang and Lei Bai and Shuyue Hu},
  title = {{Self-Harness}: Harnesses That Improve Themselves},
  journal = {arXiv preprint arXiv:2606.09498},
  year = {2026},
  doi = {10.48550/arXiv.2606.09498},
  url = {https://arxiv.org/abs/2606.09498},
  eprint = {2606.09498},
  archiveprefix = {arXiv}
}

@article{liu2026adaptiveautoharness,
  author = {Zewen Liu and Zhan Shi and Yisi Sang and Bing He and Minhua Lin and Tianxin Wei and Dakuo Wang and Benoit Dumoulin and Wei Jin and Hanqing Lu},
  title = {Adaptive Auto-Harness: Sustained Self-Improvement for Agentic System Deployment on Open-Ended Task Streams},
  journal = {arXiv preprint arXiv:2606.01770},
  year = {2026},
  doi = {10.48550/arXiv.2606.01770},
  url = {https://arxiv.org/abs/2606.01770},
  eprint = {2606.01770},
  archiveprefix = {arXiv}
}

@article{zhang2026jitagent,
  author = {Guibin Zhang and Leo Lu and Fangzhou Xie and Kang Zhu and Junhao Wang and Zhifei Xie and Zhaochen Yu and Zihang Liu and Zhongxiang Sun and Qiankun Li and Yue Liao and Heng Chang and Xiaobin Hu and Qibing Ren and Wangchunshu Zhou and Chuanrui Hu and Yafeng Deng and Shuicheng Yan},
  title = {{JIT-Agent}: Scaling Harness Intelligence via Just-in-Time Harness Evolution},
  journal = {arXiv preprint arXiv:2608.25593},
  year = {2026},
  doi = {10.48550/arXiv.2608.25593},
  url = {https://arxiv.org/abs/2608.25593},
  eprint = {2608.25593},
  archiveprefix = {arXiv}
}

@article{lin2026harnessbenefit,
  author = {Minhua Lin and Juncheng Wu and Zijun Wang and Zhan Shi and Yisi Sang and Bing He and Zewen Liu and Tianxin Wei and Zongyu Wu and Zhiwei Zhang and Dakuo Wang and Xiang Zhang and Benoit Dumoulin and Cihang Xie and Yuyin Zhou and Suhang Wang and Hanqing Lu},
  title = {Harness Updating Is Not Harness Benefit: Disentangling Evolution Capabilities in Self-Evolving {LLM} Agents},
  journal = {arXiv preprint arXiv:2605.30621},
  year = {2026},
  doi = {10.48550/arXiv.2605.30621},
  url = {https://arxiv.org/abs/2605.30621},
  eprint = {2605.30621},
  archiveprefix = {arXiv}
}

@article{wu2026modularrsi,
  author = {Siwei Wu and Jincheng Ren and Yizhi Li and Haau-Sing Li and Chengran Yang and Xudong Ma and Yuxuan Zhang and Weicheng Gu and Jian Yang and Riza Batista-Navarro and Chuanyi Zhang and Xianglong Liu and Ming Zhou and Bryan Dai and Chenghua Lin},
  title = {{ModularRSI}: Modular and Generalizable Recursive Harness Self-Improvement},
  journal = {arXiv preprint arXiv:2609.14857},
  year = {2026},
  doi = {10.48550/arXiv.2609.14857},
  url = {https://arxiv.org/abs/2609.14857},
  eprint = {2609.14857},
  archiveprefix = {arXiv},
  primaryclass = {cs.CL},
  note = {Revised 9 October 2026; first posted 14 September 2026},
  month = oct
}

@inproceedings{STaR,
  author = {Eric Zelikman and Yuhuai Wu and Jesse Mu and Noah Goodman},
  title = {{STaR}: Bootstrapping Reasoning With Reasoning},
  booktitle = {Advances in Neural Information Processing Systems},
  volume = {35},
  year = {2022},
  doi = {10.52202/068431-1126},
  url = {https://proceedings.neurips.cc/paper_files/paper/2022/hash/639a9a172c044fbb64175b5fad42e9a5-Abstract-Conference.html},
  eprint = {2203.14465},
  archiveprefix = {arXiv}
}

@article{DeepSeekR1,
  author = {Daya Guo and Dejian Yang and Haowei Zhang and others},
  title = {{DeepSeek-R1} incentivizes reasoning in {LLMs} through reinforcement learning},
  journal = {Nature},
  volume = {645},
  number = {8081},
  pages = {633--638},
  year = {2025},
  doi = {10.1038/s41586-025-09422-z},
  url = {https://www.nature.com/articles/s41586-025-09422-z}
}

@inproceedings{AgentTuning,
  author = {Zeng, Aohan and Liu, Mingdao and Lu, Rui and Wang, Bowen and Liu, Xiao and Dong, Yuxiao and Tang, Jie},
  title = {{A}gent{T}uning: Enabling Generalized Agent Abilities for {LLM}s},
  booktitle = {Findings of the {A}ssociation for {C}omputational {L}inguistics: {ACL} 2024},
  pages = {3053--3077},
  publisher = {Association for Computational Linguistics},
  year = {2024},
  doi = {10.18653/v1/2024.findings-acl.181},
  url = {https://aclanthology.org/2024.findings-acl.181/}
}

@article{gulcehre2023rest,
  author = {Caglar Gulcehre and Tom Le Paine and Srivatsan Srinivasan and Ksenia Konyushkova and Lotte Weerts and Abhishek Sharma and Aditya Siddhant and Alex Ahern and Miaosen Wang and Chenjie Gu and Wolfgang Macherey and Arnaud Doucet and Orhan Firat and Nando de Freitas},
  title = {Reinforced Self-Training ({ReST}) for Language Modeling},
  journal = {arXiv preprint arXiv:2308.08998},
  year = {2023},
  doi = {10.48550/arXiv.2308.08998},
  url = {https://arxiv.org/abs/2308.08998},
  eprint = {2308.08998},
  archiveprefix = {arXiv}
}

@inproceedings{chen2024spin,
  author = {Chen, Zixiang and Deng, Yihe and Yuan, Huizhuo and Ji, Kaixuan and Gu, Quanquan},
  title = {Self-Play Fine-Tuning Converts Weak Language Models to Strong Language Models},
  booktitle = {Proceedings of the 41st International Conference on Machine Learning},
  series = {Proceedings of Machine Learning Research},
  volume = {235},
  pages = {6621--6642},
  publisher = {PMLR},
  year = {2024},
  url = {https://proceedings.mlr.press/v235/chen24j.html}
}

@inproceedings{zelikman2024quietstar,
  author = {Eric Zelikman and Georges Harik and Yijia Shao and Varuna Jayasiri and Nick Haber and Noah D. Goodman},
  title = {{Quiet-STaR}: Language Models Can Teach Themselves to Think Before Speaking},
  booktitle = {Conference on Language Modeling},
  year = {2024},
  url = {https://openreview.net/forum?id=oRXPiSOGH9},
  eprint = {2403.09629},
  archiveprefix = {arXiv}
}

@inproceedings{hosseini2024vstar,
  author = {Arian Hosseini and Xingdi Yuan and Nikolay Malkin and Aaron Courville and Alessandro Sordoni and Rishabh Agarwal},
  title = {{V-STaR}: Training Verifiers for Self-Taught Reasoners},
  booktitle = {Conference on Language Modeling},
  year = {2024},
  url = {https://openreview.net/forum?id=stmqBSW2dV},
  eprint = {2402.06457},
  archiveprefix = {arXiv}
}

@inproceedings{zhang2024restmcts,
  author = {Dan Zhang and Sining Zhoubian and Ziniu Hu and Yisong Yue and Yuxiao Dong and Jie Tang},
  title = {{ReST-MCTS*}: {LLM} Self-Training via Process Reward Guided Tree Search},
  booktitle = {Advances in Neural Information Processing Systems},
  volume = {37},
  year = {2024},
  doi = {10.52202/079017-2066},
  url = {https://proceedings.neurips.cc/paper_files/paper/2024/hash/76ec4dc30e9faaf0e4b6093eaa377218-Abstract-Conference.html},
  eprint = {2406.03816},
  archiveprefix = {arXiv}
}

@inproceedings{guan2025rstarmath,
  author = {Guan, Xinyu and Zhang, Li Lyna and Liu, Yifei and Shang, Ning and Sun, Youran and Zhu, Yi and Yang, Fan and Yang, Mao},
  title = {{rStar-Math}: Small {LLMs} Can Master Math Reasoning with Self-Evolved Deep Thinking},
  booktitle = {Proceedings of the 42nd International Conference on Machine Learning},
  series = {Proceedings of Machine Learning Research},
  volume = {267},
  pages = {20640--20661},
  publisher = {PMLR},
  year = {2025},
  url = {https://proceedings.mlr.press/v267/guan25f.html}
}

@article{shao2024deepseekmath,
  author = {Zhihong Shao and Peiyi Wang and Qihao Zhu and Runxin Xu and Junxiao Song and Xiao Bi and Haowei Zhang and Mingchuan Zhang and Y. K. Li and Y. Wu and Daya Guo},
  title = {{DeepSeekMath}: Pushing the Limits of Mathematical Reasoning in Open Language Models},
  journal = {arXiv preprint arXiv:2402.03300},
  year = {2024},
  doi = {10.48550/arXiv.2402.03300},
  url = {https://arxiv.org/abs/2402.03300v3},
  eprint = {2402.03300},
  archiveprefix = {arXiv}
}

@article{hu2025openreasonerzero,
  author = {Jingcheng Hu and Yinmin Zhang and Qi Han and Daxin Jiang and Xiangyu Zhang and Heung-Yeung Shum},
  title = {{Open-Reasoner-Zero}: An Open Source Approach to Scaling Up Reinforcement Learning on the Base Model},
  journal = {arXiv preprint arXiv:2503.24290},
  year = {2025},
  doi = {10.48550/arXiv.2503.24290},
  url = {https://arxiv.org/abs/2503.24290},
  eprint = {2503.24290},
  archiveprefix = {arXiv}
}

@article{chen2023fireact,
  author = {Baian Chen and Chang Shu and Ehsan Shareghi and Nigel Collier and Karthik Narasimhan and Shunyu Yao},
  title = {{FireAct}: Toward Language Agent Fine-tuning},
  journal = {arXiv preprint arXiv:2310.05915},
  year = {2023},
  doi = {10.48550/arXiv.2310.05915},
  url = {https://arxiv.org/abs/2310.05915},
  eprint = {2310.05915},
  archiveprefix = {arXiv}
}

@inproceedings{yang2026toolcpt,
  author = {Yang, Yifan and Lu, Jinghui and Deng, Yaping and Yang, Ao and Yu, Peijie and Yu, Tinghao and Zhang, Feng},
  title = {{T}ool{CPT}: Improving Tool Utilization in {LLM} Agents via Continuous Pre-training},
  booktitle = {Findings of the {A}ssociation for {C}omputational {L}inguistics: {ACL} 2026},
  pages = {15830--15856},
  publisher = {Association for Computational Linguistics},
  year = {2026},
  doi = {10.18653/v1/2026.findings-acl.776},
  url = {https://aclanthology.org/2026.findings-acl.776/}
}

@inproceedings{yan2026memoryr1,
  author = {Yan, Sikuan and Yang, Xiufeng and Huang, Zuchao and Nie, Ercong and Ding, Zifeng and Li, Zonggen and Ma, Xiaowen and Bi, Jinhe and Kersting, Kristian and Pan, Jeff Z. and Schuetze, Hinrich and Tresp, Volker and Ma, Yunpu},
  title = {{M}emory-{R}1: Enhancing Large Language Model Agents to Manage and Utilize Memories via Reinforcement Learning},
  booktitle = {Proceedings of the 64th Annual Meeting of the {A}ssociation for {C}omputational {L}inguistics (Volume 1: Long Papers)},
  pages = {12805--12825},
  publisher = {Association for Computational Linguistics},
  year = {2026},
  doi = {10.18653/v1/2026.acl-long.583},
  url = {https://aclanthology.org/2026.acl-long.583/}
}

@article{li2026skillgraph,
  author = {Xiaoyuan Li and Moxin Li and Keqin Bao and Yubo Ma and Wenjie Wang and Dayiheng Liu and Fuli Feng},
  title = {{SkillGraph}: Skill-Augmented Reinforcement Learning for Agents via Evolving Skill Graphs},
  journal = {arXiv preprint arXiv:2605.12039},
  year = {2026},
  doi = {10.48550/arXiv.2605.12039},
  url = {https://arxiv.org/abs/2605.12039},
  eprint = {2605.12039},
  archiveprefix = {arXiv}
}

@article{ge2026skillgym,
  author = {Zhilong Ge and Yuting Shao and Yutao Yang and Yuxuan Cai and Jie Zhou and Kai Chen and Bo Zhang and Qin Chen and Liang He},
  title = {{SkillGym}: Internalizing Human Skills into {LLMs} for Real-World Problem Solving},
  journal = {arXiv preprint arXiv:2609.27717},
  year = {2026},
  doi = {10.48550/arXiv.2609.27717},
  url = {https://arxiv.org/abs/2609.27717},
  eprint = {2609.27717},
  archiveprefix = {arXiv}
}

@article{feng2025retool,
  author = {Jiazhan Feng and Shijue Huang and Xingwei Qu and Ge Zhang and Yujia Qin and Baoquan Zhong and Chengquan Jiang and Jinxin Chi and Wanjun Zhong},
  title = {{ReTool}: Reinforcement Learning for Strategic Tool Use in {LLMs}},
  journal = {arXiv preprint arXiv:2504.11536},
  year = {2025},
  doi = {10.48550/arXiv.2504.11536},
  url = {https://arxiv.org/abs/2504.11536},
  eprint = {2504.11536},
  archiveprefix = {arXiv}
}

@inproceedings{zou2026autotool,
  author = {Zou, Jiaru and Yang, Ling and Qi, Yunzhe and Chen, Sirui and Ai, Mengting and Shen, Ke and He, Jingrui and Wang, Mengdi},
  title = {{AutoTool}: Dynamic Tool Selection and Integration for Agentic Reasoning},
  booktitle = {Proceedings of the 43rd International Conference on Machine Learning},
  series = {Proceedings of Machine Learning Research},
  volume = {306},
  pages = {168955--168978},
  publisher = {PMLR},
  year = {2026},
  url = {https://proceedings.mlr.press/v306/zou26p.html}
}

@inproceedings{su2026toolorchestra,
  author = {Su, Hongjin and Diao, Shizhe and Lu, Ximing and Liu, Mingjie and Xu, Jiacheng and Dong, Xin and Fu, Yonggan and Belcak, Peter and Ye, Hanrong and Yin, Hongxu and Dong, Yi and Bakhturina, Evelina and Yu, Tao and Choi, Yejin and Kautz, Jan and Molchanov, Pavlo},
  title = {{ToolOrchestra}: Elevating Intelligence via Efficient Model and Tool Orchestration},
  booktitle = {Proceedings of the 43rd International Conference on Machine Learning},
  series = {Proceedings of Machine Learning Research},
  volume = {306},
  pages = {115934--115956},
  publisher = {PMLR},
  year = {2026},
  url = {https://proceedings.mlr.press/v306/su26d.html}
}

@article{luo2025agentlightning,
  author = {Xufang Luo and Yuge Zhang and Zhiyuan He and Zilong Wang and Siyun Zhao and Dongsheng Li and Luna K. Qiu and Yuqing Yang},
  title = {Agent Lightning: Train {ANY} {AI} Agents with Reinforcement Learning},
  journal = {arXiv preprint arXiv:2508.03680},
  year = {2025},
  doi = {10.48550/arXiv.2508.03680},
  url = {https://arxiv.org/abs/2508.03680},
  eprint = {2508.03680},
  archiveprefix = {arXiv}
}

@article{zhang2026aevo,
  author = {Jiayi Zhang and Yongfeng Gu and Jianhao Ruan and Maojia Song and Yiran Peng and Zhiguang Han and Jinyu Xiang and Zhitao Wang and Caiyin Yang and Yixi Ouyang and Bang Liu and Chenglin Wu and Yuyu Luo},
  title = {Harnessing Agentic Evolution},
  journal = {arXiv preprint arXiv:2605.13821},
  year = {2026},
  doi = {10.48550/arXiv.2605.13821},
  url = {https://arxiv.org/abs/2605.13821},
  eprint = {2605.13821},
  archiveprefix = {arXiv}
}

@article{qu2026bilevel,
  author = {Qu, Yaonan and Lu, Meng},
  title = {Bilevel Autoresearch: Meta-Autoresearching Itself},
  journal = {arXiv preprint arXiv:2603.23420},
  year = {2026},
  doi = {10.48550/arXiv.2603.23420},
  url = {https://arxiv.org/abs/2603.23420},
  eprint = {2603.23420},
  archiveprefix = {arXiv},
  primaryclass = {cs.AI}
}

@article{zheng2026dreamrsi,
  author = {Zheng, Tong and Wu, Xidong and Zhang, Zheng and He, Zhankui and Zhang, Chaoyi and Coleman, Benjamin and Wei, Ruoqiao and Bai, Di and Liu, Haolin and Liu, Rui and Wang, Xue and Zhuan, Yue and Kang, Wang-Cheng and Xiang, Renkai and Huang, Heng and Cheng, Xinwu and Guo, Yunsong},
  title = {{Dream-RSI}: Recursive Self-Improvement through Evolving Worlds},
  journal = {arXiv preprint arXiv:2609.14858},
  year = {2026},
  doi = {10.48550/arXiv.2609.14858},
  url = {https://arxiv.org/abs/2609.14858},
  eprint = {2609.14858},
  archiveprefix = {arXiv},
  primaryclass = {cs.CL}
}

@article{dong2026mixture,
  author = {Dong, Haoyu and Zhou, Yuhang and Lin, Zihao and Wu, Yifan and Peng, Bo and Wang, Mingyi and Fan, Xiangjun and Zhang, Lizhu and Zhao, Zhuokai},
  title = {Mixture of Self-Improving Branches For Agent Harness Optimization},
  journal = {arXiv preprint arXiv:2609.37834},
  year = {2026},
  doi = {10.48550/arXiv.2609.37834},
  url = {https://arxiv.org/abs/2609.37834},
  eprint = {2609.37834},
  archiveprefix = {arXiv},
  primaryclass = {cs.AI}
}

@article{yang2026selfsearch,
  author = {Yang, Jungwoo and Kong, Injin and Jo, Yohan},
  title = {{SelfSearch}: Reward-Free Search for Self-Improving Agents},
  journal = {arXiv preprint arXiv:2609.37968},
  year = {2026},
  doi = {10.48550/arXiv.2609.37968},
  url = {https://arxiv.org/abs/2609.37968},
  eprint = {2609.37968},
  archiveprefix = {arXiv},
  primaryclass = {cs.AI}
}

@article{xu2026specrsi,
  author = {Xu, Xin and Cheng, Ke and Wang, Hao and Huang, Yangru and Xiao, Yanghao and Chen, Zhichao and Zhong, Zhisheng and Liu, Xun and Li, Yi and Li, Qian and Lv, Haibin and Zhang, Jun and Jiang, Jie and Lin, Zhouchen},
  title = {Drafting Agents, Not Just Tokens: Speculative Recursive Self-Improvement for Agentic Systems},
  journal = {arXiv preprint arXiv:2610.16491},
  year = {2026},
  doi = {10.48550/arXiv.2610.16491},
  url = {https://arxiv.org/abs/2610.16491},
  eprint = {2610.16491},
  archiveprefix = {arXiv},
  primaryclass = {cs.LG}
}

@article{srikanth2026aide2,
  author = {Dhruv Srikanth and Bingchen Zhao and Dixing Xu and Yuxiang Wu and Zhengyao Jiang},
  title = {Recursive self-improvement of {AI} research agents},
  journal = {arXiv preprint arXiv:2609.26457},
  year = {2026},
  doi = {10.48550/arXiv.2609.26457},
  url = {https://arxiv.org/abs/2609.26457},
  eprint = {2609.26457},
  archiveprefix = {arXiv}
}

@article{wu2025sgm,
  author = {Xuening Wu and Shenqin Yin and Yanlan Kang and Xinhang Zhang and Qianya Xu and Zeping Chen and Wenqiang Zhang},
  title = {{SGM}: A Statistical {Godel} Machine for Risk-Controlled Recursive Self-Modification},
  journal = {arXiv preprint arXiv:2510.10232},
  year = {2025},
  doi = {10.48550/arXiv.2510.10232},
  url = {https://arxiv.org/abs/2510.10232},
  eprint = {2510.10232},
  archiveprefix = {arXiv}
}

@article{sage2026,
  author = {Yihao Wang and Linhan Xia and Rui Liu and Zhaofeng Zhang and Hongyu Wu and Yang Yang and Jinglu He and Yu Guo and Kai Lei},
  title = {{SAGE}: A Statistical Acceptance Gate for Self-Evolving Agents},
  journal = {arXiv preprint arXiv:2609.36043},
  year = {2026},
  doi = {10.48550/arXiv.2609.36043},
  url = {https://arxiv.org/abs/2609.36043},
  eprint = {2609.36043},
  archiveprefix = {arXiv},
  primaryclass = {cs.AI},
  note = {First posted 28 September 2026},
  month = sep
}

@article{sea2026,
  author = {Biswa Sengupta},
  title = {Self-Evolving Agents with Anytime-Valid Certificates},
  journal = {arXiv preprint arXiv:2607.00871},
  year = {2026},
  doi = {10.48550/arXiv.2607.00871},
  url = {https://arxiv.org/abs/2607.00871},
  eprint = {2607.00871},
  archiveprefix = {arXiv},
  primaryclass = {cs.AI},
  note = {First posted 1 July 2026},
  month = jul
}

@article{cai2026reliable,
  author = {Qianshu Cai and Yonggang Zhang and Jun Nie and Maohao Ran and Huajiang Zheng and Jun Song and Xinmei Tian and Yike Guo and Wei Xue},
  title = {A Theory of Reliable Self-Evolution for Agent Harnesses},
  journal = {arXiv preprint arXiv:2609.08175},
  year = {2026},
  doi = {10.48550/arXiv.2609.08175},
  url = {https://arxiv.org/abs/2609.08175},
  eprint = {2609.08175},
  archiveprefix = {arXiv}
}

@article{zou2026fmlbench,
  author = {Qiran Zou and Hou Hei Lam and Wenhao Zhao and Tingting Chen and Yiming Tang and Samson Yu and Yingtao Zhu and Srinivas Anumasa and Zufeng Zhang and Tianyi Zhang and Chang Liu and Zhengyao Jiang and Anirudh Goyal and Dianbo Liu},
  title = {{FML-bench}: A Controlled Study of {AI} Research Agent Strategies from the Perspective of Search Dynamics},
  journal = {arXiv preprint arXiv:2605.17373},
  year = {2026},
  doi = {10.48550/arXiv.2605.17373},
  url = {https://arxiv.org/abs/2605.17373},
  eprint = {2605.17373},
  archiveprefix = {arXiv}
}

@article{evopathbench2026,
  author = {Hongqiang Lin and Chao Liu and Xiaofan Bai and Xuan Jin and Yuhong Li and Nenggan Zheng and Xipeng Cao},
  title = {Beyond Endpoint Performance: Process-Level Evaluation of Self-Evolving Agents},
  journal = {arXiv preprint arXiv:2609.24663},
  year = {2026},
  doi = {10.48550/arXiv.2609.24663},
  url = {https://arxiv.org/abs/2609.24663},
  eprint = {2609.24663},
  archiveprefix = {arXiv},
  primaryclass = {cs.AI}
}

@article{xia2026rrsi,
  author = {Peng Xia and Rujun Han and Zifeng Wang and Yanfei Chen and Yufan Zhuang and Yoonho Lee and Chengsong Huang and Han Yu and Zhongying CuiZhu and Yifei Ming and Huaxiu Yao and Burak Gokturk and Tomas Pfister and Chen-Yu Lee},
  title = {{RRSI}: Regularized Recursive Self-Improvement of Agent Harnesses},
  journal = {arXiv preprint arXiv:2609.24972},
  year = {2026},
  doi = {10.48550/arXiv.2609.24972},
  url = {https://arxiv.org/abs/2609.24972},
  eprint = {2609.24972},
  archiveprefix = {arXiv}
}

@inproceedings{walle2,
  author = {Siyu Zhou and Tianyi Zhou and Yijun Yang and Guodong Long and Deheng Ye and Jing Jiang and Chengqi Zhang},
  title = {{WALL-E}: World Alignment by {NeuroSymbolic} Learning improves World Model-based {LLM} Agents},
  booktitle = {Advances in Neural Information Processing Systems},
  volume = {38},
  year = {2025},
  url = {https://proceedings.neurips.cc/paper_files/paper/2025/hash/5e772a13ccba5255331240dcd99aa38b-Abstract-Conference.html},
  eprint = {2504.15785},
  archiveprefix = {arXiv}
}

@article{proplay,
  author = {Yijun Ma and Zehong Wang and Yiyang Li and Ziming Li and Xiaoguang Guo and Weixiang Sun and Chuxu Zhang and Yanfang Ye},
  title = {{ProPlay}: Procedural World Models for Self-Evolving {LLM} Agents},
  journal = {arXiv preprint arXiv:2606.12780},
  year = {2026},
  url = {https://arxiv.org/abs/2606.12780},
  eprint = {2606.12780},
  archiveprefix = {arXiv}
}

@article{opineworld,
  author = {David Courtis and Wenhao Li and Scott Sanner},
  title = {{OPINE-World}: Programmatic World Modeling with Ontology-error-Prioritized Interactive Exploration for {ARC-AGI-3}},
  journal = {arXiv preprint arXiv:2607.01531},
  year = {2026},
  url = {https://arxiv.org/abs/2607.01531},
  eprint = {2607.01531},
  archiveprefix = {arXiv}
}

@article{evobelief,
  author = {Janet Wang and Jiang Guo and Yi Xiang and Jiaqing Yuan and Mingwen Dong and Hideo Kobayashi and Shuaichen Chang and Chandan K. Reddy and Jiarong Jiang},
  title = {{EvoBelief}: Falsifiable World Models for Self-Improving {LLM} Agents},
  journal = {arXiv preprint arXiv:2610.12789},
  year = {2026},
  url = {https://arxiv.org/abs/2610.12789},
  eprint = {2610.12789},
  archiveprefix = {arXiv}
}

@article{kepler,
  author = {Wensen Wu},
  title = {{Kepler}: Auditable World Models for {ARC-AGI-3}},
  journal = {arXiv preprint arXiv:2610.00834},
  year = {2026},
  url = {https://arxiv.org/abs/2610.00834},
  eprint = {2610.00834},
  archiveprefix = {arXiv}
}

@inproceedings{worldcoder,
  author = {Hao Tang and Darren Key and Kevin Ellis},
  title = {{WorldCoder}, a Model-Based {LLM} Agent: Building World Models by Writing Code and Interacting with the Environment},
  booktitle = {Advances in Neural Information Processing Systems},
  volume = {37},
  pages = {70148--70212},
  publisher = {Curran Associates, Inc.},
  year = {2024},
  doi = {10.52202/079017-2243},
  url = {https://proceedings.neurips.cc/paper_files/paper/2024/hash/820c61a0cd419163ccbd2c33b268816e-Abstract-Conference.html},
  eprint = {2402.12275},
  archiveprefix = {arXiv}
}

@inproceedings{gifmcts,
  author = {Nicola Dainese and Matteo Merler and Minttu Alakuijala and Pekka Marttinen},
  title = {Generating Code World Models with Large Language Models Guided by {Monte} {Carlo} Tree Search},
  booktitle = {Advances in Neural Information Processing Systems},
  volume = {37},
  year = {2024},
  url = {https://proceedings.neurips.cc/paper_files/paper/2024/hash/6f479ea488e0908ac8b1b37b27fd134c-Abstract-Conference.html},
  eprint = {2405.15383},
  archiveprefix = {arXiv}
}

@inproceedings{poeworld,
  author = {Top Piriyakulkij and Yichao Liang and Hao Tang and Adrian Weller and Marta Kryven and Kevin Ellis},
  title = {{PoE-World}: Compositional World Modeling with Products of Programmatic Experts},
  booktitle = {Advances in Neural Information Processing Systems},
  volume = {38},
  year = {2025},
  url = {https://proceedings.neurips.cc/paper_files/paper/2025/hash/262dd62fd1bbb30d6a6b4d578f5e65ff-Abstract-Conference.html},
  eprint = {2505.10819},
  archiveprefix = {arXiv}
}

@article{codeasworlds,
  author = {Hanyang Wang and Yimo Cai and Weiliang Chen and Jiawei Chi and Haowen Sun and Qiyu Dai and Yi-Hsin Hung and Xingzhuo Guo and Jinshan Ren and Runmao Yao and Ziwei Liu and Mingsheng Long and Yueqi Duan and Jun Gao and Jiangran Lyu and Fangfu Liu and Jialong Wu},
  title = {Code as Worlds: Agentic Discovery of Executable World Representations for Physical Reasoning},
  journal = {arXiv preprint arXiv:2608.27549},
  year = {2026},
  url = {https://arxiv.org/abs/2608.27549},
  eprint = {2608.27549},
  archiveprefix = {arXiv}
}

@article{twin,
  author = {Alexy Skoutnev and Kirill Acharya and Gaston Longhitano and Madeleine Udell and Kevin Ellis and Iddo Drori},
  title = {{Twin}: Playing an Unknown Game with a Test-Time Digital Twin},
  journal = {arXiv preprint arXiv:2608.14490},
  year = {2026},
  url = {https://arxiv.org/abs/2608.14490},
  eprint = {2608.14490},
  archiveprefix = {arXiv}
}

@article{memento3,
  author = {Haoyu Zhao and Zhengxu Yu and Zhiyuan He and Meng Fang and Rasul Tutunov and Haitham Bou-Ammar and Weilin Luo and Jun Wang},
  title = {Memento 3: Model-Based Recursive Self-Improvement through Reflective Rulebooks},
  journal = {arXiv preprint arXiv:2610.11794},
  year = {2026},
  url = {https://arxiv.org/abs/2610.11794},
  eprint = {2610.11794},
  archiveprefix = {arXiv}
}

@article{dworld,
  author = {Xuan Wen and Baichao Wang},
  title = {{D-World}: An Explicit World-Language-Action Model with Recursive Self-Improvement under Dual Control},
  journal = {arXiv preprint arXiv:2610.17114},
  year = {2026},
  url = {https://arxiv.org/abs/2610.17114},
  eprint = {2610.17114},
  archiveprefix = {arXiv}
}

@inproceedings{webrl,
  author = {Zehan Qi and Xiao Liu and Iat Long Iong and Hanyu Lai and Xueqiao Sun and Jiadai Sun and Xinyue Yang and Yu Yang and Shuntian Yao and Wei Xu and Jie Tang and Yuxiao Dong},
  title = {{WebRL}: Training {LLM} Web Agents via Self-Evolving Online Curriculum Reinforcement Learning},
  booktitle = {The Thirteenth International Conference on Learning Representations},
  year = {2025},
  url = {https://proceedings.iclr.cc/paper_files/paper/2025/hash/c66e1fcc9691aae706250638f36f681b-Abstract-Conference.html},
  eprint = {2411.02337},
  archiveprefix = {arXiv}
}

@article{cues,
  author = {Shinji Mai and Yunpeng Zhai and Ziqian Chen and Cheng Chen and Anni Zou and Shuchang Tao and Zhaoyang Liu and Bolin Ding},
  title = {{CuES}: A Curiosity-driven and Environment-grounded Synthesis Framework for Agentic {RL}},
  journal = {arXiv preprint arXiv:2512.01311},
  year = {2025},
  url = {https://arxiv.org/abs/2512.01311},
  eprint = {2512.01311},
  archiveprefix = {arXiv}
}

@article{envharness,
  author = {Chengsong Huang and Zifeng Wang and Rujun Han and Jun Yan and Yanfei Chen and Zoey CuiZhu and Ke Jiang and Peng Xia and Han Yu and Yufan Zhuang and Yifei Ming and Jiaqi Pan and Bhavana Dalvi Mishra and Jiaxin Huang and Burak Gokturk and Tomas Pfister and Chen-Yu Lee},
  title = {{EnvHarness}: Awakening Static Worlds for Agent Learning},
  journal = {arXiv preprint arXiv:2608.19880},
  year = {2026},
  doi = {10.48550/arXiv.2608.19880},
  url = {https://arxiv.org/abs/2608.19880},
  eprint = {2608.19880},
  archiveprefix = {arXiv},
  primaryclass = {cs.AI}
}

@article{terminalenv_evolution,
  author = {Zhiyuan Fan and Tinghao Yu and Yuanjun Cai and Jiang Zhou and Jiangtao Guan and Jincheng Liu and Yun Yang and Dingxin Hu and Zhuo Han and Xing Wu and Feng Zhang and Lilin Wang},
  title = {Environment Evolution for Terminal Agents},
  journal = {arXiv preprint arXiv:2609.04128},
  year = {2026},
  doi = {10.48550/arXiv.2609.04128},
  url = {https://arxiv.org/abs/2609.04128},
  eprint = {2609.04128},
  archiveprefix = {arXiv},
  primaryclass = {cs.AI}
}

@article{seta,
  author = {Qijia Shen and Zhiqi Huang and Vamsidhar Kamanuru and Aznaur Aliev and Jay Rainton and Ahmed Awelkair and Zhichen Zeng and Jiajun Li and Shi Dong and Yueming Yuan and Boyuan Ma and Qizheng Zhang and Jiwei Fu and Yuzhen Mao and Wendong Fan and Ping Nie and Philip Torr and Bernard Ghanem and Changran Hu and Jonathan Lingjie Li and Urmish Thakker and Guohao Li},
  title = {{SETA}: Scaling Environments for Terminal Agents},
  journal = {arXiv preprint arXiv:2607.10891},
  year = {2026},
  url = {https://arxiv.org/abs/2607.10891},
  eprint = {2607.10891},
  archiveprefix = {arXiv}
}

@article{envrethink,
  author = {Yukai Wu and Yuanjing Yang and Le Zhou and Shaokun Han and Haoyu Wang and Zirui Tang and Xuzhou Zhu and Weihuang Zheng and Maxm Pan and Xuanhe Zhou and Fan Wu},
  title = {Breaking the Environment Wall: A Unified Framework for Preparing and Evolving Agent-Native Environments},
  journal = {arXiv preprint arXiv:2609.29773},
  year = {2026},
  url = {https://arxiv.org/abs/2609.29773},
  eprint = {2609.29773},
  archiveprefix = {arXiv}
}

@article{proevolve,
  author = {Guangrui Li and Yaochen Xie and Yi Liu and Ziwei Dong and Xingyuan Pan and Tianqi Zheng and Jason Choi and Michael J. Morais and Binit Jha and Shaunak Mishra and Bingrou Zhou and Chen Luo and Monica Xiao Cheng and Dawn Song},
  title = {The World Won't Stay Still: Programmable Evolution for Agent Benchmarks},
  journal = {arXiv preprint arXiv:2603.05910},
  year = {2026},
  url = {https://arxiv.org/abs/2603.05910},
  eprint = {2603.05910},
  archiveprefix = {arXiv}
}

@article{autoenv,
  author = {Jiayi Zhang and Yiran Peng and Fanqi Kong and Cheng Yang and Yifan Wu and Zhaoyang Yu and Jinyu Xiang and Jianhao Ruan and Jinlin Wang and Maojia Song and HongZhang Liu and Xiangru Tang and Bang Liu and Chenglin Wu and Yuyu Luo},
  title = {{AutoEnv}: Automated Environments for Measuring Cross-Environment Agent Learning},
  journal = {arXiv preprint arXiv:2511.19304},
  year = {2025},
  doi = {10.48550/arXiv.2511.19304},
  url = {https://arxiv.org/abs/2511.19304},
  eprint = {2511.19304},
  archiveprefix = {arXiv},
  primaryclass = {cs.AI}
}

@article{poet,
  author = {Rui Wang and Joel Lehman and Jeff Clune and Kenneth O. Stanley},
  title = {Paired Open-Ended Trailblazer ({POET}): Endlessly Generating Increasingly Complex and Diverse Learning Environments and Their Solutions},
  journal = {arXiv preprint arXiv:1901.01753},
  year = {2019},
  doi = {10.48550/arXiv.1901.01753},
  url = {https://arxiv.org/abs/1901.01753},
  eprint = {1901.01753},
  archiveprefix = {arXiv},
  primaryclass = {cs.NE}
}

@inproceedings{paired,
  author = {Michael Dennis and Natasha Jaques and Eugene Vinitsky and Alexandre Bayen and Stuart Russell and Andrew Critch and Sergey Levine},
  title = {Emergent Complexity and Zero-shot Transfer via Unsupervised Environment Design},
  booktitle = {Advances in Neural Information Processing Systems},
  volume = {33},
  publisher = {Curran Associates, Inc.},
  year = {2020},
  url = {https://papers.nips.cc/paper/2020/hash/985e9a46e10005356bbaf194249f6856-Abstract.html},
  eprint = {2012.02096},
  archiveprefix = {arXiv}
}

@inproceedings{accel,
  author = {Jack Parker-Holder and Minqi Jiang and Michael Dennis and Mikayel Samvelyan and Jakob Foerster and Edward Grefenstette and Tim Rockt{\"a}schel},
  title = {Evolving Curricula with Regret-Based Environment Design},
  booktitle = {Proceedings of the 39th International Conference on Machine Learning},
  series = {Proceedings of Machine Learning Research},
  volume = {162},
  pages = {17473--17498},
  publisher = {PMLR},
  year = {2022},
  url = {https://proceedings.mlr.press/v162/parker-holder22a.html},
  eprint = {2203.01302},
  archiveprefix = {arXiv}
}

@article{llmpoet,
  author = {Fuma Aki and Riku Ikeda and Takumi Saito and Ciaran Regan and Mizuki Oka},
  title = {{LLM-POET}: Evolving Complex Environments using Large Language Models},
  journal = {arXiv preprint arXiv:2406.04663},
  year = {2024},
  url = {https://arxiv.org/abs/2406.04663},
  eprint = {2406.04663},
  archiveprefix = {arXiv}
}

@inproceedings{omniepic,
  author = {Maxence Faldor and Jenny Zhang and Antoine Cully and Jeff Clune},
  title = {{OMNI-EPIC}: Open-endedness via Models of human Notions of Interestingness with Environments Programmed in Code},
  booktitle = {International Conference on Learning Representations},
  year = {2025},
  url = {https://proceedings.iclr.cc/paper_files/paper/2025/hash/d40d7cbe7210f8a13ea0149eeae9c6de-Abstract-Conference.html},
  eprint = {2405.15568},
  archiveprefix = {arXiv},
  primaryclass = {cs.AI}
}

@article{traineetotrainer,
  author = {Chen, Chao and Li, Chengzu and Li, Zhiwei and Liu, Yinhong and Guo, Zhijiang},
  title = {From Trainee to Trainer: {LLM}-Designed Training Environment for {RL} with Multi-Agent Reasoning},
  journal = {arXiv preprint arXiv:2606.17682},
  year = {2026},
  doi = {10.48550/arXiv.2606.17682},
  url = {https://arxiv.org/abs/2606.17682},
  eprint = {2606.17682},
  archiveprefix = {arXiv},
  primaryclass = {cs.CL}
}

@inproceedings{vlaw,
  author = {Yanjiang Guo and Tony Lee and Lucy Xiaoyang Shi and Jianyu Chen and Percy Liang and Chelsea Finn},
  title = {{VLAW}: Iterative Co-Improvement of Vision-Language-Action Policy and World Model},
  booktitle = {Proceedings of the 43rd International Conference on Machine Learning},
  series = {Proceedings of Machine Learning Research},
  volume = {306},
  pages = {37998--38011},
  publisher = {PMLR},
  year = {2026},
  url = {https://proceedings.mlr.press/v306/guo26i.html},
  eprint = {2602.12063},
  archiveprefix = {arXiv}
}

@article{worldvlaloop,
  author = {Liu, Xiaokang and Bai, Zechen and Ci, Hai and Ma, Kevin Yuchen and Shou, Mike Zheng},
  title = {{World-VLA-Loop}: Closed-Loop Learning of Video World Model and {VLA} Policy},
  journal = {arXiv preprint arXiv:2602.06508},
  year = {2026},
  doi = {10.48550/arXiv.2602.06508},
  url = {https://arxiv.org/abs/2602.06508},
  eprint = {2602.06508},
  archiveprefix = {arXiv},
  primaryclass = {cs.RO}
}

@article{janus,
  author = {Mengdi Liu and Chenghao Jia and Hong Chang and Mingjie He and Shiguang Shan and Xilin Chen},
  title = {{JANUS}: Enhancing Long-Horizon Autonomous Research with Co-Evolving World Models},
  journal = {arXiv preprint arXiv:2610.13194},
  year = {2026},
  url = {https://arxiv.org/abs/2610.13194},
  eprint = {2610.13194},
  archiveprefix = {arXiv}
}

@inproceedings{human_preferences,
  author = {Paul F. Christiano and Jan Leike and Tom B. Brown and Miljan Martic and Shane Legg and Dario Amodei},
  title = {Deep Reinforcement Learning from Human Preferences},
  booktitle = {Advances in Neural Information Processing Systems},
  volume = {30},
  year = {2017},
  doi = {10.48550/arXiv.1706.03741},
  url = {https://proceedings.neurips.cc/paper/2017/hash/d5e2c0adad503c91f91df240d0cd4e49-Abstract.html}
}

@inproceedings{summarize,
  author = {Nisan Stiennon and Long Ouyang and Jeff Wu and Daniel M. Ziegler and Ryan Lowe and Chelsea Voss and Alec Radford and Dario Amodei and Paul Christiano},
  title = {Learning to summarize from human feedback},
  booktitle = {Advances in Neural Information Processing Systems},
  volume = {33},
  year = {2020},
  doi = {10.48550/arXiv.2009.01325},
  url = {https://proceedings.neurips.cc/paper_files/paper/2020/file/1f89885d556929e98d3ef9b86448f951-Paper.pdf}
}

@article{webgpt,
  author = {Reiichiro Nakano and Jacob Hilton and Suchir Balaji and Jeff Wu and Long Ouyang and Christina Kim and Christopher Hesse and Shantanu Jain and Vineet Kosaraju and William Saunders and Xu Jiang and Karl Cobbe and Tyna Eloundou and Gretchen Krueger and Kevin Button and Matthew Knight and Benjamin Chess and John Schulman},
  title = {{WebGPT}: Browser-assisted question-answering with human feedback},
  journal = {arXiv preprint arXiv:2112.09332},
  year = {2021},
  doi = {10.48550/arXiv.2112.09332},
  url = {https://arxiv.org/abs/2112.09332v3},
  eprint = {2112.09332},
  archiveprefix = {arXiv}
}

@article{helpful_harmless,
  author = {Yuntao Bai and Andy Jones and Kamal Ndousse and Amanda Askell and Anna Chen and Nova DasSarma and Dawn Drain and Stanislav Fort and Deep Ganguli and Tom Henighan and Nicholas Joseph and Saurav Kadavath and Jackson Kernion and Tom Conerly and Sheer El-Showk and Nelson Elhage and Zac Hatfield-Dodds and Danny Hernandez and Tristan Hume and Scott Johnston and Shauna Kravec and Liane Lovitt and Neel Nanda and Catherine Olsson and Dario Amodei and Tom Brown and Jack Clark and Sam McCandlish and Chris Olah and Ben Mann and Jared Kaplan},
  title = {Training a Helpful and Harmless Assistant with Reinforcement Learning from Human Feedback},
  journal = {arXiv preprint arXiv:2204.05862},
  year = {2022},
  doi = {10.48550/arXiv.2204.05862},
  url = {https://arxiv.org/abs/2204.05862v1},
  eprint = {2204.05862},
  archiveprefix = {arXiv}
}

@inproceedings{pebble,
  author = {Kimin Lee and Laura M Smith and Pieter Abbeel},
  title = {{PEBBLE}: Feedback-Efficient Interactive Reinforcement Learning via Relabeling Experience and Unsupervised Pre-training},
  booktitle = {Proceedings of the 38th International Conference on Machine Learning},
  volume = {139},
  pages = {6152--6163},
  year = {2021},
  url = {https://proceedings.mlr.press/v139/lee21i.html}
}

@inproceedings{surf,
  author = {Jongjin Park and Younggyo Seo and Jinwoo Shin and Honglak Lee and Pieter Abbeel and Kimin Lee},
  title = {{SURF}: Semi-supervised Reward Learning with Data Augmentation for Feedback-efficient Preference-based Reinforcement Learning},
  booktitle = {International Conference on Learning Representations},
  year = {2022},
  doi = {10.48550/arXiv.2203.10050},
  url = {https://arxiv.org/abs/2203.10050v1}
}

@article{active_preference,
  author = {William Muldrew and Peter Hayes and Mingtian Zhang and David Barber},
  title = {Active Preference Learning for Large Language Models},
  journal = {arXiv preprint arXiv:2402.08114},
  year = {2024},
  doi = {10.48550/arXiv.2402.08114},
  url = {https://arxiv.org/abs/2402.08114v2},
  eprint = {2402.08114},
  archiveprefix = {arXiv}
}

@inproceedings{active_reward,
  author = {Yunyi Shen and Hao Sun and Jean-Francois Ton},
  title = {Active Reward Modeling: Adaptive Preference Labeling for Large Language Model Alignment},
  booktitle = {Proceedings of the 42nd International Conference on Machine Learning},
  volume = {267},
  pages = {54410--54430},
  year = {2025},
  url = {https://proceedings.mlr.press/v267/shen25c.html}
}

@inproceedings{pilaf,
  author = {Yunzhen Feng and Ariel Kwiatkowski and Kunhao Zheng and Julia Kempe and Yaqi Duan},
  title = {{PILAF}: Optimal Human Preference Sampling for Reward Modeling},
  booktitle = {Proceedings of the 42nd International Conference on Machine Learning},
  volume = {267},
  pages = {16744--16776},
  year = {2025},
  url = {https://proceedings.mlr.press/v267/feng25g.html}
}

@article{alphacode,
  author = {Yujia Li and David Choi and Junyoung Chung and Nate Kushman and Julian Schrittwieser and R{\'e}mi Leblond and Tom Eccles and James Keeling and Felix Gimeno and Agustin Dal Lago and Thomas Hubert and Peter Choy and Cyprien de Masson d'Autume and Igor Babuschkin and Xinyun Chen and Po-Sen Huang and Johannes Welbl and Sven Gowal and Alexey Cherepanov and James Molloy and Daniel J. Mankowitz and Esme Sutherland Robson and Pushmeet Kohli and Nando de Freitas and Koray Kavukcuoglu and Oriol Vinyals},
  title = {Competition-Level Code Generation with {AlphaCode}},
  journal = {arXiv preprint arXiv:2203.07814},
  year = {2022},
  doi = {10.48550/arXiv.2203.07814},
  url = {https://arxiv.org/abs/2203.07814v1},
  eprint = {2203.07814},
  archiveprefix = {arXiv}
}

@inproceedings{codet,
  author = {Bei Chen and Fengji Zhang and Anh Nguyen and Daoguang Zan and Zeqi Lin and Jian-Guang Lou and Weizhu Chen},
  title = {{CodeT}: Code Generation with Generated Tests},
  booktitle = {International Conference on Learning Representations},
  year = {2023},
  doi = {10.48550/arXiv.2207.10397},
  url = {https://arxiv.org/abs/2207.10397v2}
}

@article{tulu3,
  author = {Nathan Lambert and Jacob Morrison and Valentina Pyatkin and Shengyi Huang and Hamish Ivison and Faeze Brahman and Lester James V. Miranda and Alisa Liu and Nouha Dziri and Shane Lyu and Yuling Gu and Saumya Malik and Victoria Graf and Jena D. Hwang and Jiangjiang Yang and Ronan Le Bras and Oyvind Tafjord and Chris Wilhelm and Luca Soldaini and Noah A. Smith and Yizhong Wang and Pradeep Dasigi and Hannaneh Hajishirzi},
  title = {Tulu 3: Pushing Frontiers in Open Language Model Post-Training},
  journal = {arXiv preprint arXiv:2411.15124},
  year = {2024},
  doi = {10.48550/arXiv.2411.15124},
  url = {https://arxiv.org/abs/2411.15124v5},
  eprint = {2411.15124},
  archiveprefix = {arXiv}
}

@article{rltf,
  author = {Jiate Liu and Yiqin Zhu and Kaiwen Xiao and Qiang Fu and Xiao Han and Wei Yang and Deheng Ye},
  title = {{RLTF}: Reinforcement Learning from Unit Test Feedback},
  journal = {Transactions on Machine Learning Research},
  year = {2023},
  url = {https://openreview.net/forum?id=hjYmsV6nXZ}
}

@inproceedings{rlef,
  author = {Jonas Gehring and Kunhao Zheng and Jade Copet and Vegard Mella and Taco Cohen and Gabriel Synnaeve},
  title = {{RLEF}: Grounding Code {LLMs} in Execution Feedback with Reinforcement Learning},
  booktitle = {Proceedings of the 42nd International Conference on Machine Learning},
  volume = {267},
  pages = {19034--19055},
  year = {2025},
  url = {https://proceedings.mlr.press/v267/gehring25a.html}
}

@inproceedings{geval,
  author = {Yang Liu and Dan Iter and Yichong Xu and Shuohang Wang and Ruochen Xu and Chenguang Zhu},
  title = {{G-Eval}: {NLG} Evaluation using {GPT-4} with Better Human Alignment},
  booktitle = {Proceedings of the 2023 Conference on Empirical Methods in Natural Language Processing},
  pages = {2511--2522},
  year = {2023},
  doi = {10.18653/v1/2023.emnlp-main.153},
  url = {https://aclanthology.org/2023.emnlp-main.153/}
}

@inproceedings{pandalm,
  author = {Yidong Wang and Zhuohao Yu and Zhengran Zeng and Linyi Yang and Cunxiang Wang and Hao Chen and Chaoya Jiang and Rui Xie and Jindong Wang and Xing Xie and Wei Ye and Shikun Zhang and Yue Zhang},
  title = {{PandaLM}: An Automatic Evaluation Benchmark for {LLM} Instruction Tuning Optimization},
  booktitle = {International Conference on Learning Representations},
  year = {2024},
  doi = {10.48550/arXiv.2306.05087},
  url = {https://arxiv.org/abs/2306.05087v2}
}

@inproceedings{judgelm,
  author = {Lianghui Zhu and Xinggang Wang and Xinlong Wang},
  title = {{JudgeLM}: Fine-tuned Large Language Models are Scalable Judges},
  booktitle = {International Conference on Learning Representations},
  year = {2025},
  doi = {10.48550/arXiv.2310.17631},
  url = {https://arxiv.org/abs/2310.17631v2}
}

@inproceedings{flask,
  author = {Seonghyeon Ye and Doyoung Kim and Sungdong Kim and Hyeonbin Hwang and Seungone Kim and Yongrae Jo and James Thorne and Juho Kim and Minjoon Seo},
  title = {{FLASK}: Fine-grained Language Model Evaluation based on Alignment Skill Sets},
  booktitle = {International Conference on Learning Representations},
  year = {2024},
  doi = {10.48550/arXiv.2307.10928},
  url = {https://arxiv.org/abs/2307.10928v4}
}

@inproceedings{prometheus,
  author = {Seungone Kim and Jamin Shin and Yejin Cho and Joel Jang and Shayne Longpre and Hwaran Lee and Sangdoo Yun and Seongjin Shin and Sungdong Kim and James Thorne and Minjoon Seo},
  title = {{Prometheus}: Inducing Fine-grained Evaluation Capability in Language Models},
  booktitle = {International Conference on Learning Representations},
  year = {2024},
  doi = {10.48550/arXiv.2310.08491},
  url = {https://arxiv.org/abs/2310.08491v2}
}

@inproceedings{prometheus2,
  author = {Seungone Kim and Juyoung Suk and Shayne Longpre and Bill Yuchen Lin and Jamin Shin and Sean Welleck and Graham Neubig and Moontae Lee and Kyungjae Lee and Minjoon Seo},
  title = {Prometheus 2: An Open Source Language Model Specialized in Evaluating Other Language Models},
  booktitle = {Proceedings of the 2024 Conference on Empirical Methods in Natural Language Processing},
  pages = {4334--4353},
  year = {2024},
  doi = {10.18653/v1/2024.emnlp-main.248},
  url = {https://aclanthology.org/2024.emnlp-main.248/}
}

@article{criticgpt,
  author = {Nat McAleese and Rai Michael Pokorny and Juan Felipe Ceron Uribe and Evgenia Nitishinskaya and Maja Trebacz and Jan Leike},
  title = {{LLM} Critics Help Catch {LLM} Bugs},
  journal = {arXiv preprint arXiv:2407.00215},
  year = {2024},
  doi = {10.48550/arXiv.2407.00215},
  url = {https://arxiv.org/abs/2407.00215v1},
  eprint = {2407.00215},
  archiveprefix = {arXiv}
}

@article{constitutional,
  author = {Yuntao Bai and Saurav Kadavath and Sandipan Kundu and Amanda Askell and Jackson Kernion and Andy Jones and Anna Chen and Anna Goldie and Azalia Mirhoseini and Cameron McKinnon and Carol Chen and Catherine Olsson and Christopher Olah and Danny Hernandez and Dawn Drain and Deep Ganguli and Dustin Li and Eli Tran-Johnson and Ethan Perez and Jamie Kerr and Jared Mueller and Jeffrey Ladish and Joshua Landau and Kamal Ndousse and Kamile Lukosuite and Liane Lovitt and Michael Sellitto and Nelson Elhage and Nicholas Schiefer and Noemi Mercado and Nova DasSarma and Robert Lasenby and Robin Larson and Sam Ringer and Scott Johnston and Shauna Kravec and Sheer El Showk and Stanislav Fort and Tamera Lanham and Timothy Telleen-Lawton and Tom Conerly and Tom Henighan and Tristan Hume and Samuel R. Bowman and Zac Hatfield-Dodds and Ben Mann and Dario Amodei and Nicholas Joseph and Sam McCandlish and Tom Brown and Jared Kaplan},
  title = {Constitutional {AI}: Harmlessness from {AI} Feedback},
  journal = {arXiv preprint arXiv:2212.08073},
  year = {2022},
  doi = {10.48550/arXiv.2212.08073},
  url = {https://arxiv.org/abs/2212.08073v1},
  eprint = {2212.08073},
  archiveprefix = {arXiv}
}

@inproceedings{ultrafeedback,
  author = {Ganqu Cui and Lifan Yuan and Ning Ding and Guanming Yao and Bingxiang He and Wei Zhu and Yuan Ni and Guotong Xie and Ruobing Xie and Yankai Lin and Zhiyuan Liu and Maosong Sun},
  title = {{UltraFeedback}: Boosting Language Models with Scaled {AI} Feedback},
  booktitle = {Proceedings of the 41st International Conference on Machine Learning},
  volume = {235},
  pages = {9722--9744},
  year = {2024},
  doi = {10.48550/arXiv.2310.01377},
  url = {https://proceedings.mlr.press/v235/cui24f.html}
}

@inproceedings{rlaif,
  author = {Harrison Lee and Samrat Phatale and Hassan Mansoor and Thomas Mesnard and Johan Ferret and Kellie Ren Lu and Colton Bishop and Ethan Hall and Victor Carbune and Abhinav Rastogi and Sushant Prakash},
  title = {{RLAIF} vs. {RLHF}: Scaling Reinforcement Learning from Human Feedback with {AI} Feedback},
  booktitle = {Proceedings of the 41st International Conference on Machine Learning},
  volume = {235},
  pages = {26874--26901},
  year = {2024},
  url = {https://proceedings.mlr.press/v235/lee24t.html}
}

@inproceedings{fair_evaluators,
  author = {Peiyi Wang and Lei Li and Liang Chen and Zefan Cai and Dawei Zhu and Binghuai Lin and Yunbo Cao and Lingpeng Kong and Qi Liu and Tianyu Liu and Zhifang Sui},
  title = {Large Language Models are not Fair Evaluators},
  booktitle = {Proceedings of the 62nd Annual Meeting of the Association for Computational Linguistics (Volume 1: Long Papers)},
  pages = {9440--9450},
  year = {2024},
  doi = {10.18653/v1/2024.acl-long.511},
  url = {https://aclanthology.org/2024.acl-long.511/}
}

@article{judge_survey,
  author = {Haitao Li and Qian Dong and Junjie Chen and Huixue Su and Yujia Zhou and Qingyao Ai and Ziyi Ye and Yiqun Liu},
  title = {{LLMs-as-Judges}: A Comprehensive Survey on {LLM-based} Evaluation Methods},
  journal = {arXiv preprint arXiv:2412.05579},
  year = {2024},
  doi = {10.48550/arXiv.2412.05579},
  url = {https://arxiv.org/abs/2412.05579v2},
  eprint = {2412.05579},
  archiveprefix = {arXiv}
}

@inproceedings{jetts,
  author = {Yilun Zhou and Austin Xu and Peifeng Wang and Caiming Xiong and Shafiq Joty},
  title = {Evaluating Judges as Evaluators: The {JETTS} Benchmark of {LLM-as-Judges} as Test-Time Scaling Evaluators},
  booktitle = {Proceedings of the 42nd International Conference on Machine Learning},
  volume = {267},
  pages = {79436--79471},
  year = {2025},
  url = {https://proceedings.mlr.press/v267/zhou25af.html}
}

@inproceedings{agent_judge,
  author = {Mingchen Zhuge and Changsheng Zhao and Dylan R. Ashley and Wenyi Wang and Dmitrii Khizbullin and Yunyang Xiong and Zechun Liu and Ernie Chang and Raghuraman Krishnamoorthi and Yuandong Tian and Yangyang Shi and Vikas Chandra and J{\"u}rgen Schmidhuber},
  title = {{Agent-as-a-Judge}: Evaluate Agents with Agents},
  booktitle = {Proceedings of the 42nd International Conference on Machine Learning},
  volume = {267},
  pages = {80569--80611},
  year = {2025},
  url = {https://proceedings.mlr.press/v267/zhuge25a.html}
}

@article{codevisionary,
  author = {Xinchen Wang and Pengfei Gao and Chao Peng and Ruida Hu and Cuiyun Gao},
  title = {{CodeVisionary}: An Agent-based Framework for Evaluating Large Language Models in Code Generation},
  journal = {arXiv preprint arXiv:2504.13472},
  year = {2025},
  doi = {10.48550/arXiv.2504.13472},
  url = {https://arxiv.org/abs/2504.13472v2},
  eprint = {2504.13472},
  archiveprefix = {arXiv}
}

@article{mind2web2,
  author = {Boyu Gou and Zanming Huang and Yuting Ning and Yu Gu and Michael Lin and Weijian Qi and Andrei Kopanev and Botao Yu and Bernal Jim{\'e}nez Guti{\'e}rrez and Yiheng Shu and Chan Hee Song and Jiaman Wu and Shijie Chen and Hanane Nour Moussa and Tianshu Zhang and Jian Xie and Yifei Li and Tianci Xue and Zeyi Liao and Kai Zhang and Boyuan Zheng and Zhaowei Cai and Viktor Rozgic and Morteza Ziyadi and Huan Sun and Yu Su},
  title = {{Mind2Web 2}: Evaluating Agentic Search with Agent-as-a-Judge},
  journal = {arXiv preprint arXiv:2506.21506},
  year = {2025},
  doi = {10.48550/arXiv.2506.21506},
  url = {https://arxiv.org/abs/2506.21506v2},
  eprint = {2506.21506},
  archiveprefix = {arXiv}
}

@article{intellagent,
  author = {Elad Levi and Ilan Kadar},
  title = {{IntellAgent}: A Multi-Agent Framework for Evaluating Conversational {AI} Systems},
  journal = {arXiv preprint arXiv:2501.11067},
  year = {2025},
  doi = {10.48550/arXiv.2501.11067},
  url = {https://arxiv.org/abs/2501.11067v1},
  eprint = {2501.11067},
  archiveprefix = {arXiv}
}

@article{auto_eval,
  author = {Roshita Bhonsle and Rishav Dutta and Sneha Vavilapalli and Harsh Seth and Abubakarr Jaye and Yapei Chang and Mukund Rungta and Emmanuel Aboah Boateng and Sadid Hasan and Ehi Nosakhare and Soundar Srinivasan},
  title = {Auto-Eval Judge: Towards a General Agentic Framework for Task Completion Evaluation},
  journal = {arXiv preprint arXiv:2508.05508},
  year = {2025},
  doi = {10.48550/arXiv.2508.05508},
  url = {https://arxiv.org/abs/2508.05508v1},
  eprint = {2508.05508},
  archiveprefix = {arXiv}
}

@article{ajbench,
  author = {Wentao Shi and Yu Wang and Yuyang Zhao and Yuxin Chen and Fuli Feng and Xueyuan Hao and Xi Su and Qi Gu and Hui Su and Xunliang Cai and Xiangnan He},
  title = {{AJ-Bench}: Benchmarking Agent-as-a-Judge for Environment-Aware Evaluation},
  journal = {arXiv preprint arXiv:2604.18240},
  year = {2026},
  doi = {10.48550/arXiv.2604.18240},
  url = {https://arxiv.org/abs/2604.18240v1},
  eprint = {2604.18240},
  archiveprefix = {arXiv}
}

@inproceedings{themis,
  author = {Lei Li and Yekun Chai and Shuohuan Wang and Yu Sun and Hao Tian and Ningyu Zhang and Hua Wu},
  title = {Tool-Augmented Reward Modeling},
  booktitle = {International Conference on Learning Representations},
  year = {2024},
  doi = {10.48550/arXiv.2310.01045},
  url = {https://arxiv.org/abs/2310.01045v2}
}

@inproceedings{verifiagent,
  author = {Jiuzhou Han and Wray Buntine and Ehsan Shareghi},
  title = {{VerifiAgent}: a Unified Verification Agent in Language Model Reasoning},
  booktitle = {Findings of the Association for Computational Linguistics: EMNLP 2025},
  pages = {16410--16431},
  year = {2025},
  doi = {10.18653/v1/2025.findings-emnlp.891},
  url = {https://aclanthology.org/2025.findings-emnlp.891/}
}

@inproceedings{evalagent,
  author = {Manya Wadhwa and Zayne Sprague and Chaitanya Malaviya and Philippe Laban and Junyi Jessy Li and Greg Durrett},
  title = {{EvalAgent}: Discovering Implicit Evaluation Criteria from the Web},
  booktitle = {Conference on Language Modeling},
  year = {2025},
  doi = {10.48550/arXiv.2504.15219},
  url = {https://arxiv.org/abs/2504.15219v2}
}

@article{fact_audit,
  author = {Hongzhan Lin and Yang Deng and Yuxuan Gu and Wenxuan Zhang and Jing Ma and See-Kiong Ng and Tat-Seng Chua},
  title = {{FACT-AUDIT}: An Adaptive Multi-Agent Framework for Dynamic Fact-Checking Evaluation of Large Language Models},
  journal = {arXiv preprint arXiv:2502.17924},
  year = {2025},
  doi = {10.48550/arXiv.2502.17924},
  url = {https://arxiv.org/abs/2502.17924v2},
  eprint = {2502.17924},
  archiveprefix = {arXiv}
}

@article{gui_pra,
  author = {Tao Xiong and Xavier Hu and Yurun Chen and Yuhang Liu and Changqiao Wu and Pengzhi Gao and Wei Liu and Jian Luan and Shengyu Zhang},
  title = {{GUI-PRA}: Process Reward Agent for {GUI} Tasks},
  journal = {arXiv preprint arXiv:2509.23263},
  year = {2025},
  doi = {10.48550/arXiv.2509.23263},
  url = {https://arxiv.org/abs/2509.23263v3},
  eprint = {2509.23263},
  archiveprefix = {arXiv}
}

@article{vagen,
  author = {Chaoqun Cui and Jing Huang and Shijing Wang and Liming Zheng and Qingchao Kong and Zhixiong Zeng},
  title = {Agentic Reward Modeling: Verifying {GUI} Agent via Progressive Trajectory-Grounded Interaction},
  journal = {arXiv preprint arXiv:2602.00575},
  year = {2026},
  doi = {10.48550/arXiv.2602.00575},
  url = {https://arxiv.org/abs/2602.00575v2},
  eprint = {2602.00575},
  archiveprefix = {arXiv}
}

@inproceedings{rewardagent,
  author = {Hao Peng and others},
  title = {Agentic Reward Modeling: Integrating Human Preferences with Verifiable Correctness Signals for Reliable Reward Systems},
  booktitle = {Proceedings of the 63rd Annual Meeting of the Association for Computational Linguistics (Volume 1: Long Papers)},
  pages = {15934--15949},
  year = {2025},
  doi = {10.18653/v1/2025.acl-long.775},
  url = {https://aclanthology.org/2025.acl-long.775/}
}

@article{openreward,
  author = {Ziyou Hu and Zhengliang Shi and Minghang Zhu and Haitao Li and Teng Sun and Pengjie Ren and Suzan Verberne and Zhaochun Ren},
  title = {{OpenReward}: Learning to Reward Long-form Agentic Tasks via Reinforcement Learning},
  journal = {arXiv preprint arXiv:2510.24636},
  year = {2025},
  doi = {10.48550/arXiv.2510.24636},
  url = {https://arxiv.org/abs/2510.24636v3},
  eprint = {2510.24636},
  archiveprefix = {arXiv}
}

@article{agentv_rl,
  author = {Jiazheng Zhang and Ziche Fu and Zhiheng Xi and Wenqing Jing and Mingxu Chai and Wei He and Guoqiang Zhang and Chenghao Fan and Chenxin An and Wenxiang Chen and Zhicheng Liu and Haojie Pan and Dingwei Zhu and Tao Gui and Qi Zhang and Xuanjing Huang},
  title = {{AgentV-RL}: Scaling Reward Modeling with Agentic Verifier},
  journal = {arXiv preprint arXiv:2604.16004},
  year = {2026},
  doi = {10.48550/arXiv.2604.16004},
  url = {https://arxiv.org/abs/2604.16004v1},
  eprint = {2604.16004},
  archiveprefix = {arXiv}
}

@article{scaling_verifier,
  author = {Zeyao Ma and Jing Zhang and Xiaokang Zhang and Jiaxi Yang and Zongmeng Zhang and Jiajun Zhang and Yuheng Jing and Lei Zhang and Hao Zheng and Wenting Zhao and Junyang Lin and Binyuan Hui},
  title = {Scaling Agentic Verifier for Competitive Coding},
  journal = {arXiv preprint arXiv:2602.04254},
  year = {2026},
  doi = {10.48550/arXiv.2602.04254},
  url = {https://arxiv.org/abs/2602.04254v1},
  eprint = {2602.04254},
  archiveprefix = {arXiv}
}

@inproceedings{pairranker,
  author = {Dongfu Jiang and Xiang Ren and Bill Yuchen Lin},
  title = {{LLM-Blender}: Ensembling Large Language Models with Pairwise Ranking and Generative Fusion},
  booktitle = {Proceedings of the 61st Annual Meeting of the Association for Computational Linguistics (Volume 1: Long Papers)},
  pages = {14165--14178},
  year = {2023},
  doi = {10.18653/v1/2023.acl-long.792},
  url = {https://aclanthology.org/2023.acl-long.792/}
}

@inproceedings{minicheck,
  author = {Liyan Tang and Philippe Laban and Greg Durrett},
  title = {{MiniCheck}: Efficient Fact-Checking of {LLMs} on Grounding Documents},
  booktitle = {Proceedings of the 2024 Conference on Empirical Methods in Natural Language Processing},
  pages = {8818--8847},
  year = {2024},
  doi = {10.18653/v1/2024.emnlp-main.499},
  url = {https://aclanthology.org/2024.emnlp-main.499/}
}

@article{bert_judge,
  author = {Hippolyte Gisserot-Boukhlef and Nicolas Boizard and Emmanuel Malherbe and C{\'e}line Hudelot and Pierre Colombo},
  title = {{BERT-as-a-Judge}: A Robust Alternative to Lexical Methods for Efficient Reference-Based {LLM} Evaluation},
  journal = {arXiv preprint arXiv:2604.09497},
  year = {2026},
  doi = {10.48550/arXiv.2604.09497},
  url = {https://arxiv.org/abs/2604.09497v2},
  eprint = {2604.09497},
  archiveprefix = {arXiv}
}

@article{luna2,
  author = {Vatsal Goel and Rishon Dsouza and Nikhil Ega and Amey Ramesh Rambatla and Rob Friel and Shuai Shao and Yash Sheth},
  title = {{Luna-2}: Scalable Single-Token Evaluation with Small Language Models},
  journal = {arXiv preprint arXiv:2602.18583},
  year = {2026},
  doi = {10.48550/arXiv.2602.18583},
  url = {https://arxiv.org/abs/2602.18583v1},
  eprint = {2602.18583},
  archiveprefix = {arXiv}
}

@inproceedings{skywork,
  author = {Chris Yuhao Liu and Liang Zeng and Yuzhen Xiao and Jujie He and Jiacai Liu and Chaojie Wang and Rui Yan and Wei Shen and Fuxiang Zhang and Jiacheng Xu and Yang Liu and Yahui Zhou},
  title = {{Skywork-Reward-V2}: Scaling Preference Data Curation via {Human-AI} Synergy},
  booktitle = {International Conference on Learning Representations},
  year = {2026},
  doi = {10.48550/arXiv.2507.01352},
  url = {https://arxiv.org/abs/2507.01352v3}
}

@inproceedings{rewardbench,
  author = {Nathan Lambert and others},
  title = {{RewardBench}: Evaluating Reward Models for Language Modeling},
  booktitle = {Findings of the Association for Computational Linguistics: NAACL 2025},
  pages = {1755--1797},
  year = {2025},
  doi = {10.18653/v1/2025.findings-naacl.96},
  url = {https://aclanthology.org/2025.findings-naacl.96/}
}

@article{jev,
  author = {Yubo Li and Yidi Miao and Ramayya Krishnan and Rema Padman},
  title = {{JEV-as-a-Judge}: Accept When Confident, Escalate When Unsure},
  journal = {arXiv preprint arXiv:2609.26550},
  year = {2026},
  doi = {10.48550/arXiv.2609.26550},
  url = {https://arxiv.org/abs/2609.26550v5},
  eprint = {2609.26550},
  archiveprefix = {arXiv}
}

@article{trust_escalate,
  author = {Jaehun Jung and Faeze Brahman and Yejin Choi},
  title = {Trust or Escalate: {LLM} Judges with Provable Guarantees for Human Agreement},
  journal = {arXiv preprint arXiv:2407.18370},
  year = {2024},
  doi = {10.48550/arXiv.2407.18370},
  url = {https://arxiv.org/abs/2407.18370v1},
  eprint = {2407.18370},
  archiveprefix = {arXiv}
}

@inproceedings{ask_strong,
  author = {Zhenghao Xu and Qin Lu and Qingru Zhang and Liang Qiu and Ilgee Hong and Changlong Yu and Wenlin Yao and Yao Liu and Haoming Jiang and Lihong Li and Hyokun Yun and Tuo Zhao},
  title = {Ask a Strong {LLM} Judge when Your Reward Model is Uncertain},
  booktitle = {Advances in Neural Information Processing Systems},
  year = {2025},
  doi = {10.48550/arXiv.2510.20369},
  url = {https://arxiv.org/abs/2510.20369v1}
}

@article{jev_rubric,
  author = {Delip Rao and Chris Callison-Burch},
  title = {Decision Models vs. {LLMs} as Rubric Judges: Cheaper, Faster, and Wrong in the Same Places},
  journal = {arXiv preprint arXiv:2609.29769},
  year = {2026},
  doi = {10.48550/arXiv.2609.29769},
  url = {https://arxiv.org/abs/2609.29769v3},
  eprint = {2609.29769},
  archiveprefix = {arXiv}
}

@article{jev_benchmark,
  author = {Tobias Deu{\ss}er and Lorenz Sparrenberg and Rafet Sifa},
  title = {Evaluating and Benchmarking the System One Model {Jev}},
  journal = {arXiv preprint arXiv:2609.37647},
  year = {2026},
  doi = {10.48550/arXiv.2609.37647},
  url = {https://arxiv.org/abs/2609.37647v1},
  eprint = {2609.37647},
  archiveprefix = {arXiv}
}

@article{dynamic_rubric,
  author = {Beining Wang and Weihang Su and Hongtao Tian and Hao Kong and Tao Yang and Ting Yao and Qingyi Pan and Yueyue Wu and Qingyao Ai and Min Zhang and Yiqun Liu},
  title = {Co-Evolving {LLM} Evaluators and Policies via {DynamicRubric}},
  journal = {arXiv preprint arXiv:2607.20083},
  year = {2026},
  doi = {10.48550/arXiv.2607.20083},
  url = {https://arxiv.org/abs/2607.20083v2},
  eprint = {2607.20083},
  archiveprefix = {arXiv}
}

@article{evo_rubric,
  author = {Xin Guan and Xiaomeng Hu and Shen Huang and Zhenyi Wang and Bo Zhang and Zijian Li and Pengjun Xie and Bo Liu and Jiuxin Cao},
  title = {{EvoRubric}: Self-Evolving Rubric-Driven {RL} for Open-Ended Generation},
  journal = {arXiv preprint arXiv:2605.29847},
  year = {2026},
  doi = {10.48550/arXiv.2605.29847},
  url = {https://arxiv.org/abs/2605.29847},
  eprint = {2605.29847},
  archiveprefix = {arXiv}
}

@article{genrubric,
  author = {Yifan Chen and Haitao Li and Qingyao Ai and Fengbin Zhu and Tat-Seng Chua and Min Zhang and Yiqun Liu},
  title = {{GenRubric}: Self-Evolving Rubric Generation for Scalable {LLM} Evaluation},
  journal = {arXiv preprint arXiv:2608.29856},
  year = {2026},
  doi = {10.48550/arXiv.2608.29856},
  url = {https://arxiv.org/abs/2608.29856v1},
  eprint = {2608.29856},
  archiveprefix = {arXiv}
}

@article{serpo,
  author = {Jianze Wang and Kunwang Zheng and Ying Liu and Yu Cao and Qilong Zhang and Jinlong Chen and Hua Yang and Qianglong Chen},
  title = {{SERPO}: Self-Evolving Rubric Policy Optimization for Open-Ended Test-Time Reinforcement Learning},
  journal = {arXiv preprint arXiv:2607.26873},
  year = {2026},
  doi = {10.48550/arXiv.2607.26873},
  url = {https://arxiv.org/abs/2607.26873v2},
  eprint = {2607.26873},
  archiveprefix = {arXiv}
}

@article{evors,
  author = {Weiyuan Li and Aili Chen and Xintao Wang and Yikai Zhang and Qingqing Dong and Jinghan Xu and Hongru Hou and Wenxuan Zhao and Chengkun Lang and Jun Gao and Yuanli Guo and Hongcheng Guo and Yanghua Xiao and Deqing Yang},
  title = {{EvoRS}: On-Policy Self-Evolution of Reward Systems for Open-Ended Reinforcement Learning},
  journal = {arXiv preprint arXiv:2609.12459},
  year = {2026},
  doi = {10.48550/arXiv.2609.12459},
  url = {https://arxiv.org/abs/2609.12459v1},
  eprint = {2609.12459},
  archiveprefix = {arXiv}
}

@inproceedings{self_rewarding,
  author = {Weizhe Yuan and Richard Yuanzhe Pang and Kyunghyun Cho and Xian Li and Sainbayar Sukhbaatar and Jing Xu and Jason E Weston},
  title = {Self-Rewarding Language Models},
  booktitle = {Proceedings of the 41st International Conference on Machine Learning},
  volume = {235},
  pages = {57905--57923},
  year = {2024},
  url = {https://proceedings.mlr.press/v235/yuan24d.html}
}

@article{self_taught,
  author = {Tianlu Wang and Ilia Kulikov and Olga Golovneva and Ping Yu and Weizhe Yuan and Jane Dwivedi-Yu and Richard Yuanzhe Pang and Maryam Fazel-Zarandi and Jason Weston and Xian Li},
  title = {Self-Taught Evaluators},
  journal = {arXiv preprint arXiv:2408.02666},
  year = {2024},
  doi = {10.48550/arXiv.2408.02666},
  url = {https://arxiv.org/abs/2408.02666v2},
  eprint = {2408.02666},
  archiveprefix = {arXiv}
}

@inproceedings{echo,
  author = {Zhicong Li and Lingjie Jiang and Yulan Hu and Xingchen Zeng and Yixia Li and Xiangwen Zhang and Guanhua Chen and Zheng Pan and Xin Li and Yong Liu},
  title = {No More Stale Feedback: Co-Evolving Critics for Open-World Agent Learning},
  booktitle = {Proceedings of the 64th Annual Meeting of the Association for Computational Linguistics (Volume 1: Long Papers)},
  pages = {12643--12660},
  year = {2026},
  doi = {10.18653/v1/2026.acl-long.576},
  url = {https://aclanthology.org/2026.acl-long.576/},
  eprint = {2601.06794},
  archiveprefix = {arXiv}
}

@article{shang2026backfires,
  author = {Linfang Shang and Ming Xu and Yiding Sun and Tianle Xia and Lingxiang Hu and Lan Xu and Ning Zheng},
  title = {When Self-Evolution Backfires: Pre-Commit Gating against Skill Contamination in {LLM} Agents},
  journal = {arXiv preprint arXiv:2608.05810},
  year = {2026},
  doi = {10.48550/arXiv.2608.05810},
  url = {https://arxiv.org/abs/2608.05810},
  eprint = {2608.05810},
  archiveprefix = {arXiv}
}

@article{roesner2026trustingtrust,
  author = {Franziska Roesner and Tadayoshi Kohno},
  title = {Reflections on Trusting Trust, Revisited: Contaminating Self-Modifying {AI} Coding Agents with Poisoned Benchmarks},
  journal = {arXiv preprint arXiv:2609.17817},
  year = {2026},
  doi = {10.48550/arXiv.2609.17817},
  url = {https://arxiv.org/abs/2609.17817},
  eprint = {2609.17817},
  archiveprefix = {arXiv}
}

@article{ord2026dynamics,
  author = {Toby Ord},
  title = {The Dynamics of Intelligence Explosions},
  journal = {arXiv preprint arXiv:2608.14426},
  year = {2026},
  doi = {10.48550/arXiv.2608.14426},
  url = {https://arxiv.org/abs/2608.14426},
  eprint = {2608.14426},
  archiveprefix = {arXiv}
}

@article{cunningham2026economics,
  author = {Tom Cunningham and Lukas Althoff and Basil Halperin and Brian Jabarian and Andrew Koh and Arjun Ramani and Phil Trammell and Parker Whitfill and Cheryl Wu},
  title = {The Economics of Recursive Self-Improvement},
  journal = {arXiv preprint arXiv:2609.15802},
  year = {2026},
  doi = {10.48550/arXiv.2609.15802},
  url = {https://arxiv.org/abs/2609.15802},
  eprint = {2609.15802},
  archiveprefix = {arXiv}
}

@article{whitfill2025compute,
  author = {Parker Whitfill and Cheryl Wu},
  title = {Will Compute Bottlenecks Prevent an Intelligence Explosion?},
  journal = {arXiv preprint arXiv:2507.23181},
  year = {2025},
  doi = {10.48550/arXiv.2507.23181},
  url = {https://arxiv.org/abs/2507.23181},
  eprint = {2507.23181},
  archiveprefix = {arXiv}
}

@article{evolvent2026rsibenchdata,
  author = {Meng, Fanqing and Du, Lingxiao and Chen, Qiguang and Zhao, Ziqi and Lu, Haocheng and Hu, Mengkang and Shieh, Michael Qizhe},
  title = {{RSIBench-Data}: Benchmarking Data-Centric Research for Recursive Self-Improvement},
  journal = {arXiv preprint arXiv:2607.25886},
  year = {2026},
  doi = {10.48550/arXiv.2607.25886},
  url = {https://arxiv.org/abs/2607.25886},
  eprint = {2607.25886},
  archiveprefix = {arXiv},
  note = {code and protocol: \url{https://github.com/evolvent-ai/RSIBench-Data}}
}

@inproceedings{zhong2026skilllearnbench,
  author = {Shanshan Zhong and Yi Lu and Jingjie Ning and Yibing Wan and Lihan Feng and Yuyi Ao and Leonardo F. R. Ribeiro and Markus Dreyer and Sean Ammirati and Chenyan Xiong},
  title = {{SkillLearnBench}: Benchmarking Continual Learning Methods for Agent Skill Generation on Real-World Tasks},
  booktitle = {Conference on Language Modeling},
  year = {2026},
  url = {https://arxiv.org/abs/2604.20087},
  eprint = {2604.20087},
  archiveprefix = {arXiv}
}

@article{zheng2026servelearnbench,
  author = {Haizhong Zheng and Yizhuo Di and Ranajoy Sadhukhan and Shuowei Jin and Beidi Chen},
  title = {{ServeLearnBench}: How Well Can Agents Self-Improve from Serving Experience?},
  journal = {arXiv preprint arXiv:2610.07792},
  year = {2026},
  doi = {10.48550/arXiv.2610.07792},
  url = {https://arxiv.org/abs/2610.07792},
  eprint = {2610.07792},
  archiveprefix = {arXiv}
}

@article{yu2026rsebench,
  author = {Tao Yu and Zhongtian Luo and Yuchen Long and Yiheng Lu and Yuxuan Zhou and Yanpei Gong and Siyi Cao and Yuqing Wu and Haopeng Jin and Zhenghao Zhang and Yujia Yang and Jingmin Zhu and Yunyun Hou and Xinming Wang and Hongzhu Yi and Jiachen Liu and Sirui Wang and Nan Duan and Yan Huang and Liang Wang and Tianjin Huang and Yanyan Zou},
  title = {When Non-Adversarial Noise Hurts Learning: Benchmarking Skill Self-Evolution in Agents},
  journal = {arXiv preprint arXiv:2610.19910},
  year = {2026},
  doi = {10.48550/arXiv.2610.19910},
  url = {https://arxiv.org/abs/2610.19910},
  eprint = {2610.19910},
  archiveprefix = {arXiv}
}

@article{shi2026s3gym,
  author = {Jiajun Shi and Siyuan Tao and Yuhao Wu and Zexuan Wang and Jingyuan Zhang and Jiaheng Liu and Xinping Lei and Xinrong Zhang and Siyuan Fang and Zhewen Tan and Tianle Cai and Junhao Fang and Jiameng Huang and Yueyang Wang and Jinkai Liu and Yuxuan Zhang and Jian Yang and Zhoujun Li and Shen Yan and Wenhao Huang and Ge Zhang},
  title = {{S\textsuperscript{3}Gym}: Can {LLMs} Turn Self-Testing and Self-Judging into Self-Improvement?},
  journal = {arXiv preprint arXiv:2608.31100},
  year = {2026},
  doi = {10.48550/arXiv.2608.31100},
  url = {https://arxiv.org/abs/2608.31100},
  eprint = {2608.31100},
  archiveprefix = {arXiv}
}

@article{wu2026harnessdev,
  author = {Yuhao Wu and Jingyuan Zhang and Jiajun Shi and Xinping Lei and Qingshui Gu and Yuxuan Zhang and Zexuan Wang and Chen He and Chen Huang and Maojia Song and Zhiyuan Zeng and Shaowen Wang and Jinkai Liu and Yunfeng Shi and Jiaheng Liu and Shen Yan and Wenhao Huang and Ge Zhang and Wenxuan Zhang},
  title = {{HarnessDev}: Can {LLMs} Create and Evolve Their Own Agent Harness?},
  journal = {arXiv preprint arXiv:2609.01437},
  year = {2026},
  doi = {10.48550/arXiv.2609.01437},
  url = {https://arxiv.org/abs/2609.01437},
  eprint = {2609.01437},
  archiveprefix = {arXiv}
}

@article{ge2026doc2harness,
  author = {Yuyao Ge and Shenghua Liu and Peizhong Gao and Chendi Ge and Chenxu Liu and Yiwei Wang and Lingrui Mei and Baolong Bi and Jiayu Yao and Jiafeng Guo and Xueqi Cheng},
  title = {{Doc2Harness}: Benchmarking {LLMs} on Long-Horizon Harness Engineering of Coding Agents},
  journal = {arXiv preprint arXiv:2610.20344},
  year = {2026},
  doi = {10.48550/arXiv.2610.20344},
  url = {https://arxiv.org/abs/2610.20344},
  eprint = {2610.20344},
  archiveprefix = {arXiv}
}

@article{ke2026evoharnessbench,
  author = {Zixuan Ke and Vaidehi Patil and Haizhou Shi and Yang Li and Ye Liu and Sarath Shekkizhar and Anurag Koul and Jiayu Wang and Xuan Phi Nguyen and Semih Yavuz and Mohit Bansal and Shafiq Joty},
  title = {{EvoHarnessBench}: Can Your Agents Keep Pace with an Evolving Harness?},
  journal = {arXiv preprint arXiv:2609.04280},
  year = {2026},
  doi = {10.48550/arXiv.2609.04280},
  url = {https://arxiv.org/abs/2609.04280},
  eprint = {2609.04280},
  archiveprefix = {arXiv}
}

@article{lei2026skillevolbench,
  author = {Yingtie Lei and Zhongwei Wan and Jiankun Zhang and Samiul Alam and Zixuan Zhong and Peizhou Huang and Xin Wang and Jingxuan Zhang and Donghao Zhou and Yunta Hsieh and Zhihao Dou and Hui Shen and Yan Xu and Dimitrios Dimitriadis and Tuo Zhang and Mi Zhang},
  title = {{SkillEvolBench}: Benchmarking the Evolution from Episodic Experience to Procedural Skills},
  journal = {arXiv preprint arXiv:2605.24117},
  year = {2026},
  doi = {10.48550/arXiv.2605.24117},
  url = {https://arxiv.org/abs/2605.24117},
  eprint = {2605.24117},
  archiveprefix = {arXiv}
}

@article{guan2026continualskillbench,
  author = {Tianyi Guan and Yiding Wang and Haotong Yang and Siyuan Cao and Shirui Liu and Yi Hu and Jiaqi Li and Muhan Zhang},
  title = {{ContinualSkillBench}: Can {LLM} Agents Truly Evolve Their Capabilities?},
  journal = {arXiv preprint arXiv:2608.03874},
  year = {2026},
  doi = {10.48550/arXiv.2608.03874},
  url = {https://arxiv.org/abs/2608.03874},
  eprint = {2608.03874},
  archiveprefix = {arXiv}
}

@article{xia2026toolgenesis,
  author = {Bowei Xia and Mengkang Hu and Shijian Wang and Jiarui Jin and Wenxiang Jiao and Yuan Lu and Kexin Li and Ping Luo},
  title = {{Tool-Genesis}: A Task-Driven Tool Creation Benchmark for Self-Evolving Language Agent},
  journal = {arXiv preprint arXiv:2603.05578},
  year = {2026},
  doi = {10.48550/arXiv.2603.05578},
  url = {https://arxiv.org/abs/2603.05578},
  eprint = {2603.05578},
  archiveprefix = {arXiv}
}

@article{zheng2026seagym,
  author = {Congjie Zheng and Chuanyi Xue and Bin Liang and Jun Yang and Changshui Zhang},
  title = {{SEAGym}: An Evaluation Environment for Self-Evolving {LLM} Agents},
  journal = {arXiv preprint arXiv:2606.17546},
  year = {2026},
  doi = {10.48550/arXiv.2606.17546},
  url = {https://arxiv.org/abs/2606.17546},
  eprint = {2606.17546},
  archiveprefix = {arXiv}
}

@article{liang2026pahf,
  author = {Kaiqu Liang and Julia Kruk and Shengyi Qian and Xianjun Yang and Shengjie Bi and Yuanshun Yao and Shaoliang Nie and Mingyang Zhang and Lijuan Liu and Jaime Fern{\'a}ndez Fisac and Shuyan Zhou and Saghar Hosseini},
  title = {Learning Personalized Agents from Human Feedback},
  journal = {arXiv preprint arXiv:2602.16173},
  year = {2026},
  doi = {10.48550/arXiv.2602.16173},
  url = {https://arxiv.org/abs/2602.16173},
  eprint = {2602.16173},
  archiveprefix = {arXiv}
}

@misc{openai_codex_cli,
  author = {{OpenAI}},
  title = {{Codex} {CLI}},
  url = {https://learn.chatgpt.com/docs/codex/cli},
  howpublished = {Official documentation},
  note = {Accessed: 2026-10-20}
}

@misc{anthropic_claude_code_memory,
  author = {{Anthropic}},
  title = {How {Claude} remembers your project},
  url = {https://code.claude.com/docs/en/memory},
  howpublished = {Official documentation},
  note = {Accessed: 2026-10-20}
}

@misc{openhands_persistent_memory,
  author = {{OpenHands}},
  title = {Persistent Memory},
  url = {https://docs.openhands.dev/sdk/guides/persistent-memory},
  howpublished = {Official documentation},
  note = {Accessed: 2026-10-20}
}

@misc{karpathy2026autoresearch,
  author = {Karpathy, Andrej},
  title = {autoresearch},
  year = {2026},
  url = {https://github.com/karpathy/autoresearch},
  howpublished = {Official source repository},
  note = {Accessed: 2026-10-20}
}

@inproceedings{lange2025shinka,
  author = {Lange, Robert Tjarko and Imajuku, Yuki and Cetin, Edoardo},
  title = {{ShinkaEvolve}: Towards Open-Ended And Sample-Efficient Program Evolution},
  booktitle = {International Conference on Learning Representations},
  year = {2026},
  doi = {10.48550/arXiv.2509.19349},
  url = {https://arxiv.org/abs/2509.19349},
  eprint = {2509.19349},
  archiveprefix = {arXiv}
}

@inproceedings{agrawal2025gepareflectivepromptevolution,
  author = {Agrawal, Lakshya A and Tan, Shangyin and Soylu, Dilara and Ziems, Noah and Khare, Rishi and Opsahl-Ong, Krista and Singhvi, Arnav and Shandilya, Herumb and Ryan, Michael J and Jiang, Meng and Potts, Christopher and Sen, Koushik and Dimakis, Alexandros G. and Stoica, Ion and Klein, Dan and Zaharia, Matei and Khattab, Omar},
  title = {{GEPA}: Reflective Prompt Evolution Can Outperform Reinforcement Learning},
  booktitle = {International Conference on Learning Representations},
  year = {2026},
  doi = {10.48550/arXiv.2507.19457},
  url = {https://arxiv.org/abs/2507.19457},
  eprint = {2507.19457},
  archiveprefix = {arXiv}
}

@article{wang2025evoagentx,
  author = {Wang, Yingxu and Liu, Siwei and Fang, Jinyuan and Meng, Zaiqiao},
  title = {{EvoAgentX}: An Automated Framework for Evolving Agentic Workflows},
  journal = {arXiv preprint arXiv:2507.03616},
  year = {2025},
  doi = {10.48550/arXiv.2507.03616},
  url = {https://arxiv.org/abs/2507.03616},
  eprint = {2507.03616},
  archiveprefix = {arXiv}
}

@misc{cursor_project_rules,
  author = {{Cursor}},
  title = {Rules},
  url = {https://cursor.com/docs/rules},
  howpublished = {Official documentation},
  note = {Accessed: 2026-10-20}
}

@misc{anomaly_opencode_rules,
  author = {{Anomaly}},
  title = {{OpenCode}: Rules},
  url = {https://opencode.ai/docs/rules/},
  howpublished = {Official documentation},
  note = {Accessed: 2026-10-20}
}

@article{meng2026swarmrsirecursiveselfimprovementevolving,
  author = {Qianru Meng and Xi Zhang and Konstantinos Papakostas and Congfeng Cao and Wenxiang Hu and Tianyi Lu and Xiao Zhang and Lun Zhou and Zaiqiao Meng and Zhaochun Ren and Xiaohui Yan and Yougang Lyu},
  title = {{SwarmRSI}: Recursive Self-Improvement through Evolving Scientific Agent Swarm},
  journal = {arXiv preprint arXiv:2610.20431},
  year = {2026},
  url = {https://arxiv.org/abs/2610.20431},
  eprint = {2610.20431},
  archiveprefix = {arXiv},
  primaryclass = {cs.AI}
}

@article{yang2026terminalrsirecursiveselfimprovingterminal,
  author = {Junyao Yang and Zongxia Li and Yucheng Shi and Zhongzhi Li and Ruhan Wang and Ruiwen Zhou and Haitao Mi and Shuicheng Yan and Leowei Liang},
  title = {{Terminal-RSI}: Recursive Self-Improving Terminal Agents via Recursive Self-Rewrite and Policy-Refine},
  journal = {arXiv preprint arXiv:2610.22192},
  year = {2026},
  url = {https://arxiv.org/abs/2610.22192},
  eprint = {2610.22192},
  archiveprefix = {arXiv},
  primaryclass = {cs.AI}
}

@article{zhang2026rsiflowadaptiveselfimprovementworkflows,
  author = {Huan Zhang and Mingju Chen and Dongxu Zhou and Can Lv and Sen Cui and Faguo Wu and Shiji Zhou},
  title = {{RSIFlow}: Adaptive Self-Improvement Workflows with an Evolving Meta-Agent},
  journal = {arXiv preprint arXiv:2610.21791},
  year = {2026},
  url = {https://arxiv.org/abs/2610.21791},
  eprint = {2610.21791},
  archiveprefix = {arXiv},
  primaryclass = {cs.AI}
}

@article{meng2026rsigymflexibleenvironmentrecursive,
  author = {Fanqing Meng and Lingxiao Du and Haocheng Lu and Qiguang Chen and Ziqi Zhao and Zijian Wu and Jiayuan Zhuo and Mengkang Hu and Michael Qizhe Shieh},
  title = {{RSIGym}: A Flexible Environment for Recursive Self-Improvement},
  journal = {arXiv preprint arXiv:2610.10310},
  year = {2026},
  url = {https://arxiv.org/abs/2610.10310},
  eprint = {2610.10310},
  archiveprefix = {arXiv},
  primaryclass = {cs.LG}
}

@article{weng2026groupevolvingagentsopenendedselfimprovement,
  author = {Zhaotian Weng and Antonis Antoniades and Deepak Nathani and Zhen Zhang and Xiao Pu and Xin Eric Wang},
  title = {Group-Evolving Agents: Open-Ended Self-Improvement via Experience Sharing},
  journal = {arXiv preprint arXiv:2602.04837},
  year = {2026},
  url = {https://arxiv.org/abs/2602.04837},
  eprint = {2602.04837},
  archiveprefix = {arXiv},
  primaryclass = {cs.AI}
}

@article{paul2026hgmduallineageawareguidedagent,
  author = {Sudipta Paul and Pengrui Quan and Aounon Kumar and Nicholas Tzou and Vivek Kulkarni and Srinivas Chappidi},
  title = {{HGM-Dual}: Lineage-Aware Guided Agent Harness Optimization for Long-Horizon Planning},
  journal = {arXiv preprint arXiv:2610.30042},
  year = {2026},
  url = {https://arxiv.org/abs/2610.30042},
  eprint = {2610.30042},
  archiveprefix = {arXiv},
  primaryclass = {cs.AI}
}

@article{zhou2026rsicontrolaimodels,
  author = {Kaiwen Zhou and Udari Madhushani Sehwag},
  title = {{RSI} Control: Can {AI} Models Perform Covert Sabotage in the Self-Improvement Process?},
  journal = {arXiv preprint arXiv:2610.27735},
  year = {2026},
  url = {https://arxiv.org/abs/2610.27735},
  eprint = {2610.27735},
  archiveprefix = {arXiv},
  primaryclass = {cs.AI}
}

@article{zhang2026r2flowrecursiveselfimprovement,
  author = {Mingda Zhang and Qiang Huang and Yanjin Li and Zijia Wang and Qika Lin and Xiaoying Tang and Tiesunlong Shen},
  title = {R$^2$ Flow: Recursive Self-Improvement via Recursive Skill Evolution},
  journal = {arXiv preprint arXiv:2609.33867},
  year = {2026},
  url = {https://arxiv.org/abs/2609.33867},
  eprint = {2609.33867},
  archiveprefix = {arXiv},
  primaryclass = {cs.AI}
}

@article{lee2026recursiveharnessselfimprovement,
  author = {Hyunin Lee and Jinglue Xu and Jeffrey Seely and Donghyun Lee and Matei Zaharia and Yujin Tang},
  title = {Recursive Harness Self-Improvement},
  journal = {arXiv preprint arXiv:2607.15524},
  year = {2026},
  url = {https://arxiv.org/abs/2607.15524},
  eprint = {2607.15524},
  archiveprefix = {arXiv},
  primaryclass = {cs.LG}
}

@article{wei2026selfevolvingcodingagentsharnesspolicy,
  author = {Zhepei Wei and Wenlin Yao and Baolin Peng and Danqing Wang and Qianhui Wu and Hao Cheng and Yu Meng and Jianfeng Gao},
  title = {Self-Evolving Coding Agents via Harness-Policy Optimization},
  journal = {arXiv preprint arXiv:2610.22791},
  year = {2026},
  url = {https://arxiv.org/abs/2610.22791},
  eprint = {2610.22791},
  archiveprefix = {arXiv},
  primaryclass = {cs.CL}
}

@article{chen2026coevocoevolvingcontextparameters,
  author = {Siwei Chen and Xinping Bao and Xinyu Cai and Yuan Cao and Wan Jiang and Shaohong Chen},
  title = {{COEVO}: Co-Evolving Context and Parameters for Recursive Self-Improvement},
  journal = {arXiv preprint arXiv:2609.33398},
  year = {2026},
  url = {https://arxiv.org/abs/2609.33398},
  eprint = {2609.33398},
  archiveprefix = {arXiv},
  primaryclass = {cs.AI}
}

@inproceedings{du2026generalizableendtoendtooluserl,
  author = {Weihua Du and Hailei Gong and Zhan Ling and Kang Liu and Lingfeng Shen and Xuesong Yao and Yufei Xu and Dingyuan Shi and Yiming Yang and Jiecao Chen},
  title = {Generalizable End-to-End Tool-Use {RL} with Synthetic {CodeGym}},
  booktitle = {International Conference on Learning Representations},
  year = {2026},
  url = {https://proceedings.iclr.cc/paper_files/paper/2026/hash/1e4322fddd833f83c855660ac65e428d-Abstract-Conference.html},
  eprint = {2509.17325},
  archiveprefix = {arXiv},
  primaryclass = {cs.LG}
}

@article{lv2026agentsgeneralizeopenworld,
  author = {Song-Lin Lv and Weiming Wu and Rui Zhu and Zi-Jian Cheng and Lan-Zhe Guo},
  title = {Can Agents Generalize to the Open World? Unveiling the Fragility of Static Training in Tool Use},
  journal = {arXiv preprint arXiv:2607.01084},
  year = {2026},
  url = {https://arxiv.org/abs/2607.01084},
  eprint = {2607.01084},
  archiveprefix = {arXiv},
  primaryclass = {cs.AI}
}
